\documentclass{article}

\PassOptionsToPackage{numbers}{natbib}
\usepackage[preprint]{neurips_2025}

\usepackage{changepage}
\usepackage[utf8]{inputenc} 
\usepackage[T1]{fontenc}    
\usepackage{hyperref}       
\usepackage{url}            
\usepackage{booktabs}       
\usepackage{amsfonts}       
\usepackage{nicefrac}       
\usepackage{microtype}      
\usepackage{graphicx}
\usepackage{caption}
\usepackage{xspace}
\usepackage{enumitem}
\usepackage{array}
\usepackage[table]{xcolor}
\usepackage{multirow} 
\usepackage{tabularx}
\usepackage{lineno}
\usepackage{amsmath}
\usepackage{booktabs,adjustbox}
\newcommand{\datasetname}{OCRBench v2\xspace}

\title{OCRBench v2: An Improved Benchmark for Evaluating Large Multimodal Models on Visual Text Localization and Reasoning}

\author{
\textbf{Ling Fu}$^1$ \quad 
\textbf{Zhebin Kuang}$^{1}$ \quad \textbf{Jiajun Song}$^1$ \quad \textbf{Mingxin Huang}$^2$ \quad \textbf{Biao Yang}$^1$ \quad \\
\textbf{Yuzhe Li}$^1$ \quad \textbf{Linghao Zhu}$^{1}$ \quad \textbf{Qidi Luo}$^{1}$ \quad \textbf{Xinyu Wang}$^{1}$ \quad \textbf{Hao Lu}$^{1}$ \quad \textbf{Zhang Li}$^{1}$\\
\textbf{Guozhi Tang}$^{4}$ \quad \textbf{Bin Shan}$^{4}$ \quad \textbf{Chunhui Lin}$^{4}$ \quad \textbf{Qi Liu}$^{4}$ \quad \textbf{Binghong Wu}$^{4}$ \\
\textbf{Hao Feng}$^{4}$ \quad \textbf{Hao Liu}$^{4}$ \quad \textbf{Can Huang}$^{4}$ \quad \textbf{Jingqun Tang}$^{4}$ \quad \textbf{Wei Chen}$^{1}$ \\
\textbf{Lianwen Jin}$^{2}$ \quad \textbf{Yuliang Liu}$^{1}$ \quad \textbf{Xiang Bai}$^{1}$ \\
\\
$^1$Huazhong University of Science and Technology \quad $^2$South China University of Technology \\ $^{3}$University of Adelaide \quad $^{4}$ByteDance
}

\begin{document}

\maketitle

\begin{abstract}
  Scoring the Optical Character Recognition (OCR) capabilities of Large Multimodal Models (LMMs) has witnessed growing interest. Existing benchmarks have highlighted the impressive performance of LMMs in text recognition; however, their abilities in certain challenging tasks, such as text localization, handwritten content extraction, and logical reasoning, remain underexplored. To bridge this gap, we introduce \textbf{\datasetname}, a large-scale bilingual text-centric benchmark with currently the most comprehensive set of tasks ($4\times$ more tasks than the previous multi-scene benchmark OCRBench), the widest coverage of scenarios ($31$ diverse scenarios), and thorough evaluation metrics, with $10,000$ human-verified question-answering pairs and a high proportion of difficult samples. Moreover, we construct a private test set with $1,500$ manually annotated images. The consistent evaluation trends observed across both public and private test sets validate the \datasetname's reliability. After carefully benchmarking state-of-the-art LMMs, we find that most LMMs score below $50$ ($100$ in total) and suffer from five-type limitations, including less frequently encountered text recognition, fine-grained perception, layout perception, complex element parsing, and logical reasoning. The project website is at: 
\href{https://99franklin.github.io/ocrbench_v2/}{https://99franklin.github.io/ocrbench\_v2}.
\end{abstract}

\vspace{-2mm}
\section{Introduction}
\label{Introduction}
The emergence of Large Language Models (LLMs)~\cite{achiam2023gpt, touvron2023llama, brown2020language} has greatly improved the understanding and generation of structured text. However, in reality, much of the textual content is unstructured; it appears within images, videos, and other non-textual media in varied positions, orientations, and shapes. The need for processing such unstructured content leads to the study of Large Multimodal Models (LMMs)~\cite{bai2023qwen, liu2024visual, zhu2023minigpt} that extend the text-only LLMs to additional modalities. By pretraining on multimodal data, LMMs acquire the zero-shot ability to interpret across diverse media, such as recognizing and understanding complex visual scene text~\cite{liu2024textmonkey}. Such capability represents a significant advancement over standard Optical Character Recognition (OCR), because LMMs not only spot text but also interpret its semantic relevance to a scene.

\begin{figure}[t]
    \centering
    \includegraphics[width=0.95\linewidth]{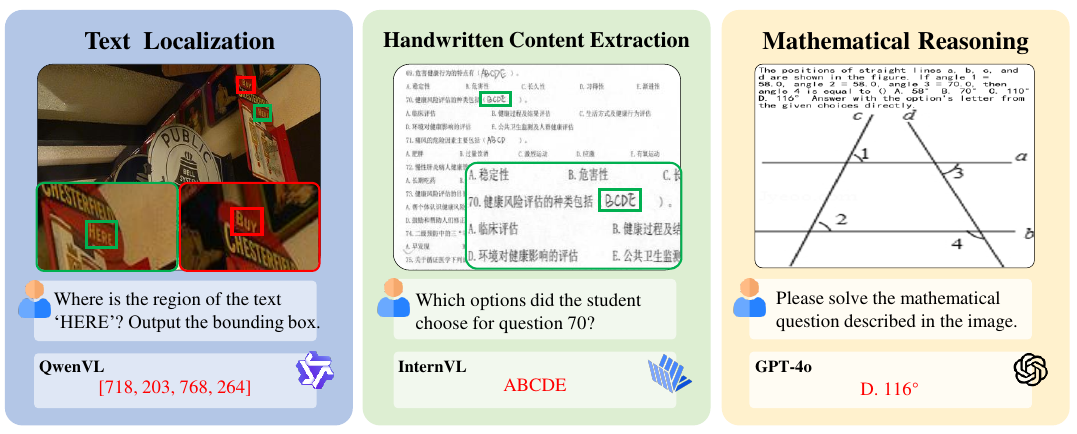}
    \captionsetup{width=0.95\linewidth}
    \caption{\textbf{Large multimodal models struggle with text-intensive tasks accurately}. They are prone to errors in tasks like text localization, handwritten content extraction, and mathematical reasoning, revealing limitations in tackling complex textual information within images.}
    \label{fig:gpt_limitations}
    \vspace{-15pt}
\end{figure}

Compared with classic OCR that typically relies on task-specific models to spot text, the increasing capability of LMMs to process multimodal inputs has opened new potential to redefine the area of OCR. OCR has therefore become an important aspect of recent LMM evaluations. Some text-focused tasks have been included in standard benchmarks to assess the proficiency of LMMs in recognizing and interpreting textual content~\cite{fu2023mme, ying2024mmt}. Typically, text-based Visual Question Answering (VQA) datasets~\cite{singh2019towards, biten2019scene, wang2020general} are repurposed to evaluate OCR by framing generic VQA into questions that require accurate reading of embedded text. However, many of these datasets are initially created for classic OCR models, which are of limited diversity, depth, and suitability for evaluating LMMs. A common drawback is that, many questions lack sufficient complexity to assess the reasoning abilities of LMMs on scene text, and some can even be answered without visual input~\cite{chen2024we, wang2020general}. 


More recently, several customized benchmarks~\cite{liu2023hidden, li2024seed, wadhawancontextual, liu2024focus, kim2024tablevqa} have explored the OCR capabilities of LMMs. For example, OCRBench~\cite{liu2023hidden} consolidates $5$ core text-oriented tasks to evaluate LMM performance across traditional OCR functions. Other datasets, such as ComTQA~\cite{zhao2024tabpedia} and ChartX~\cite{xia2024chartx}, focus on structured text interpretation like table and chart understanding. While such effort represents a leap over standard OCR benchmarks, they remain limited in both data diversity and quantity (see Tab.~\ref{tab:comparison-existing}), often leading to rapid performance saturation. For example, recent LMMs such as Qwen2.5-VL~\cite{qwen2.5-VL} have achieved 96.4\% accuracy on the DocVQA dataset~\cite{mathew2021docvqa}, nearly matching human performance at 98.1\%, and 88.8\% on OCRBench~\cite{liu2023hidden}. This raises an important question for the community:
\textit{Do models perform well enough on text-oriented visual understanding tasks in the LMM era, or do existing benchmarks fail to capture the broader challenges in diverse environments?}

To answer the question above, we conducted preliminary tests with several state-of-the-art LMMs, including Qwen2.5-VL-7B~\cite{qwen2.5-VL}, InternVL3-14B~\cite{chen2024expanding}, and GPT-4o~\cite{openai2024gpt4o}. These tests assessed performance on text-oriented tasks, such as text localization, handwritten content extraction, and document-based logical reasoning. As illustrated in Fig.~\ref{fig:gpt_limitations}, each model can fail on one of the text-intensive tasks. These failures reveal a gap in detailed visual perception across different models, which constrains their effectiveness in tasks requiring accurate text localization, recognition, and contextual understanding within images. Recent benchmarks, such as OmniDocBench~\cite{ouyang2024omnidocbench}, CC-OCR~\cite{yang2024cc}, and MMLONGBENCH-DOC~\cite{ma2024mmlongbench}, have broadened evaluation to cover more comprehensive scenarios, including fine-grained document parsing and multi-page document understanding. Their analyses reveal the limited capabilities of LMMs for practical OCR applications and highlight the growing need for benchmarks that allow for more robust and varied evaluation of LMMs.

To bridge this gap, we propose \textit{\datasetname}, a comprehensive benchmark designed to assess LMMs across diverse text-oriented visual understanding tasks. As shown in Fig.~\ref{fig:all_tasks}, \textit{\datasetname} assesses eight core text-reading abilities, including \textit{text recognition}, \textit{text referring}, \textit{text spotting}, \textit{relation extraction}, \textit{element parsing}, \textit{mathematical calculation}, \textit{visual text understanding}, and \textit{knowledge reasoning}, organized into a total of $23$ concrete tasks. This benchmark provides $10,000$ high-quality, human-validated instruction-response pairs and also six types of evaluation metrics, which offers a rigorous framework for evaluating LMM performance in complex, practical OCR scenarios. For better evaluation quality, we further collect and label $1,500$ additional text-images from scratch, reserved as the private test set. This private data serves as an independently curated test set to validate model generalization. In summary, the contributions of this work are three-fold:

\vspace{-5pt}

\begin{itemize}[leftmargin=1em, itemsep=0.1em]
    \item \textit{\datasetname}: an improved benchmark designed to assess eight core OCR competencies and covers $23$ tasks across $31$ diverse scenarios, which provides a thorough evaluation framework encapsulating fundamental and advanced text-centric challenges.
    
    \item We systematically evaluate state-of-the-art LMMs, ranging from commercial APIs to open-source models, which establishes broad baselines for OCR performance and enables a comparative understanding of model capabilities across varied text-oriented visual understanding tasks.
    
    \item We provide a detailed analysis to identify factors affecting the OCR capabilities of LMMs. The analysis examines performance across various dimensions such as model generalization to diverse text types, model robustness, and the ability to tackle complex visual-textual relations.
\end{itemize}

\begin{table}[t]
\centering
\small
\renewcommand{\arraystretch}{0.9}
\setlength{\tabcolsep}{3pt}
\addtolength{\tabcolsep}{2pt}
  \centering
  \begin{tabular}{@{}lcccc@{}}
    \toprule
    Benchmark & \#Scenario & \#Task & \#Image & \#Instruction \\
    \midrule
    OCRbench~\cite{liu2023hidden} & $ \sim 14 $  & 5 & 0.9k & 1k \\
    Seed-bench-2-plus~\cite{li2024seed} & $ \sim 8 $ & 1 & 0.6k & 2.3k \\
    CONTEXTUAL~\cite{wadhawancontextual} & $ \sim 11 $ & 1 & 0.5k & 0.5k \\
    Fox~\cite{liu2024focus} & 2 & 9 & 0.7k & 2.2k \\
    MMTab-eval~\cite{DBLP:conf/acl/ZhengFSS0J024} & 1 & 9 & 23k & 49k \\
    ComTQA~\cite{zhao2024tabpedia} & 1 & 4 & 1.6k & 9k \\
    ChartX~\cite{xia2024chartx} & 1 & 7  & 6k & 6k \\
    MMC~\cite{liu2024mmc} & 1 & 9 & 1.7k & 2.9k \\
    OmniDocBench~\cite{ouyang2024omnidocbench} & 9 & 5 & 1k & 1k \\
    MMLONGBENCH-DOC~\cite{ma2024mmlongbench} & 7 & 2 & 6.4k & 1.1k \\
    \datasetname (Ours) & 31 & 23 & 9.5k & 10k \\
    \bottomrule
  \end{tabular}
  \vspace{4pt}
  \caption{Comparison between the proposed benchmark and existing text-centric datasets.}
  \label{tab:comparison-existing}
  \vspace{-18pt}
\end{table}

\section{Related Work}
\label{RelatedWork}



\paragraph{OCR-Enhanced LMMs.} Inspired by LLMs, visual encoders are integrated into them to create LMMs capable of processing both images and text. Early LMMs exhibit strong zero-shot OCR capabilities, motivating the exploration of text-centric LMMs. For instance, some work~\cite{zhang2023llavar, ye2023mplug} use text-centric instruction-tuning to enhance OCR-related abilities. But they are restricted to low-res inputs, limiting the ability to recognize dense and small text. To address this, several studies~\cite{feng2023docpedia, ye2023ureader, luo2024layoutllm} shift attention to increasing the input resolution. As the resolution of inputs increases, so does computational cost. To tackle this issue, TextMonkey~\cite{liu2024textmonkey} introduces a Token Resampler to compress redundant visual feature tokens, mPLUG-DocOwl2~\cite{hu2024mplug} presents a DocCompressor module for compressing high-res images, and DocKylin~\cite{zhang2025dockylin} adopts adaptive pixel slimming and dynamic token slimming modules to reduce redundant regions. To enhance layout perception, DocLayLLM~\cite{liao2024doclayllm} integrates layout information into LMMs inputs, LayTokenLLM~\cite{zhu2025simple} shares position IDs between text and layout tokens, DocMark~\cite{xiao2025docmark} utilizes adaptive generation of markup languages to build structured document representations, while Marten~\cite{wang2025marten} introduces an additional mask generator during pre-training. Despite strong results on existing benchmarks, challenges remain unsolved in certain key areas such as text localization, entity extraction, and logical reasoning. 

\vspace{-5pt}

\paragraph{Benchmarks for Text-Centric LMMs.}
Previous efforts have focused on creating scenario-specific benchmarks to assess LMMs. For example, DocVQA~\cite{mathew2021docvqa}, ChartQA~\cite{masry2022chartqa}, Infographics VQA~\cite{mathew2022infographicvqa}, and TextVQA~\cite{singh2019towards} evaluate models on document understanding, chart reasoning, infographic interpretation, and scene text comprehension, respectively. To broaden evaluation scope, OCRBench~\cite{liu2023hidden} introduces a holistic evaluation framework covering five text-oriented tasks, while CONTEXTUAL~\cite{wadhawancontextual} and SEED-Bench-2-Plus~\cite{li2024seed} introduce context-sensitive and diverse real-world images. Other benchmarks target specific challenges such as dense text understanding~\cite{zhang2024exploring}, complex structure parsing~\cite{yang2024cc}, and fine-grained document analysis~\cite{ouyang2024omnidocbench}. To provide a more thorough assessment, some benchmarks design multiple tasks within a specific scenario. TableVQA-Bench~\cite{kim2024tablevqa}, MMTab~\cite{DBLP:conf/acl/ZhengFSS0J024}, and ComTQA~\cite{zhao2024tabpedia} explore table-based tasks, while ChartY~\cite{chen2024onechart}, ChartX~\cite{xia2024chartx}, and MMC~\cite{liu2024mmc} focus on chart information extraction and reasoning. Recently, DUDE~\cite{van2023document}, MM-NIAH~\cite{wang2025needle}, MP-DocVQA~\cite{tito2023hierarchical}, MMLONGBENCH-DOC~\cite{ma2024mmlongbench}, and LongDocURL~\cite{deng2024longdocurl} explore the long document understanding capability of LMMs. In this work, we establish \textit{\datasetname}, a systematic benchmark to reveal the limitations of LMMs in diverse single-image, text-related scenarios.

\section{Why Do We Need OCRBench v2?}
\label{WhyNeedOCRBenchv2}
\begin{figure}[!t]
  \centering   \includegraphics[width=1\linewidth]{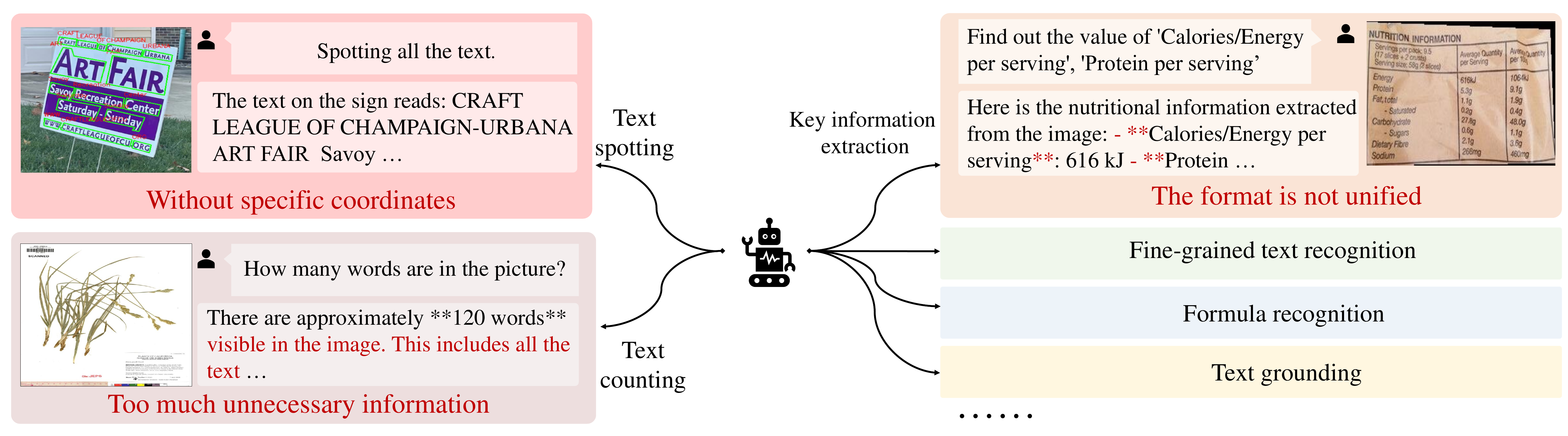}
   \captionsetup{width=1\linewidth}
   \caption{As evaluation for LMMs expands to diverse text-oriented tasks, existing datasets often require task-specific handling, making unified and scalable evaluation difficult.}
   \label{fig:why_need}
   \vspace{-5pt}
\end{figure}

\paragraph{Limitations of Existing Benchmarks.}
Recent evaluations of LMMs' OCR capabilities have made significant progress, yet most existing benchmarks exhibit limitations. Datasets like DocVQA, ChartQA, and TextVQA are often narrow in scope, focusing predominantly on text recognition within specific domains such as forms, tables, or documents. While useful for isolated capabilities, they fall short in task diversity, instruction complexity, and structured output formats that better reflect the multimodal nature of LMMs. In particular, many of these benchmarks were originally tailored for traditional OCR systems that prior to the emergence of LMMs. Furthermore, as illustrated in Fig.~\ref{fig:why_need}, complex task-specific processes are needed for LMMs when extended to more text-oriented tasks, which limits the evaluation of their broader capabilities.

\vspace{-5pt}
\paragraph{The Necessity of Unified Multi-task Evaluation.}
With the emergence of LMMs, current models show generalization ability to handle multiple tasks. To assess these multi-task models, unified benchmarks like LongDocURL~\cite{deng2024longdocurl}, OmniDocBench~\cite{ouyang2024omnidocbench}, CCOCR~\cite{yang2024cc}, OCRBench~\cite{liu2023hidden}, CONTEXTUAL~\cite{wadhawancontextual}, SEED-Bench-2-Plus~\cite{li2024seed}, have been proposed and successfully demonstrated the value of evaluating text-oriented models across diverse tasks. These benchmarks show the importance of unified evaluation frameworks in guiding model development. 
However, as model capabilities expand, existing benchmarks with limited task coverage result in fragmented and sometimes misleading insights. To address this, a unified benchmark is essential to: 1) \textit{Understand generalization}: Can a model perform consistently across varied text-centric tasks? 2) \textit{Diagnose failure models}: Does a model that excels in recognition also succeed in reasoning, localization, and parsing? 3) \textit{Guide model development}: Unified evaluation provides clearer signals for architecture and training improvements.

\begin{figure*}[t]
  \centering   
  \includegraphics[width=\linewidth]{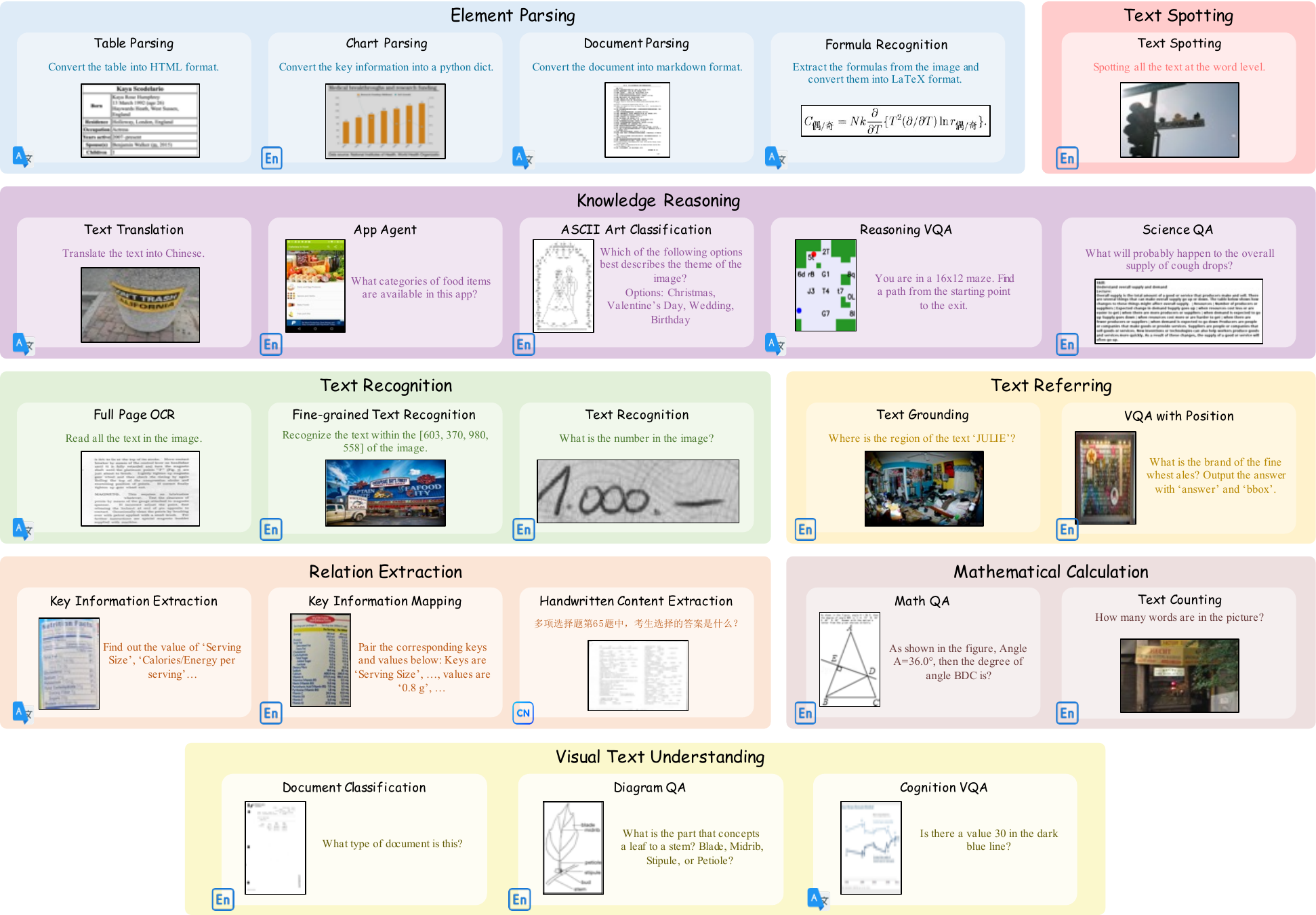}
  \caption{\textbf{Sample visualizations for each task.} OCRBench v2 comprises 23 sub-tasks grouped under 8 core OCR capabilities. Tasks marked with~\includegraphics[width=0.02\linewidth]{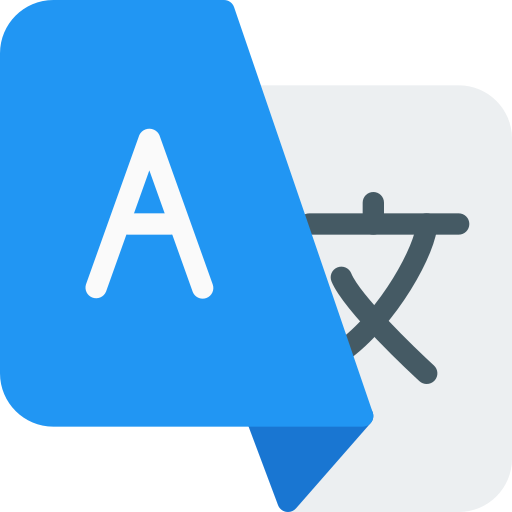} contain both English and Chinese instructions, while other tasks are either English-only~\includegraphics[width=0.02\linewidth]{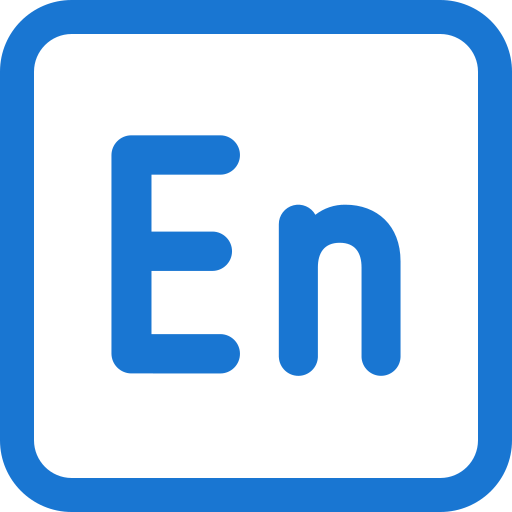} or Chinese-only~\includegraphics[width=0.02\linewidth]{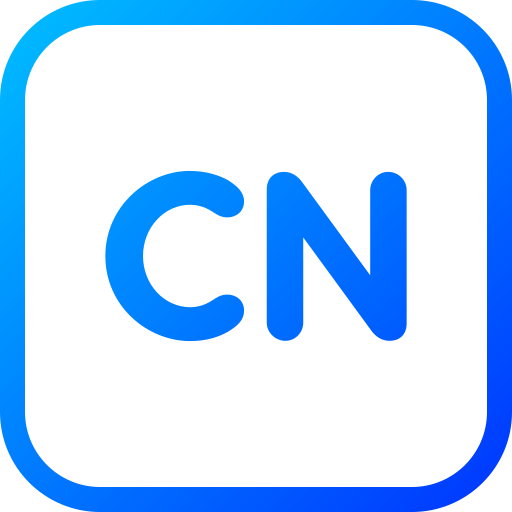}  (Zoomed in for better clarity).}
   \label{fig:all_tasks}
   \vspace{-5pt}
\end{figure*}

As shown in Fig.~\ref{fig:all_tasks}, \textit{\datasetname} tackles this by combining 23 tasks under 8 core capabilities within one framework. This holistic design enables systematic comparison of models and highlights trade-offs (e.g., performance on reasoning vs. recognition) that isolated benchmarks cannot reveal.

\vspace{-5pt}
\paragraph{How OCRBench v2 Addresses the Gaps.}

\textit{\datasetname} is a comprehensive, and high-difficulty benchmark specifically built to evaluate LMMs in realistic OCR settings, with key advantages: 1) \textit{Breadth of coverage}: With 31 scenarios, we ensure diverse contextual challenges; 2) \textit{Task variety}: The benchmark spans 8 OCR-related capabilities, many of which are poorly handled by current LMMs; 3) \textit{Instruction complexity}: Human-authored prompts and structured outputs (e.g., Markdown, JSON, LaTeX) raise the bar beyond simple answer extraction; 4) \textit{Private evaluation test set}: To prevent overfitting and training contamination, we additionally provide a private test set. 

Ultimately, \textit{\datasetname} fills a critical gap by offering a unified and challenging benchmark that reflects the practical needs of OCR in the LMM era. It not only measures what current models can do, but more importantly, reveals what they still cannot.

\vspace{-5pt}
\paragraph{Design Rationale: Focusing on Single-Image Text Tasks.}
While designing \textit{\datasetname}, we focus on challenges in single-image, text-related scenarios, and do not extend our study to multi-image tasks. This design choice is grounded in two considerations: 1) Single-image understanding is the foundation for more complex multimodal tasks. Many existing models still perform unsatisfactorily in various single-image scenarios, which motivates our work; 2) Given long-context inputs, multi-page tasks have more emphasis on long-sequence modeling, requiring specific benchmarks to assess this capability individually. For example, MMLONGBENCH-DOC focuses on evaluating the ability of LMMs to locate and understand content across pages in long documents.

\vspace{-5pt}
\paragraph{Private Dataset for Reliable Evaluation.}
To further enhance the assessment quality, we also construct a private test set. This data comprises $1,500$ manually collected text-rich images with human-annotated labels, covering 23 tasks aligned with the distribution of the public data. The images are from diverse sources, including printed books, e-books, scanned documents, and web content. During data collection and annotation, we meticulously curated samples to align with practical text-oriented applications. Given that benchmarks may be contaminated in massive internet-scraped pre-training data of LMMs, this data will not be released. Instead, we maintain a regularly updated leaderboard to reflect the performance of advanced LMMs. Moreover, consistent performance trends and model rankings observed on both the public and private test sets (see Section~\ref{main_results}) indicate the benchmark’s well-founded design and its effectiveness in identifying model capabilities.

\section{Benchmark Construction}
\label{DataConstruction}

In this section, we describe the task description, annotation curation, statistics, and evaluation criteria. Due to space limitations, more details can be found in the Appendix.

\subsection{Task Description}
\label{sec:3_1_task_description}

To provide a comprehensive evaluation framework for text-reading tasks, we categorize OCR capabilities into eight core areas, each encompassing specific sub-tasks that address various aspects of text comprehension and interpretation. Fig.~\ref{fig:all_tasks} exhibits samples for each task, with visual inputs and corresponding instructions. Detailed descriptions of these core capabilities are as follows.

\noindent\textbf{Text Recognition.} This fundamental capability focuses on perceiving textual content. The related tasks include (fine-grained) text recognition and full-page OCR. 

\noindent\textbf{Text Referring.} Determining the location of texts accurately is necessary for real-world OCR applications. This ability is evaluated with text grounding and VQA with position tasks.

\noindent\textbf{Text Spotting.} 
Text spotting is a widely studied OCR task that requires models to output both the location and content of text. We consider it a distinct capability due to this unique output format.

\noindent\textbf{Relation Extraction.} Given that texts are often densely arranged in images, the ability to extract and map visual components is essential. This capability is assessed through key information extraction, key information mapping, and handwritten content extraction. 

\noindent\textbf{Element Parsing.} LMMs face the need of parsing complex elements for downstream applications. This ability is evaluated via table parsing, chart parsing, document parsing, and formula recognition. 

\noindent\textbf{Mathematical Calculation.} Math calculation is essential for LMMs to address numerical reasoning tasks. Hence, text counting is introduced to assess the textual perception ability. Besides, we enhance the math QA data by rendering textual questions into images, accompanied by geometric figures.

\noindent\textbf{Visual Text Understanding.} To tackle sophisticated tasks involving human interaction, LMMs need to comprehend the semantic information of texts, a capability we term visual text understanding. This ability is evaluated by document classification and diagram QA. Additionally, we include basic VQA instructions where answers are located directly within the image, which refers to cognition VQA.

\noindent\textbf{Knowledge Reasoning.} Some tasks require complex inference and world knowledge, including science QA, APP agent interactions, ASCII art classification, text translation, and reasoning VQA (where answers are not directly visible in images).

\subsection{Annotation Curation}

\noindent\textbf{Dataset Collection.} To ensure data diversity, we manually harvest and screen $81$ text-rich academic datasets. To ensure diverse scenario coverage, we also supplement them with additional private data. In all, our dataset comprises $31$ typical scenarios (see Appendix for the full list).

\noindent\textbf{Instruction Formatting.} To convert existing annotations into the LMM-compatible instruction format, we design specific prompts for each task. For complex tasks such as document parsing that require structured output, we include a format example to minimize the impact of instruction-following ability and focus the evaluation on OCR capabilities. Additionally, considering the distinct training strategies for localization tasks in LMMs, we standardize the coordinates by normalizing them with image sizes and by scaling to the range of $[0, 1000]$. Such a standardization is explicitly specified in the related prompts. For the completeness of the tasks and scenarios, we carefully annotate the additional images to ensure they align with the designed task requirements. 

\noindent\textbf{Manual Verification.} To ensure data quality, we manually review all instructions for public data and correct approximately $1\%$ annotation errors.

\begin{figure}[!t]
  \centering   
  \includegraphics[width=\linewidth]{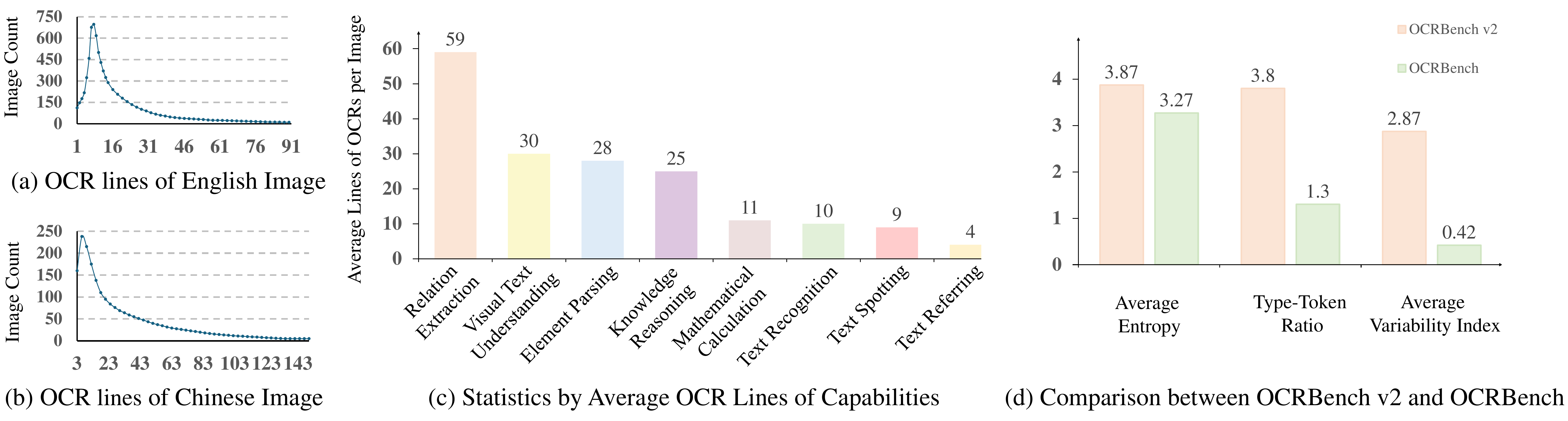}
   \captionsetup{width=1\linewidth}\vspace{-5pt}
   \caption{\textbf{OCR-related statistics and prompt quality assessment of OCRBench v2}.}
   \label{fig:ocr_statistics}
   \vspace{-5pt}
\end{figure}

\subsection{Statistics of OCRBench v2}
Here we present the OCR-related statistics and the measurement of prompt quality. As shown in Fig.~\ref{fig:ocr_statistics} (a) and (b), we count the distribution of line-level OCR results of $7,400$ English and $2,600$ Chinese images. And Fig.~\ref{fig:ocr_statistics} (c) exhibits the average number of line-level OCR results per category. These statistics demonstrate that the text information is sufficiently rich in \textit{\datasetname}. In addition, Fig.~\ref{fig:ocr_statistics} (d) compares the Average Entropy, Type-Token Ratio, and Average Variability Index of the questions between \textit{\datasetname} and OCRBench. \textit{\datasetname} presents higher values across all three metrics, indicating more diverse, less redundant, and structurally varied questions. This suggests it provides a more comprehensive and challenging benchmark for LMMs.

\begin{table*}
  \centering
  \renewcommand{\arraystretch}{0.9} 
  \addtolength{\tabcolsep}{-2pt} 
  \fontsize{9}{11}\selectfont
  \resizebox{\textwidth}{!}{
  \begin{tabular}{ 
                  @{}
                  l
                  >{\centering\arraybackslash}p{1.25cm}
                  >{\centering\arraybackslash}p{1.25cm}
                  >{\centering\arraybackslash}p{1.25cm}
                  >{\centering\arraybackslash}p{1.25cm}
                  >{\centering\arraybackslash}p{1.25cm}
                  >{\centering\arraybackslash}p{1.25cm}
                  >{\centering\arraybackslash}p{1.25cm}
                  >{\centering\arraybackslash}p{1.25cm}
                  >{\centering\arraybackslash}p{1.25cm}
                  @{}
                  }
    \toprule
    {\fontsize{7.5}{9}\selectfont Method} & 
    \multicolumn{1}{>{\centering\arraybackslash}p{1.25cm}}{\fontsize{7.5}{9}\selectfont Recognition} & 
    \multicolumn{1}{>{\centering\arraybackslash}p{1.25cm}}{\fontsize{7.5}{9}\selectfont Referring} & 
    \multicolumn{1}{>{\centering\arraybackslash}p{1.25cm}}{\fontsize{7.5}{9}\selectfont Spotting} &{\fontsize{8}{9}\selectfont Extraction} & 
    \multicolumn{1}{>{\centering\arraybackslash}p{1.25cm}}{\fontsize{7.5}{9}\selectfont Parsing} & 
    \multicolumn{1}{>{\centering\arraybackslash}p{1.25cm}}{\fontsize{7.5}{9}\selectfont Calculation} & 
    \multicolumn{1}{>{\centering\arraybackslash}p{1.25cm}}{\fontsize{7.5}{9}\selectfont Understanding} & 
    \multicolumn{1}{>{\centering\arraybackslash}p{1.25cm}}{\fontsize{7.5}{9}\selectfont Reasoning} & 
    \multicolumn{1}{>{\centering\arraybackslash}p{1.25cm}}{\fontsize{7.5}{9}\selectfont Average} \\
    \midrule
    \rowcolor{gray!20}
    \multicolumn{10}{c}{Open-source LMMs} \\
    LLaVA-Next-8B~\cite{liu2024llava} & 41.3 & 18.8 & 0 & 49.5 & 21.2 & 17.3 & 55.2 & 48.9 & 31.5 \\
    LLaVA-OV-7B~\cite{li2024llava} & 46.0 & 20.8 & 0.1 & 58.3 & 25.3 & 23.3 & 64.4 & 53.0 & 36.4 \\
    Monkey~\cite{li2024monkey} & 35.2 & 0 & 0 & 16.6 & 16.3 & 14.4 & 59.8 & 42.3 & 23.1 \\
    TextMonkey~\cite{liu2024textmonkey} & 39.1 & 0.7 & 0 & 19.0 & 12.2 & 19.0 & 61.1 & 40.2 & 23.9 \\
    Molmo-7B~\cite{deitke2024molmo} & 52.4 & 21.3 & 0.1 & 45.5 & 7.6 & 28.5 & 65.3 & 55.0 & 34.5 \\
    Cambrian-1-8B~\cite{tongcambrian} & 45.3 & 21.5 & 0 & 53.6 & 19.2 & 19.5 & 63.5 & 55.5 & 34.7 \\
    Pixtral-12B~\cite{agrawal2024pixtral} & 48.9 & 21.6 & 0 & 66.3 & 35.5 & 29.8 & 66.9 & 53.7 & 40.3 \\
    Qwen2.5-VL-7B~\cite{qwen2.5-VL} & \underline{68.8} & 25.7 & 1.2 & \underline{80.2} & 30.4 & 38.2 & 73.2 & 56.2 & 46.7 \\
    InternVL3-14B~\cite{chen2024expanding} & 67.3 & \underline{36.9} & \underline{11.2} & \textbf{89.0} & 38.4 & 38.4 & \textbf{79.2} & 60.5 & \textbf{52.6} \\
    Deepseek-VL2-Small~\cite{wu2024deepseek} & 62.7 & 28.0 & 0.1 & 77.5 & 32.7 & 14.3 & \underline{77.1} & 53.9 & 43.3 \\
    MiniCPM-o-2.6~\cite{yao2024minicpm} & 66.9 & 29.5 & 0.5 & 70.8 & 33.4 & 31.9 & 69.9 & 57.9 & 45.1 \\
    GLM-4V-9B~\cite{glm2024chatglm} & 61.8 & 22.6 & 0 & 71.7 & 31.6 & 22.6 & 72.1 & 58.4 & 42.6 \\
    Ovis2-8B~\cite{lu2024ovis} & \textbf{73.2} & 24.6 & 0.7 & 62.4 & \textbf{44.8} & 40.6 & 72.7 & \textbf{62.6} & 47.7 \\
    \rowcolor{gray!20}
    \multicolumn{10}{c}{Closed-source LMMs} \\
    GPT-4o~\cite{achiam2023gpt} & 61.2 & 26.7 & 0 & 77.5 & 36.3 & \underline{43.4} & 71.1 & 55.5 & 46.5 \\
    GPT-4o-mini~\cite{gpt4omini} & 57.9 & 23.3 & 0.6 & 70.8 & 31.5 & 38.8 & 65.9 & 55.1 & 43.0 \\
    Gemini-Pro~\cite{team2023gemini} & 61.2 & \textbf{39.5} & \textbf{13.5} & 79.3 & \underline{39.2} & \textbf{47.7} & 75.5 & 59.3 & \underline{51.9} \\
    Claude3.5-sonnet~\cite{Anthropic2024Claude3.5Sonnet} & 62.2 & 28.4 & 1.3 & 56.6 & 37.8 & 40.8 & 73.5 &  \underline{60.9} & 45.2 \\
    Step-1V~\cite{website_step1v} & 67.8 & 31.3 & 7.2 & 73.6 & 37.2 & 27.8 & 69.8 & 58.6 & 46.7 \\
    \bottomrule
  \end{tabular}
  }
  \caption{\textbf{Evaluation of existing LMMs on English tasks of OCRBench v2's public data}. ``Recognition'', ``Referring'', ``Spotting'', ``Extraction'', ``Parsing'', ``Calculation'', ``Understanding'', and ``Reasoning'' refer to text recognition, text referring, text spotting, relation extraction, element parsing, mathematical calculation, visual text understanding, and knowledge reasoning, respectively. Higher values indicate better performance. Best performance is in boldface, and the second best is underlined. The notations apply to all subsequent figures.}
  \label{tab:English_subsets}
  \vspace{-5pt}
\end{table*}

\subsection{Evaluation Criteria}
\label{sec:3_3_evaluation_method}

We adopt six types of evaluation metrics tailored to specific task categories. In the following, we present an overview of the evaluation metrics and their applicability to specific tasks.

\noindent\textbf{Parsing Type.} To evaluate the element parsing ability of LMMs, we assess their performance in transforming input images into structured formats, including HTML, Markdown, and JSON. TEDS~\cite{zhong2020image} is employed to measure the structural similarity between outputs and the desired format.

\noindent\textbf{Localization Type.} For text referring, the IoU score is applied to quantify the distance between the predicted regions and the ground truth.

\noindent\textbf{Extraction Type.} To evaluate relation extraction, we employ the F1 score to assess key information extraction and mapping. Since this evaluation requires structural extraction of information from the output of LMMs, the format is provided in the given prompt.    

\noindent\textbf{Long Reading Type.} To assess performance on long text reading tasks, BLEU~\cite{papineni2002bleu}, METEOR~\cite{banerjee2005meteor}, F1 score, and edit distance are used to assess the similarity between predicted text and ground truth.

\noindent\textbf{Counting Type.} In text counting, LMMs are required to count the number of text instances. Thus, we use the L1 distance to measure the absolute difference between predicted and ground truth counts. The final score is then normalized to the range of $[0,1]$ based on the ground truth.

\noindent\textbf{Basic VQA Type.} For questions where the original data provides options, we use exact string matching to compute accuracy. In other cases, we follow the approach of OCRBench to check whether the ground truth is contained in the prediction for short answers (fewer than $5$ words) and employ ANLS to measure prediction quality for longer answers ($5$ words or more).

\section{Results and Findings}
\label{Experiments}

\begin{table*}
  \centering
  \renewcommand{\arraystretch}{0.85} 
  \addtolength{\tabcolsep}{-0.5pt} 
  \fontsize{9}{11}\selectfont
  \resizebox{\textwidth}{!}{
  \begin{tabular}{
                  @{}
                  l
                  >{\centering\arraybackslash}p{1.5cm}
                  >{\centering\arraybackslash}p{1.5cm}
                  >{\centering\arraybackslash}p{1.25cm}
                  >{\centering\arraybackslash}p{1.25cm}
                  >{\centering\arraybackslash}p{1.25cm}
                  >{\centering\arraybackslash}p{1.25cm}
                  >{\centering\arraybackslash}p{1.25cm}
                  >{\centering\arraybackslash}p{1.25cm}
                  @{}
                  }
    \toprule
    Method & 
    \multicolumn{1}{>{\centering\arraybackslash}p{1.5cm}}{\fontsize{8}{9}\selectfont LLM Size} & 
    \multicolumn{1}{>{\centering\arraybackslash}p{1.25cm}}{\fontsize{8}{9}\selectfont Recognition} & 
    \multicolumn{1}{>{\centering\arraybackslash}p{1.25cm}}{\fontsize{8}{9}\selectfont Extraction} & 
    \multicolumn{1}{>{\centering\arraybackslash}p{1.25cm}}{\fontsize{8}{9}\selectfont Parsing} & 
    \multicolumn{1}{>{\centering\arraybackslash}p{1.25cm}}{\fontsize{8}{9}\selectfont Understanding} & 
    \multicolumn{1}{>{\centering\arraybackslash}p{1.25cm}}{\fontsize{8}{9}\selectfont Reasoning} & 
    \multicolumn{1}{>{\centering\arraybackslash}p{1.25cm}}{\fontsize{8}{9}\selectfont Average} \\
    \midrule
    \rowcolor{gray!20}
    \multicolumn{8}{c}{Open-source LMMs} \\
    LLaVA-Next-8B~\cite{liu2024llava} & 8B & 5.7 & 2.9 & 12.2 & 7.5 & 17.2 & 9.1 \\
    LLaVA-OV-7B~\cite{li2024llava} & 8B & 14.8 & 15.7 & 13.7 & 16.0 & 28.7 & 17.8 \\
    Monkey~\cite{li2024monkey} & 8B & 4.6 & 11.2 & 8.4 & 21.5 & 20.0 & 13.1 \\
    TextMonkey~\cite{liu2024textmonkey} & 8B &  23.5 & 14.8 & 8.4 & 19.9 & 12.2 & 15.8 \\
    Molmo-7B~\cite{deitke2024molmo} & 8B & 7.1 & 15.0 & 9.2 & 9.0 & 23.7 & 12.8 \\
    Cambrian-1-8B~\cite{tongcambrian} & 8B & 5.3 & 14.9 & 12.6 & 8.5 & 8.1 & 9.9 \\
    Pixtral-12B~\cite{agrawal2024pixtral} & 12B & 13.4 & 10.9 & 21.0 & 7.0 & 20.7 & 14.6 \\
    Qwen2.5-VL-7B~\cite{qwen2.5-VL} & 8B & \textbf{75.3} & \underline{61.4} & \textbf{41.8} & \underline{59.3} & \underline{40.4} & \underline{55.6} \\
    InternVL3-14B~\cite{chen2024expanding} & 14B & 66.2 & \textbf{64.8} & 33.5 & \textbf{63.4} & \textbf{50.6} & \textbf{55.7} \\
    Deepseek-VL2-Small~\cite{wu2024deepseek} & 16B & 60.9 & 50.6 & 28.3 & 53.0 & 20.5 & 42.7 \\
    MiniCPM-o-2.6~\cite{yao2024minicpm} & 7B & 53.0 & 49.4 & 27.1 & 43.5 & 32.7 & 41.1 \\
    GLM-4V-9B~\cite{glm2024chatglm} & 9B & 24.4 & 60.6 & 20.4 & 52.8 & 25.2 & 36.6 \\
    Ovis2-8B~\cite{lu2024ovis} & 7B & \underline{72.2} & 50.8 & \underline{37.7} & 47.9 & 37.4 & 49.2 \\
    \rowcolor{gray!20}
    \multicolumn{8}{c}{Closed-source LMMs} \\
    GPT-4o~\cite{achiam2023gpt} & - & 21.6 & 53.0 & 29.8 & 38.5 & 18.2 & 32.2 \\
    GPT-4o-mini~\cite{gpt4omini} & - & 13.1 & 38.9 & 27.2 & 28.8 & 16.9 & 25.0 \\
    Gemini-Pro~\cite{team2023gemini} & - & 52.5 & 47.3 & 30.9 & 51.5 & 33.4 & 43.1 \\
    Claude3.5-sonnet~\cite{Anthropic2024Claude3.5Sonnet} & - & 21.0 & 56.2 & 35.2 & 55.0 & 30.5 & 39.6 \\
    Step-1V~\cite{website_step1v} & - & 56.7 & 41.1 & 37.6 & 38.3 & 39.2 & 42.6 \\
    \bottomrule
  \end{tabular}
  }
  \caption{\textbf{Evaluation of existing LMMs on Chinese tasks of OCRBench v2' public data}. ``LLM Size'' indicates the number of parameters of the language model employed in each method.}
  \label{tab:Chinese_subsets}
  \vspace{-5pt}
\end{table*}

Here we first benchmark state-of-the-art LMMs on \textit{\datasetname}, presenting the quantitative analysis, then summarize key findings of current limitations for LMMs. All results are presented as percentages.

\subsection{Baselines}
The tested LMMs in the section includes LLaVA-Next-8B~\cite{liu2024llava}, LLaVA-OV-7B~\cite{li2024llava}, Monkey~\cite{li2024monkey}, TextMonkey~\cite{liu2024textmonkey}, Molmo-7B~\cite{deitke2024molmo}, Cambrian-1-8B~\cite{tongcambrian}, Pixtral-12B~\cite{agrawal2024pixtral}, Qwen2.5-VL-7B~\cite{qwen2.5-VL}, InternVL3-14B~\cite{chen2024expanding}, Deepseek-VL2-Tiny~\cite{wu2024deepseek}, MiniCPM-o-2.6~\cite{yao2024minicpm}, GLM-4v-9B~\cite{glm2024chatglm}, Ovis2-8B~\cite{lu2024ovis}, GPT4o~\cite{openai2024gpt4o}, GPT4o-mini~\cite{gpt4omini}, Gemini-1.5-Pro~\cite{team2023gemini}, Claude3.5-sonnet~\cite{Anthropic2024Claude3.5Sonnet}, and Step-1V~\cite{website_step1v}. Due to space limitations, more LMM evaluation results can be found in the Sec.~\ref{sec:Evaluation results}.

\subsection{Main Results}
\label{main_results}

\noindent\textbf{Evaluation results on public data} are shown in Tab.~\ref{tab:English_subsets} and Tab.~\ref{tab:Chinese_subsets}. 
While LMMs perform well on some basic capabilities such as text recognition and visual text understanding, most LMMs achieve low scores in other capabilities, such as text spotting and element parsing, mostly below $50$. In particular, some LMMs show significant limitations in text spotting capabilities, failing to precisely locate and recognize the texts. Additionally, LMMs demonstrate inadequate abilities in element parsing and mathematical calculation, which are crucial for complicated tasks like document analysis and mathematical reasoning. Besides, after comparing the performance of LMMs on visual text understanding and knowledge reasoning capabilities, we find that they perform poorly in knowledge reasoning. This suggests the deficiency of LMMs in logical reasoning. 

\noindent\textbf{Evaluation results on private data} are shown in Tab.~\ref{tab:private_English_subsets} and Tab.~\ref{tab:private_Chinese_subsets}. We observe similar evaluation trends to those in the public test set experiments. Overall, LMMs exhibit unsatisfactory performance in text referring, text spotting, element parsing, mathematical calculation, and knowledge reasoning capabilities. In addition, closed-source LMMs outperform their open-source counterparts, demonstrating stronger generalization capabilities. The consistent results across both public and private test sets confirm the soundness of \textit{\datasetname}'s task design, data collection process, and evaluation metrics, and demonstrate its effectiveness in revealing the capability limitations of current LMMs.

\begin{table*}
  \centering
  \renewcommand{\arraystretch}{0.9} 
  \addtolength{\tabcolsep}{-2pt} 
  \fontsize{9}{11}\selectfont
  \resizebox{\textwidth}{!}{
  \begin{tabular}{ 
                  @{}
                  l
                  >{\centering\arraybackslash}p{1.25cm}
                  >{\centering\arraybackslash}p{1.25cm}
                  >{\centering\arraybackslash}p{1.25cm}
                  >{\centering\arraybackslash}p{1.25cm}
                  >{\centering\arraybackslash}p{1.25cm}
                  >{\centering\arraybackslash}p{1.25cm}
                  >{\centering\arraybackslash}p{1.25cm}
                  >{\centering\arraybackslash}p{1.25cm}
                  >{\centering\arraybackslash}p{1.25cm}
                  @{}
                  }
    \toprule
    {\fontsize{7.5}{9}\selectfont Method} & 
    \multicolumn{1}{>{\centering\arraybackslash}p{1.25cm}}{\fontsize{7.5}{9}\selectfont Recognition} & 
    \multicolumn{1}{>{\centering\arraybackslash}p{1.25cm}}{\fontsize{7.5}{9}\selectfont Referring} & 
    \multicolumn{1}{>{\centering\arraybackslash}p{1.25cm}}{\fontsize{7.5}{9}\selectfont Spotting} &{\fontsize{8}{9}\selectfont Extraction} & 
    \multicolumn{1}{>{\centering\arraybackslash}p{1.25cm}}{\fontsize{7.5}{9}\selectfont Parsing} & 
    \multicolumn{1}{>{\centering\arraybackslash}p{1.25cm}}{\fontsize{7.5}{9}\selectfont Calculation} & 
    \multicolumn{1}{>{\centering\arraybackslash}p{1.25cm}}{\fontsize{7.5}{9}\selectfont Understanding} & 
    \multicolumn{1}{>{\centering\arraybackslash}p{1.25cm}}{\fontsize{7.5}{9}\selectfont Reasoning} & 
    \multicolumn{1}{>{\centering\arraybackslash}p{1.25cm}}{\fontsize{7.5}{9}\selectfont Average} \\
    \midrule
    \rowcolor{gray!20}
    \multicolumn{10}{c}{Open-source LMMs} \\
    LLaVA-Next-8B~\cite{liu2024llava} & 41.4 & 17.0 & 0 & 49.0 & 12.9 & 16.1 & 60.9 & 30.5 & 28.5 \\
    LLaVA-OV-7B~\cite{li2024llava} & 45.4 & 18.5 & 0 & 60.0 & 15.5 & 32.0 & 59.0 & 39.3 & 33.7 \\
    Monkey~\cite{li2024monkey} & 31.5 & 0.1 & 0 & 34.4 & \underline{26.3} & 17.7 & 61.4 & 22.4 & 24.2 \\
    TextMonkey~\cite{liu2024textmonkey} & 39.8 & 1.6 & 0 & 27.6 & 24.8 & 10.2 & 62.3 & 21.2 & 23.4 \\
    Molmo-7B~\cite{deitke2024molmo} & 40.8 & 19.5 & 0 & 51.7 & 10.0 & 33.9 & 67.0 & 48.0 & 33.9 \\
    Cambrian-1-8B~\cite{tongcambrian} & 44.0 & 19.0 & 0 & 52.3 & 19.0 & 20.7 & 64.0 & 39.3 & 32.3 \\
    Pixtral-12B~\cite{agrawal2024pixtral} & 45.1 & 21.8 & 0 & 71.6 & 21.7 & 30.4 & 77.3 & 39.5 & 38.4 \\
    Qwen2.5-VL-7B~\cite{wang2024qwen2} & 51.5 & 24.5 & \underline{3.1} & 64.8 & 13.1 & 53.3 & \underline{78.6} & 45.5 & 41.8 \\
    InternVL3-14B~\cite{chen2024expanding} & 55.8 & 24.5 & 2.1 & \underline{89.3} & 21.0 & \underline{59.5} & 72.0 & 50.0 & 46.8  \\
    Deepseek-VL2-Small~\cite{wu2024deepseek} & 56.6 & 23.7 & 0 & 86.4 & 18.9 & 30.6 & 72.2 & 39.5 & 41.0 \\
    MiniCPM-o-2.6~\cite{yao2024minicpm} & 54.1 & 24.7 & 0.3 & 74.4 & 17.6 & 39.2 & 75.7 & 47.0 & 41.6 \\
    GLM-4v-9B~\cite{glm2024chatglm} & 52.7 & 20.6 & 0 & 79.4 & 15.9 & 21.5 & 74.7 & 32.0 & 37.1 \\
    Ovis2-8B~\cite{lu2024ovis} & 54.2 & 20.9 & 0 & 83.6 & 24.2 & 54.7 & 74.1 & 57.3 & 46.1 \\
    \rowcolor{gray!20}
    \multicolumn{10}{c}{Closed-source LMMs} \\
    GPT-4o~\cite{achiam2023gpt} & \underline{58.6} & 23.4 & 0 & 87.4 & 23.1 & 51.6 & 74.4 & \textbf{62.3} & \underline{47.6} \\
    GPT-4o-mini~\cite{gpt4omini} & 55.3 & 21.8 & 0 & 85.4 & 20.6 & 45.2 & 75.5 & 49.0 & 44.1 \\
    Gemini1.5-Pro~\cite{team2023gemini} & \textbf{59.1} & \textbf{41.2} & \textbf{6.6} & \textbf{89.5} & 22.4 & 54.7 & \textbf{78.8} & \underline{60.3} & \textbf{51.6} \\
    Claude3.5-sonnet~\cite{Anthropic2024Claude3.5Sonnet} & 52.9 & 24.9 & 2.5 & 86.9 & 23.8 & \textbf{61.4} & 74.4 & 53.0 & 47.5 \\
    Step-1V~\cite{website_step1v} & 56.7 & \underline{27.4} & 2.6 & 86.3 & \textbf{33.3} & 42.6 & 76.6 & 48.7 & 46.8 \\
    \bottomrule
  \end{tabular}
  }
  \caption{\textbf{Evaluation of existing LMMs on English tasks of OCRBench v2's private data}.}
  \label{tab:private_English_subsets}
\end{table*}

\begin{table*}
  \centering
  \renewcommand{\arraystretch}{0.85} 
  \addtolength{\tabcolsep}{-0.5pt} 
  \fontsize{9}{11}\selectfont
  \resizebox{\textwidth}{!}{
  \begin{tabular}{
                  @{}
                  l
                  >{\centering\arraybackslash}p{1.5cm}
                  >{\centering\arraybackslash}p{1.5cm}
                  >{\centering\arraybackslash}p{1.25cm}
                  >{\centering\arraybackslash}p{1.25cm}
                  >{\centering\arraybackslash}p{1.25cm}
                  >{\centering\arraybackslash}p{1.25cm}
                  >{\centering\arraybackslash}p{1.25cm}
                  >{\centering\arraybackslash}p{1.25cm}
                  @{}
                  }
    \toprule
    Method & 
    \multicolumn{1}{>{\centering\arraybackslash}p{1.5cm}}{\fontsize{8}{9}\selectfont LLM Size} & 
    \multicolumn{1}{>{\centering\arraybackslash}p{1.25cm}}{\fontsize{8}{9}\selectfont Recognition} & 
    \multicolumn{1}{>{\centering\arraybackslash}p{1.25cm}}{\fontsize{8}{9}\selectfont Extraction} & 
    \multicolumn{1}{>{\centering\arraybackslash}p{1.25cm}}{\fontsize{8}{9}\selectfont Parsing} & 
    \multicolumn{1}{>{\centering\arraybackslash}p{1.25cm}}{\fontsize{8}{9}\selectfont Understanding} & 
    \multicolumn{1}{>{\centering\arraybackslash}p{1.25cm}}{\fontsize{8}{9}\selectfont Reasoning} & 
    \multicolumn{1}{>{\centering\arraybackslash}p{1.25cm}}{\fontsize{8}{9}\selectfont Average} \\
    \midrule
    \rowcolor{gray!20}
    \multicolumn{8}{c}{Open-source LMMs} \\
    LLaVA-Next-8B~\cite{liu2024llava} & 8B & 2.8 & 0.9 & 14.9 & 20.0 & 7.4 & 9.2 \\
    LLaVA-OV-7B~\cite{li2024llava} & 8B & 5.4 & 13.6 & 20.3 & 34.0 & 13.6 & 17.4 \\
    Monkey~\cite{li2024monkey} & 8B & 1.5 & 28.4 & 29.1 & 40.0 & 8.3 & 21.5 \\
    TextMonkey~\cite{liu2024textmonkey} & 8B & 10.5 & 15.2 & 30.2 & 44.0 & 7.6 & 21.5 \\
    Molmo-7B~\cite{deitke2024molmo} & 8B & 3.4 & 29.8 & 6.6 & 24.0 & 11.1 & 15.0 \\
    Cambrian-1-8B~\cite{tongcambrian} & 8B & 2.4 & 19.8 & 26.7 & 36.0 & 7.6 & 18.5 \\
    Pixtral-12B~\cite{agrawal2024pixtral} & 12B & 6.2 & 22.3 & 11.4 & 26.0 & 14.0 & 16.0 \\
    Qwen2.5-VL-7B~\cite{wang2024qwen2} & 8B & 24.4 & \textbf{78.9} & 33.1 & \textbf{82.0} & 29.0 & 49.5 \\
    InternVL3-14B~\cite{chen2024expanding} & 14B & 62.1 & 59.5 & 33.2 & 80.0 & 29.2 & 52.8 \\
    DeepSeek-VL2-Small~\cite{wu2024deepseek} & 16B & 51.6 & 56.3 & 27.8 & 79.6 & 25.3 & 48.1 \\
    MiniCPM-o-2.6~\cite{yao2024minicpm} & 7B & 54.0 & 62.4 & 24.1 & 68.0 & 29.8 & 47.7 \\
    GLM-4v-9B~\cite{glm2024chatglm} & 9B & 60.6 & 65.2 & 32.4 & \textbf{82.0} & 18.2 & 51.7 \\
    Ovis2-8B~\cite{lu2024ovis} & 7B & 61.0 & \underline{67.7} & \textbf{43.6} & \textbf{82.0} & 25.6 & \textbf{56.0} \\
    \rowcolor{gray!20}
    \multicolumn{8}{c}{Closed-source LMMs} \\
    GPT-4o~\cite{achiam2023gpt} & - & 41.7 & 52.1 & 29.0 & 76.0 & 29.4 & 45.7 \\
    GPT-4o-mini~\cite{gpt4omini} & - & 20.0 & 53.6 & 27.9 & 66.0 & 19.6 & 37.4 \\
    Gemini1.5-Pro~\cite{team2023gemini} & - & \textbf{71.4} & 63.8 & 30.5 & \textbf{82.0} & \underline{29.9} & \underline{55.5} \\
    Claude3.5-sonnet~\cite{Anthropic2024Claude3.5Sonnet} & - & 34.2 & 62.5 & \underline{35.2} & 78.0 & \textbf{32.2} & 48.4 \\
    Step-1V~\cite{website_step1v} & - & \underline{65.2} & 64.9 & 33.1 & 78.0 & 25.5 & 53.4 \\
    \bottomrule
  \end{tabular}
  }
  \caption{\textbf{Evaluation of existing LMMs on Chinese tasks of OCRBench v2's private data}.}
  \label{tab:private_Chinese_subsets}
\end{table*}

\subsection{Main Findings}
We provide in-depth analyses for LMMs' common limitations, including rare text recognition, fine-grained spatial perception, layout perception, complex element analysis, and logical reasoning.

\noindent\textbf{Finding 1.} LMMs still face challenges with less frequently encountered texts, such as dot matrix texts and mathematical formulas. This performance gap highlights the continuing challenges LMMs face in real-world text recognition. For instance, occluded text, CAPTCHA, and dot-matrix text are considered low-frequency text, whereas other types belong to high-frequency text. Notably, recognition accuracy varies significantly across these categories. For example, InternVL3-14B achieves 79.1\% accuracy on high-frequency texts but drops to 46.7\% on low-frequency ones.

\noindent\textbf{Finding 2.} Current LMMs still exhibit limited performance in tasks requiring precise spatial understanding, such as text referring and text spotting. For instance, while InternVL3-14B achieves a response accuracy of 78.3\% on the VQA with position task, its IoU score for answer region localization is only 12.9\%. This suggests that although LMMs can roughly identify where the answer is located, they struggle to output the exact region.

\noindent\textbf{Finding 3.} While LMMs achieve good performance on basic text recognition, they struggle with complex layouts such as overlapping or rotated texts. For example, GPT-4o fails to detect the characters in overlapping handwritten text and misrecognizes numbers in 90° rotated images, revealing LMMs' limitations in handling texts with complex layouts. Rotating images in the DocVQA dataset led to a significant performance drop of $55.7\%$ for InternVL3-14B (from $90.9\%$ to $35.2\%$).

\noindent\textbf{Finding 4.} LMMs still struggle to parse text into structured formats in downstream applications such as document digitalization. For instance, InternVL3-14B achieves 94.4\% accuracy in unpaired entities matching, but its performance drops to 84.9\% in key information extraction, where the model is required to identify the corresponding value given an entity. The performance further degrades in element parsing tasks that demand structured outputs. 

\noindent\textbf{Finding 5.} Despite recent advances, LMMs still face challenges in complex mathematical and textual reasoning tasks. To assess their capabilities, we evaluated InternVL3-14B on the private test set covering reasoning VQA, ScienceQA, and APP agent tasks. Questions were categorized into five types: common sense reasoning, visual-text understanding, pattern recognition, calculation, and expert knowledge. Human ratings showed the model achieved accuracies of $72.9\%$, $83.0\%$, $69.2\%$, $56.5\%$, and $71.8\%$, respectively, indicating notable variation. 

\section{Conclusion}
\label{Conclusion}
In this work, we introduce \textit{\datasetname}, a comprehensive benchmark designed to evaluate the OCR capabilities of LMMs. Covering $23$ tasks across $31$ diverse scenarios, our benchmark systematically assesses eight core capabilities that are essential for text-oriented visual understanding tasks. It includes $10,000$ high-quality QA pairs and six rigorous evaluation metrics. In addition, we curate a private test set of $1,500$ manually labeled images to ensure robust generalization evaluation. Leveraging this benchmark, we conduct extensive experiments on representative LMMs. Through in-depth analysis of experimental results, we identify critical limitations of current models and uncover key factors that affect their OCR performance. We hope \textit{\datasetname} could aid future research on enhancing LMMs' text understanding ability. 


\bibliographystyle{IEEEtran}     
\bibliography{references.bib}  

\begin{thebibliography}{100}
\providecommand{\url}[1]{#1}
\csname url@samestyle\endcsname
\providecommand{\newblock}{\relax}
\providecommand{\bibinfo}[2]{#2}
\providecommand{\BIBentrySTDinterwordspacing}{\spaceskip=0pt\relax}
\providecommand{\BIBentryALTinterwordstretchfactor}{4}
\providecommand{\BIBentryALTinterwordspacing}{\spaceskip=\fontdimen2\font plus
\BIBentryALTinterwordstretchfactor\fontdimen3\font minus \fontdimen4\font\relax}
\providecommand{\BIBforeignlanguage}[2]{{%
\expandafter\ifx\csname l@#1\endcsname\relax
\typeout{** WARNING: IEEEtran.bst: No hyphenation pattern has been}%
\typeout{** loaded for the language `#1'. Using the pattern for}%
\typeout{** the default language instead.}%
\else
\language=\csname l@#1\endcsname
\fi
#2}}
\providecommand{\BIBdecl}{\relax}
\BIBdecl

\bibitem{achiam2023gpt}
J.~Achiam, S.~Adler, S.~Agarwal, L.~Ahmad, I.~Akkaya, F.~L. Aleman, D.~Almeida, J.~Altenschmidt, S.~Altman, S.~Anadkat \emph{et~al.}, ``{Gpt-4 technical report},'' \emph{arXiv preprint arXiv:2303.08774}, 2023.

\bibitem{touvron2023llama}
H.~Touvron, T.~Lavril, G.~Izacard, X.~Martinet, M.-A. Lachaux, T.~Lacroix, B.~Rozi{\`e}re, N.~Goyal, E.~Hambro, F.~Azhar \emph{et~al.}, ``{Llama: Open and efficient foundation language models},'' \emph{arXiv preprint arXiv:2302.13971}, 2023.

\bibitem{brown2020language}
T.~Brown, B.~Mann, N.~Ryder, M.~Subbiah, J.~D. Kaplan, P.~Dhariwal, A.~Neelakantan, P.~Shyam, G.~Sastry, A.~Askell \emph{et~al.}, ``{Language models are few-shot learners},'' \emph{Advances in Neural Information Processing Systems}, 2020.

\bibitem{bai2023qwen}
J.~Bai, S.~Bai, S.~Yang, S.~Wang, S.~Tan, P.~Wang, J.~Lin, C.~Zhou, and J.~Zhou, ``{Qwen-vl: A frontier large vision-language model with versatile abilities},'' \emph{arXiv preprint arXiv:2308.12966}, 2023.

\bibitem{liu2024visual}
H.~Liu, C.~Li, Q.~Wu, and Y.~J. Lee, ``{Visual instruction tuning},'' \emph{Advances in Neural Information Processing Systems}, vol.~36, 2024.

\bibitem{zhu2023minigpt}
D.~Zhu, J.~Chen, X.~Shen, X.~Li, and M.~Elhoseiny, ``{Minigpt-4: Enhancing vision-language understanding with advanced large language models},'' \emph{Proceedings of the International Conference on Learning Representations}, 2024.

\bibitem{liu2024textmonkey}
Y.~Liu, B.~Yang, Q.~Liu, Z.~Li, Z.~Ma, S.~Zhang, and X.~Bai, ``{Textmonkey: An ocr-free large multimodal model for understanding document},'' \emph{arXiv preprint arXiv:2403.04473}, 2024.

\bibitem{fu2023mme}
C.~Fu, P.~Chen, Y.~Shen, Y.~Qin, M.~Zhang, X.~Lin, J.~Yang, X.~Zheng, K.~Li, X.~Sun \emph{et~al.}, ``{MME}: A comprehensive evaluation benchmark for multimodal large language models,'' \emph{arXiv preprint arXiv:2306.13394}, 2023.

\bibitem{ying2024mmt}
K.~Ying, F.~Meng, J.~Wang, Z.~Li, H.~Lin, Y.~Yang, H.~Zhang, W.~Zhang, Y.~Lin, S.~Liu \emph{et~al.}, ``{Mmt-bench: A comprehensive multimodal benchmark for evaluating large vision-language models towards multitask agi},'' \emph{arXiv preprint arXiv:2404.16006}, 2024.

\bibitem{singh2019towards}
A.~Singh, V.~Natarajan, M.~Shah, Y.~Jiang, X.~Chen, D.~Batra, D.~Parikh, and M.~Rohrbach, ``{Towards vqa models that can read},'' in \emph{Proceedings of the IEEE/CVF Conference on Computer Vision and Pattern Recognition}, 2019, pp. 8317--8326.

\bibitem{biten2019scene}
A.~F. Biten, R.~Tito, A.~Mafla, L.~Gomez, M.~Rusinol, E.~Valveny, C.~Jawahar, and D.~Karatzas, ``{Scene text visual question answering},'' in \emph{Proceedings of the IEEE/CVF Conference on Computer Vision and Pattern Recognition}, 2019, pp. 4291--4301.

\bibitem{wang2020general}
X.~Wang, Y.~Liu, C.~Shen, C.~C. Ng, C.~Luo, L.~Jin, C.~S. Chan, A.~v.~d. Hengel, and L.~Wang, ``{On the general value of evidence, and bilingual scene-text visual question answering},'' in \emph{Proceedings of the IEEE/CVF Conference on Computer Vision and Pattern Recognition}, 2020, pp. 10\,126--10\,135.

\bibitem{chen2024we}
L.~Chen, J.~Li, X.~Dong, P.~Zhang, Y.~Zang, Z.~Chen, H.~Duan, J.~Wang, Y.~Qiao, D.~Lin \emph{et~al.}, ``{Are We on the Right Way for Evaluating Large Vision-Language Models?}'' \emph{arXiv preprint arXiv:2403.20330}, 2024.

\bibitem{liu2023hidden}
Y.~Liu, Z.~Li, B.~Yang, C.~Li, X.~Yin, C.-l. Liu, L.~Jin, and X.~Bai, ``{On the hidden mystery of ocr in large multimodal models},'' \emph{arXiv preprint arXiv:2305.07895}, 2023.

\bibitem{li2024seed}
B.~Li, Y.~Ge, Y.~Chen, Y.~Ge, R.~Zhang, and Y.~Shan, ``{Seed-bench-2-plus: Benchmarking multimodal large language models with text-rich visual comprehension},'' \emph{arXiv preprint arXiv:2404.16790}, 2024.

\bibitem{wadhawancontextual}
R.~Wadhawan, H.~Bansal, K.-W. Chang, and N.~Peng, ``{ConTextual: Evaluating Context-Sensitive Text-Rich Visual Reasoning in Large Multimodal Models},'' in \emph{Proceedings of International Conference on Machine Learning}, 2024.

\bibitem{liu2024focus}
C.~Liu, H.~Wei, J.~Chen, L.~Kong, Z.~Ge, Z.~Zhu, L.~Zhao, J.~Sun, C.~Han, and X.~Zhang, ``{Focus Anywhere for Fine-grained Multi-page Document Understanding},'' \emph{arXiv preprint arXiv:2405.14295}, 2024.

\bibitem{kim2024tablevqa}
Y.~Kim, M.~Yim, and K.~Y. Song, ``{TableVQA-Bench}: A visual question answering benchmark on multiple table domains,'' \emph{arXiv preprint arXiv:2404.19205}, 2024.

\bibitem{zhao2024tabpedia}
W.~Zhao, H.~Feng, Q.~Liu, J.~Tang, S.~Wei, B.~Wu, L.~Liao, Y.~Ye, H.~Liu, H.~Li \emph{et~al.}, ``{TabPedia: Towards Comprehensive Visual Table Understanding with Concept Synergy},'' \emph{arXiv preprint arXiv:2406.01326}, 2024.

\bibitem{xia2024chartx}
R.~Xia, B.~Zhang, H.~Ye, X.~Yan, Q.~Liu, H.~Zhou, Z.~Chen, M.~Dou, B.~Shi, J.~Yan \emph{et~al.}, ``{Chartx \& chartvlm: A versatile benchmark and foundation model for complicated chart reasoning},'' \emph{arXiv preprint arXiv:2402.12185}, 2024.

\bibitem{qwen2.5-VL}
\BIBentryALTinterwordspacing
Q.~Team, ``Qwen2.5-vl,'' January 2025. [Online]. Available: \url{https://qwenlm.github.io/blog/qwen2.5-vl/}
\BIBentrySTDinterwordspacing

\bibitem{mathew2021docvqa}
M.~Mathew, D.~Karatzas, and C.~Jawahar, ``{Docvqa: A dataset for vqa on document images},'' in \emph{Proceedings of the IEEE Winter Conference on Applications of Computer Vision}, 2021, pp. 2200--2209.

\bibitem{chen2024expanding}
Z.~Chen, W.~Wang, Y.~Cao, Y.~Liu, Z.~Gao, E.~Cui, J.~Zhu, S.~Ye, H.~Tian, Z.~Liu \emph{et~al.}, ``Expanding performance boundaries of open-source multimodal models with model, data, and test-time scaling,'' \emph{arXiv preprint arXiv:2412.05271}, 2024.

\bibitem{openai2024gpt4o}
OpenAI, ``{Hello GPT-4o},'' \url{https://openai.com/index/gpt-4v-system-card}, 2024, accessed: 2024-12-29.

\bibitem{ouyang2024omnidocbench}
L.~Ouyang, Y.~Qu, H.~Zhou, J.~Zhu, R.~Zhang, Q.~Lin, B.~Wang, Z.~Zhao, M.~Jiang, X.~Zhao \emph{et~al.}, ``Omnidocbench: Benchmarking diverse pdf document parsing with comprehensive annotations,'' \emph{arXiv preprint arXiv:2412.07626}, 2024.

\bibitem{yang2024cc}
Z.~Yang, J.~Tang, Z.~Li, P.~Wang, J.~Wan, H.~Zhong, X.~Liu, M.~Yang, P.~Wang, Y.~Liu \emph{et~al.}, ``Cc-ocr: A comprehensive and challenging ocr benchmark for evaluating large multimodal models in literacy,'' \emph{arXiv preprint arXiv:2412.02210}, 2024.

\bibitem{ma2024mmlongbench}
Y.~Ma, Y.~Zang, L.~Chen, M.~Chen, Y.~Jiao, X.~Li, X.~Lu, Z.~Liu, Y.~Ma, X.~Dong \emph{et~al.}, ``Mmlongbench-doc: Benchmarking long-context document understanding with visualizations,'' \emph{arXiv preprint arXiv:2407.01523}, 2024.

\bibitem{DBLP:conf/acl/ZhengFSS0J024}
\BIBentryALTinterwordspacing
M.~Zheng, X.~Feng, Q.~Si, Q.~She, Z.~Lin, W.~Jiang, and W.~Wang, ``{Multimodal Table Understanding},'' in \emph{Proceedings of Annual Meeting of the Association for Computational Linguistics}, L.~Ku, A.~Martins, and V.~Srikumar, Eds.\hskip 1em plus 0.5em minus 0.4em\relax Association for Computational Linguistics, 2024, pp. 9102--9124. [Online]. Available: \url{https://doi.org/10.18653/v1/2024.acl-long.493}
\BIBentrySTDinterwordspacing

\bibitem{liu2024mmc}
F.~Liu, X.~Wang, W.~Yao, J.~Chen, K.~Song, S.~Cho, Y.~Yacoob, and D.~Yu, ``{MMC: Advancing Multimodal Chart Understanding with Large-scale Instruction Tuning},'' in \emph{Proceedings of the 2024 Conference of the North American Chapter of the Association for Computational Linguistics: Human Language Technologies}, 2024, pp. 1287--1310.

\bibitem{zhang2023llavar}
Y.~Zhang, R.~Zhang, J.~Gu, Y.~Zhou, N.~Lipka, D.~Yang, and T.~Sun, ``{Llavar: Enhanced visual instruction tuning for text-rich image understanding},'' \emph{arXiv preprint arXiv:2306.17107}, 2023.

\bibitem{ye2023mplug}
J.~Ye, A.~Hu, H.~Xu, Q.~Ye, M.~Yan, Y.~Dan, C.~Zhao, G.~Xu, C.~Li, J.~Tian \emph{et~al.}, ``{mplug-docowl: Modularized multimodal large language model for document understanding},'' \emph{arXiv preprint arXiv:2307.02499}, 2023.

\bibitem{feng2023docpedia}
H.~Feng, Q.~Liu, H.~Liu, W.~Zhou, H.~Li, and C.~Huang, ``{Docpedia: Unleashing the power of large multimodal model in the frequency domain for versatile document understanding},'' \emph{arXiv preprint arXiv:2311.11810}, 2023.

\bibitem{ye2023ureader}
J.~Ye, A.~Hu, H.~Xu, Q.~Ye, M.~Yan, G.~Xu, C.~Li, J.~Tian, Q.~Qian, J.~Zhang \emph{et~al.}, ``{Ureader: Universal ocr-free visually-situated language understanding with multimodal large language model},'' \emph{arXiv preprint arXiv:2310.05126}, 2023.

\bibitem{luo2024layoutllm}
C.~Luo, Y.~Shen, Z.~Zhu, Q.~Zheng, Z.~Yu, and C.~Yao, ``{LayoutLLM: Layout Instruction Tuning with Large Language Models for Document Understanding},'' in \emph{Proceedings of the IEEE/CVF Conference on Computer Vision and Pattern Recognition}, 2024, pp. 15\,630--15\,640.

\bibitem{hu2024mplug}
A.~Hu, H.~Xu, J.~Ye, M.~Yan, L.~Zhang, B.~Zhang, C.~Li, J.~Zhang, Q.~Jin, F.~Huang \emph{et~al.}, ``{mplug-docowl 1.5: Unified structure learning for ocr-free document understanding},'' \emph{arXiv preprint arXiv:2403.12895}, 2024.

\bibitem{zhang2025dockylin}
J.~Zhang, W.~Yang, S.~Lai, Z.~Xie, and L.~Jin, ``Dockylin: A large multimodal model for visual document understanding with efficient visual slimming,'' in \emph{Proceedings of the AAAI Conference on Artificial Intelligence}, vol.~39, no.~9, 2025, pp. 9923--9932.

\bibitem{liao2024doclayllm}
W.~Liao, J.~Wang, H.~Li, C.~Wang, J.~Huang, and L.~Jin, ``Doclayllm: An efficient and effective multi-modal extension of large language models for text-rich document understanding,'' \emph{arXiv preprint arXiv:2408.15045}, 2024.

\bibitem{zhu2025simple}
Z.~Zhu, C.~Luo, Z.~Shao, F.~Gao, H.~Xing, Q.~Zheng, and J.~Zhang, ``A simple yet effective layout token in large language models for document understanding,'' in \emph{Proceedings of the IEEE/CVF Conference on Computer Vision and Pattern Recognition}, 2025.

\bibitem{xiao2025docmark}
H.~Xiao, Y.~Xie, G.~Tan, Y.~Chen, R.~Hu, K.~Wang, A.~Zhou, H.~Li, H.~Shao, X.~Lu, P.~Gao, Y.~Wen, X.~Chen, S.~Ren, and H.~Li, ``Adaptive markup language generation for contextually-grounded visual document understanding,'' in \emph{Proceedings of the IEEE/CVF Conference on Computer Vision and Pattern Recognition}, 2025.

\bibitem{wang2025marten}
Z.~Wang, T.~Guan, P.~Fu, C.~Duan, Q.~Jiang, Z.~Guo, S.~Guo, J.~Luo, W.~Shen, and X.~Yang, ``Marten: Visual question answering with mask generation for multi-modal document understanding,'' in \emph{Proceedings of the IEEE/CVF Conference on Computer Vision and Pattern Recognition}, 2025.

\bibitem{masry2022chartqa}
A.~Masry, D.~X. Long, J.~Q. Tan, S.~Joty, and E.~Hoque, ``{Chartqa: A benchmark for question answering about charts with visual and logical reasoning},'' \emph{arXiv preprint arXiv:2203.10244}, 2022.

\bibitem{mathew2022infographicvqa}
M.~Mathew, V.~Bagal, R.~Tito, D.~Karatzas, E.~Valveny, and C.~Jawahar, ``{Infographicvqa},'' in \emph{Proceedings of the IEEE Winter Conference on Applications of Computer Vision}, 2022, pp. 1697--1706.

\bibitem{zhang2024exploring}
S.~Zhang, B.~Yang, Z.~Li, Z.~Ma, Y.~Liu, and X.~Bai, ``{Exploring the Capabilities of Large Multimodal Models on Dense Text},'' in \emph{Proceedings of International Conference on Document Analysis and Recognition}.\hskip 1em plus 0.5em minus 0.4em\relax Springer, 2024, pp. 281--298.

\bibitem{chen2024onechart}
J.~Chen, L.~Kong, H.~Wei, C.~Liu, Z.~Ge, L.~Zhao, J.~Sun, C.~Han, and X.~Zhang, ``{Onechart: Purify the chart structural extraction via one auxiliary token},'' in \emph{Proceedings of the ACM International Conference on Multimedia}, 2024, pp. 147--155.

\bibitem{van2023document}
J.~Van~Landeghem, R.~Tito, {\L}.~Borchmann, M.~Pietruszka, P.~Joziak, R.~Powalski, D.~Jurkiewicz, M.~Coustaty, B.~Anckaert, E.~Valveny \emph{et~al.}, ``Document understanding dataset and evaluation (dude),'' in \emph{Proceedings of the IEEE/CVF Conference on Computer Vision and Pattern Recognition}, 2023, pp. 19\,528--19\,540.

\bibitem{wang2025needle}
W.~Wang, S.~Zhang, Y.~Ren, Y.~Duan, T.~Li, S.~Liu, M.~Hu, Z.~Chen, K.~Zhang, L.~Lu \emph{et~al.}, ``Needle in a multimodal haystack,'' \emph{Advances in Neural Information Processing Systems}, vol.~37, pp. 20\,540--20\,565, 2025.

\bibitem{tito2023hierarchical}
R.~Tito, D.~Karatzas, and E.~Valveny, ``Hierarchical multimodal transformers for multipage docvqa,'' \emph{Pattern Recognition}, vol. 144, p. 109834, 2023.

\bibitem{deng2024longdocurl}
C.~Deng, J.~Yuan, P.~Bu, P.~Wang, Z.-Z. Li, J.~Xu, X.-H. Li, Y.~Gao, J.~Song, B.~Zheng \emph{et~al.}, ``Longdocurl: a comprehensive multimodal long document benchmark integrating understanding, reasoning, and locating,'' \emph{arXiv preprint arXiv:2412.18424}, 2024.

\bibitem{liu2024llava}
H.~Liu, C.~Li, Y.~Li, B.~Li, Y.~Zhang, S.~Shen, and Y.~J. Lee, ``{Llava-next: Improved reasoning, ocr, and world knowledge},'' 2024.

\bibitem{li2024llava}
B.~Li, Y.~Zhang, D.~Guo, R.~Zhang, F.~Li, H.~Zhang, K.~Zhang, Y.~Li, Z.~Liu, and C.~Li, ``{Llava-onevision: Easy visual task transfer},'' \emph{arXiv preprint arXiv:2408.03326}, 2024.

\bibitem{li2024monkey}
Z.~Li, B.~Yang, Q.~Liu, Z.~Ma, S.~Zhang, J.~Yang, Y.~Sun, Y.~Liu, and X.~Bai, ``{Monkey: Image resolution and text label are important things for large multi-modal models},'' in \emph{Proceedings of the IEEE/CVF Conference on Computer Vision and Pattern Recognition}, 2024, pp. 26\,763--26\,773.

\bibitem{deitke2024molmo}
M.~Deitke, C.~Clark, S.~Lee, R.~Tripathi, Y.~Yang, J.~S. Park, M.~Salehi, N.~Muennighoff, K.~Lo, L.~Soldaini \emph{et~al.}, ``{Molmo and pixmo: Open weights and open data for state-of-the-art multimodal models},'' \emph{arXiv preprint arXiv:2409.17146}, 2024.

\bibitem{tongcambrian}
S.~Tong, E.~L. Brown~II, P.~Wu, S.~Woo, A.~J. IYER, S.~C. Akula, S.~Yang, J.~Yang, M.~Middepogu, Z.~Wang \emph{et~al.}, ``Cambrian-1: A fully open, vision-centric exploration of multimodal llms,'' in \emph{Advances in Neural Information Processing Systems}, 2024.

\bibitem{agrawal2024pixtral}
P.~Agrawal, S.~Antoniak, E.~B. Hanna, B.~Bout, D.~Chaplot, J.~Chudnovsky, D.~Costa, B.~De~Monicault, S.~Garg, T.~Gervet \emph{et~al.}, ``Pixtral 12b,'' \emph{arXiv preprint arXiv:2410.07073}, 2024.

\bibitem{wu2024deepseek}
Z.~Wu, X.~Chen, Z.~Pan, X.~Liu, W.~Liu, D.~Dai, H.~Gao, Y.~Ma, C.~Wu, B.~Wang \emph{et~al.}, ``Deepseek-vl2: Mixture-of-experts vision-language models for advanced multimodal understanding,'' \emph{arXiv preprint arXiv:2412.10302}, 2024.

\bibitem{yao2024minicpm}
Y.~Yao, T.~Yu, A.~Zhang, C.~Wang, J.~Cui, H.~Zhu, T.~Cai, H.~Li, W.~Zhao, Z.~He \emph{et~al.}, ``{Minicpm-v: A gpt-4v level mllm on your phone},'' \emph{arXiv preprint arXiv:2408.01800}, 2024.

\bibitem{glm2024chatglm}
T.~GLM, A.~Zeng, B.~Xu, B.~Wang, C.~Zhang, D.~Yin, D.~Rojas, G.~Feng, H.~Zhao, H.~Lai \emph{et~al.}, ``{ChatGLM: A Family of Large Language Models from GLM-130B to GLM-4 All Tools},'' \emph{arXiv preprint arXiv:2406.12793}, 2024.

\bibitem{lu2024ovis}
S.~Lu, Y.~Li, Q.-G. Chen, Z.~Xu, W.~Luo, K.~Zhang, and H.-J. Ye, ``Ovis: Structural embedding alignment for multimodal large language model,'' \emph{arXiv preprint arXiv:2405.20797}, 2024.

\bibitem{gpt4omini}
{OpenAI}, ``{GPT}-4o mini: advancing cost-efficient intelligence,'' \url{https://openai.com/index/gpt-4o-mini-advancing-cost-efficient-intelligence}, 2024, accessed: 2024-12-29.

\bibitem{team2023gemini}
G.~Team, R.~Anil, S.~Borgeaud, J.-B. Alayrac, J.~Yu, R.~Soricut, J.~Schalkwyk, A.~M. Dai, A.~Hauth, K.~Millican \emph{et~al.}, ``{Gemini: a family of highly capable multimodal models},'' \emph{arXiv preprint arXiv:2312.11805}, 2023.

\bibitem{Anthropic2024Claude3.5Sonnet}
Anthropic, ``{Claude 3.5 Sonnet},'' \url{https://www.anthropic.com/news/claude-3-5-sonnet}, 2024, accessed: 2024-12-29.

\bibitem{website_step1v}
StepFun, ``{Step-1V},'' \url{https://www.stepfun.com/##step1v}, 2024, accessed: 2024-12-29.

\bibitem{zhong2020image}
X.~Zhong, E.~ShafieiBavani, and A.~Jimeno~Yepes, ``{Image-based table recognition: data, model, and evaluation},'' in \emph{Proceedings of European Conference on Computer Vision}.\hskip 1em plus 0.5em minus 0.4em\relax Springer, 2020, pp. 564--580.

\bibitem{papineni2002bleu}
K.~Papineni, S.~Roukos, T.~Ward, and W.-J. Zhu, ``{Bleu: a method for automatic evaluation of machine translation},'' in \emph{Proceedings of Annual Meeting of the Association for Computational Linguistics}, 2002, pp. 311--318.

\bibitem{banerjee2005meteor}
S.~Banerjee and A.~Lavie, ``{METEOR}: An automatic metric for mt evaluation with improved correlation with human judgments,'' in \emph{Proceedings of the ACL Workshop on Intrinsic and Extrinsic Evaluation Measures for Machine Translation and/or Summarization}, 2005, pp. 65--72.

\bibitem{wang2024qwen2}
P.~Wang, S.~Bai, S.~Tan, S.~Wang, Z.~Fan, J.~Bai, K.~Chen, X.~Liu, J.~Wang, W.~Ge \emph{et~al.}, ``{Qwen2-vl: Enhancing vision-language model's perception of the world at any resolution},'' \emph{arXiv preprint arXiv:2409.12191}, 2024.

\bibitem{shi2016end}
B.~Shi, X.~Bai, and C.~Yao, ``An end-to-end trainable neural network for image-based sequence recognition and its application to scene text recognition,'' \emph{IEEE Transactions on Pattern Analysis and Machine Intelligence}, vol.~39, no.~11, pp. 2298--2304, 2016.

\bibitem{fang2021read}
S.~Fang, H.~Xie, Y.~Wang, Z.~Mao, and Y.~Zhang, ``Read like humans: Autonomous, bidirectional and iterative language modeling for scene text recognition,'' in \emph{Proceedings of the IEEE/CVF Conference on Computer Vision and Pattern Recognition}, 2021, pp. 7098--7107.

\bibitem{shi2018aster}
B.~Shi, M.~Yang, X.~Wang, P.~Lyu, C.~Yao, and X.~Bai, ``Aster: An attentional scene text recognizer with flexible rectification,'' \emph{IEEE Transactions on Pattern Analysis and Machine Intelligence}, vol.~41, no.~9, pp. 2035--2048, 2018.

\bibitem{lu2021master}
N.~Lu, W.~Yu, X.~Qi, Y.~Chen, P.~Gong, R.~Xiao, and X.~Bai, ``Master: Multi-aspect non-local network for scene text recognition,'' \emph{Pattern Recognition}, vol. 117, p. 107980, 2021.

\bibitem{DBLP:conf/ijcai/DuCJYZLDJ22}
\BIBentryALTinterwordspacing
Y.~Du, Z.~Chen, C.~Jia, X.~Yin, T.~Zheng, C.~Li, Y.~Du, and Y.~Jiang, ``{SVTR:} scene text recognition with a single visual model,'' in \emph{Proceedings of the International Joint Conference on Artificial Intelligence}, L.~D. Raedt, Ed.\hskip 1em plus 0.5em minus 0.4em\relax ijcai.org, 2022, pp. 884--890. [Online]. Available: \url{https://doi.org/10.24963/ijcai.2022/124}
\BIBentrySTDinterwordspacing

\bibitem{liu2020abcnet}
Y.~Liu, H.~Chen, C.~Shen, T.~He, L.~Jin, and L.~Wang, ``Abcnet: Real-time scene text spotting with adaptive bezier-curve network,'' in \emph{Proceedings of the IEEE/CVF Conference on Computer Vision and Pattern Recognition}, 2020, pp. 9809--9818.

\bibitem{liu2021abcnet}
Y.~Liu, C.~Shen, L.~Jin, T.~He, P.~Chen, C.~Liu, and H.~Chen, ``Abcnet v2: Adaptive bezier-curve network for real-time end-to-end text spotting,'' \emph{IEEE Transactions on Pattern Analysis and Machine Intelligence}, vol.~44, no.~11, pp. 8048--8064, 2021.

\bibitem{zhang2022text}
X.~Zhang, Y.~Su, S.~Tripathi, and Z.~Tu, ``Text spotting transformers,'' in \emph{Proceedings of the IEEE/CVF Conference on Computer Vision and Pattern Recognition}, 2022, pp. 9519--9528.

\bibitem{ch2017total}
C.~K. Ch'ng and C.~S. Chan, ``Total-text: A comprehensive dataset for scene text detection and recognition,'' in \emph{Proceedings of International Conference on Document Analysis and Recognition}, vol.~1.\hskip 1em plus 0.5em minus 0.4em\relax IEEE, 2017, pp. 935--942.

\bibitem{wei2024general}
H.~Wei, C.~Liu, J.~Chen, J.~Wang, L.~Kong, Y.~Xu, Z.~Ge, L.~Zhao, J.~Sun, Y.~Peng \emph{et~al.}, ``General ocr theory: Towards ocr-2.0 via a unified end-to-end model,'' \emph{arXiv preprint arXiv:2409.01704}, 2024.

\bibitem{6628859}
D.~Karatzas, F.~Shafait, S.~Uchida, M.~Iwamura, L.~G.~i. Bigorda, S.~R. Mestre, J.~Mas, D.~F. Mota, J.~A. Almazàn, and L.~P. de~las Heras, ``Icdar 2013 robust reading competition,'' in \emph{Proceedings of International Conference on Document Analysis and Recognition}, 2013, pp. 1484--1493.

\bibitem{Shi2014EndtoendST}
\BIBentryALTinterwordspacing
C.~Shi, C.~Wang, B.~Xiao, S.~Gao, and J.~Hu, ``End-to-end scene text recognition using tree-structured models,'' \emph{Pattern Recognition}, vol.~47, pp. 2853--2866, 2014. [Online]. Available: \url{https://api.semanticscholar.org/CorpusID:30201169}
\BIBentrySTDinterwordspacing

\bibitem{MishraBMVC12}
A.~Mishra, K.~Alahari, and C.~V. Jawahar, ``Scene text recognition using higher order language priors,'' in \emph{British Machine Vision Conference}, 2012.

\bibitem{karatzas2015icdar}
D.~Karatzas, L.~Gomez-Bigorda, A.~Nicolaou, S.~Ghosh, A.~Bagdanov, M.~Iwamura, J.~Matas, L.~Neumann, V.~R. Chandrasekhar, S.~Lu \emph{et~al.}, ``Icdar 2015 competition on robust reading,'' in \emph{Proceedings of International Conference on Document Analysis and Recognition}.\hskip 1em plus 0.5em minus 0.4em\relax IEEE, 2015, pp. 1156--1160.

\bibitem{10.1016/j.patcog.2019.02.002}
\BIBentryALTinterwordspacing
Y.~Liu, L.~Jin, S.~Zhang, C.~Luo, and S.~Zhang, ``Curved scene text detection via transverse and longitudinal sequence connection,'' \emph{Pattern Recognition}, vol.~90, no.~C, p. 337–345, Jun. 2019. [Online]. Available: \url{https://doi.org/10.1016/j.patcog.2019.02.002}
\BIBentrySTDinterwordspacing

\bibitem{veit2016coco}
A.~Veit, T.~Matera, L.~Neumann, J.~Matas, and S.~Belongie, ``Coco-text: Dataset and benchmark for text detection and recognition in natural images,'' \emph{arXiv preprint arXiv:1601.07140}, 2016.

\bibitem{Risnumawan2014ARA}
\BIBentryALTinterwordspacing
A.~Risnumawan, P.~Shivakumara, C.~S. Chan, and C.~L. Tan, ``A robust arbitrary text detection system for natural scene images,'' \emph{Expert Systems with Applications}, vol.~41, pp. 8027--8048, 2014. [Online]. Available: \url{https://api.semanticscholar.org/CorpusID:15559857}
\BIBentrySTDinterwordspacing

\bibitem{DBLP:conf/iccv/PhanSTT13}
\BIBentryALTinterwordspacing
T.~Q. Phan, P.~Shivakumara, S.~Tian, and C.~L. Tan, ``Recognizing text with perspective distortion in natural scenes,'' in \emph{Proceedings of IEEE/CVF International Conference on Computer Vision}.\hskip 1em plus 0.5em minus 0.4em\relax {IEEE} Computer Society, 2013, pp. 569--576. [Online]. Available: \url{https://doi.org/10.1109/ICCV.2013.76}
\BIBentrySTDinterwordspacing

\bibitem{10.1007/978-3-031-19815-1_18}
X.~Xie, L.~Fu, Z.~Zhang, Z.~Wang, and X.~Bai, ``Toward understanding wordart: Corner-guided transformer for scene text recognition,'' in \emph{Proceedings of European Conference on Computer Vision}, S.~Avidan, G.~Brostow, M.~Ciss{\'e}, G.~M. Farinella, and T.~Hassner, Eds.\hskip 1em plus 0.5em minus 0.4em\relax Cham: Springer Nature Switzerland, 2022, pp. 303--321.

\bibitem{marti2002iam}
U.-V. Marti and H.~Bunke, ``The iam-database: an english sentence database for offline handwriting recognition,'' \emph{International Journal on Document Analysis and Recognition}, vol.~5, pp. 39--46, 2002.

\bibitem{6981115}
M.~Diem, S.~Fiel, F.~Kleber, R.~Sablatnig, J.~M. Saavedra, D.~Contreras, J.~M. Barrios, and L.~S. Oliveira, ``Proceedings of ieee international conference on frontiers in handwriting recognition,'' in \emph{2014 14th International Conference on Frontiers in Handwriting Recognition}, 2014, pp. 779--784.

\bibitem{wang2021two}
Y.~Wang, H.~Xie, S.~Fang, J.~Wang, S.~Zhu, and Y.~Zhang, ``From two to one: A new scene text recognizer with visual language modeling network,'' in \emph{Proceedings of IEEE/CVF International Conference on Computer Vision}, 2021, pp. 14\,194--14\,203.

\bibitem{yuliang2017detecting}
L.~Yuliang, J.~Lianwen, Z.~Shuaitao, and Z.~Sheng, ``Detecting curve text in the wild: New dataset and new solution,'' \emph{arXiv preprint arXiv:1712.02170}, 2017.

\bibitem{long2022towards}
S.~Long, S.~Qin, D.~Panteleev, A.~Bissacco, Y.~Fujii, and M.~Raptis, ``Towards end-to-end unified scene text detection and layout analysis,'' in \emph{Proceedings of the IEEE/CVF Conference on Computer Vision and Pattern Recognition}, 2022, pp. 1049--1059.

\bibitem{1557128390388-1539606084}
\BIBentryALTinterwordspacing
T.-L. Yuan, Z.~Zhu, K.~Xu, C.-J. Li, T.-J. Mu, and S.-M. Hu, ``A large chinese text dataset in the wild,'' \emph{Journal of Computer Science and Technology}, vol.~34, no.~3, pp. 509--521, 2019. [Online]. Available: \url{https://jcst.ict.ac.cn/en/article/doi/10.1007/s11390-019-1923-y}
\BIBentrySTDinterwordspacing

\bibitem{shi2017icdar2017}
B.~Shi, C.~Yao, M.~Liao, M.~Yang, P.~Xu, L.~Cui, S.~Belongie, S.~Lu, and X.~Bai, ``Icdar2017 competition on reading chinese text in the wild (rctw-17),'' in \emph{Proceedings of International Conference on Document Analysis and Recognition}, vol.~1.\hskip 1em plus 0.5em minus 0.4em\relax IEEE, 2017, pp. 1429--1434.

\bibitem{liu2019icdar2019robustreading}
\BIBentryALTinterwordspacing
X.~Liu, R.~Zhang, Y.~Zhou, Q.~Jiang, Q.~Song, N.~Li, K.~Zhou, L.~Wang, D.~Wang, M.~Liao, M.~Yang, X.~Bai, B.~Shi, D.~Karatzas, S.~Lu, and C.~V. Jawahar, ``Icdar 2019 robust reading challenge on reading chinese text on signboard,'' 2019. [Online]. Available: \url{https://arxiv.org/abs/1912.09641}
\BIBentrySTDinterwordspacing

\bibitem{sun2020chinesestreetviewtext}
\BIBentryALTinterwordspacing
Y.~Sun, J.~Liu, W.~Liu, J.~Han, E.~Ding, and J.~Liu, ``Chinese street view text: Large-scale chinese text reading with partially supervised learning,'' 2020. [Online]. Available: \url{https://arxiv.org/abs/1909.07808}
\BIBentrySTDinterwordspacing

\bibitem{cheng2023m6doclargescalemultiformatmultitype}
\BIBentryALTinterwordspacing
H.~Cheng, P.~Zhang, S.~Wu, J.~Zhang, Q.~Zhu, Z.~Xie, J.~Li, K.~Ding, and L.~Jin, ``M$^{6}$doc: A large-scale multi-format, multi-type, multi-layout, multi-language, multi-annotation category dataset for modern document layout analysis,'' 2023. [Online]. Available: \url{https://arxiv.org/abs/2305.08719}
\BIBentrySTDinterwordspacing

\bibitem{deka2017rico}
B.~Deka, Z.~Huang, C.~Franzen, J.~Hibschman, D.~Afergan, Y.~Li, J.~Nichols, and R.~Kumar, ``Rico: A mobile app dataset for building data-driven design applications,'' in \emph{Proceedings of the 30th annual ACM symposium on user interface software and technology}, 2017, pp. 845--854.

\bibitem{jaume2019funsd}
G.~Jaume, H.~K. Ekenel, and J.-P. Thiran, ``Funsd: A dataset for form understanding in noisy scanned documents,'' in \emph{Proceedings of International Conference on Document Analysis and Recognition Workshops}, vol.~2.\hskip 1em plus 0.5em minus 0.4em\relax IEEE, 2019, pp. 1--6.

\bibitem{huang2019icdar2019}
Z.~Huang, K.~Chen, J.~He, X.~Bai, D.~Karatzas, S.~Lu, and C.~Jawahar, ``Icdar2019 competition on scanned receipt ocr and information extraction,'' in \emph{Proceedings of International Conference on Document Analysis and Recognition}.\hskip 1em plus 0.5em minus 0.4em\relax IEEE, 2019, pp. 1516--1520.

\bibitem{kuang2023visual}
J.~Kuang, W.~Hua, D.~Liang, M.~Yang, D.~Jiang, B.~Ren, and X.~Bai, ``Visual information extraction in the wild: practical dataset and end-to-end solution,'' in \emph{Proceedings of International Conference on Document Analysis and Recognition}.\hskip 1em plus 0.5em minus 0.4em\relax Springer, 2023, pp. 36--53.

\bibitem{xu-etal-2022-xfund}
\BIBentryALTinterwordspacing
Y.~Xu, T.~Lv, L.~Cui, G.~Wang, Y.~Lu, D.~Florencio, C.~Zhang, and F.~Wei, ``{XFUND}: A benchmark dataset for multilingual visually rich form understanding,'' in \emph{Proceedings of Annual Meeting of the Association for Computational Linguistics}.\hskip 1em plus 0.5em minus 0.4em\relax Dublin, Ireland: Association for Computational Linguistics, May 2022, pp. 3214--3224. [Online]. Available: \url{https://aclanthology.org/2022.findings-acl.253}
\BIBentrySTDinterwordspacing

\bibitem{yu2023icdar2023competitionstructured}
\BIBentryALTinterwordspacing
W.~Yu, C.~Zhang, H.~Cao, W.~Hua, B.~Li, H.~Chen, M.~Liu, M.~Chen, J.~Kuang, M.~Cheng, Y.~Du, S.~Feng, X.~Hu, P.~Lyu, K.~Yao, Y.~Yu, Y.~Liu, W.~Che, E.~Ding, C.-L. Liu, J.~Luo, S.~Yan, M.~Zhang, D.~Karatzas, X.~Sun, J.~Wang, and X.~Bai, ``Icdar 2023 competition on structured text extraction from visually-rich document images,'' 2023. [Online]. Available: \url{https://arxiv.org/abs/2306.03287}
\BIBentrySTDinterwordspacing

\bibitem{Long_2021_ICCV}
L.~Rujiao, W.~Wen, X.~Nan, G.~Feiyu, Y.~Zhibo, W.~Yongpan, and X.~Gui-Song, ``Parsing table structures in the wild,'' in \emph{Proceedings of IEEE/CVF International Conference on Computer Vision}, October 2021.

\bibitem{yang2023large}
F.~Yang, L.~Hu, X.~Liu, S.~Huang, and Z.~Gu, ``A large-scale dataset for end-to-end table recognition in the wild,'' \emph{Scientific Data}, vol.~10, no.~1, p. 110, 2023.

\bibitem{liang2024document}
Y.~Liang, Y.~Zhang, C.~Ma, Z.~Zhang, Y.~Zhao, L.~Xiang, C.~Zong, and Y.~Zhou, ``Document image machine translation with dynamic multi-pre-trained models assembling,'' in \emph{Proceedings of the 2024 Conference of the North American Chapter of the Association for Computational Linguistics: Human Language Technologies}, 2024, pp. 7077--7088.

\bibitem{DBLP:conf/wacv/MathewKJ21}
M.~Mathew, D.~Karatzas, and C.~V. Jawahar, ``Docvqa: {A} dataset for {VQA} on document images,'' in \emph{Proceedings of the IEEE Winter Conference on Applications of Computer Vision}.\hskip 1em plus 0.5em minus 0.4em\relax {IEEE}, 2021, pp. 2199--2208.

\bibitem{yuan2022syntax}
Y.~Yuan, X.~Liu, W.~Dikubab, H.~Liu, Z.~Ji, Z.~Wu, and X.~Bai, ``Syntax-aware network for handwritten mathematical expression recognition,'' \emph{arXiv preprint arXiv:2203.01601}, 2022.

\bibitem{DBLP:journals/corr/abs-2402-14804}
K.~Wang, J.~Pan, W.~Shi, Z.~Lu, M.~Zhan, and H.~Li, ``Measuring multimodal mathematical reasoning with math-vision dataset,'' \emph{CoRR}, vol. abs/2402.14804, 2024.

\bibitem{DBLP:conf/mm/YangLPJHY23}
W.~Yang, Z.~Li, D.~Peng, L.~Jin, M.~He, and C.~Yao, ``Read ten lines at one glance: Line-aware semi-autoregressive transformer for multi-line handwritten mathematical expression recognition,'' in \emph{Proceedings of the ACM International Conference on Multimedia}, A.~El{-}Saddik, T.~Mei, R.~Cucchiara, M.~Bertini, D.~P.~T. Vallejo, P.~K. Atrey, and M.~S. Hossain, Eds.\hskip 1em plus 0.5em minus 0.4em\relax {ACM}, 2023, pp. 2066--2077.

\bibitem{DBLP:journals/corr/abs-2404-10690}
P.~Gervais, A.~Fadeeva, and A.~Maksai, ``Mathwriting: {A} dataset for handwritten mathematical expression recognition,'' \emph{CoRR}, vol. abs/2404.10690, 2024.

\bibitem{9956654}
L.~Ding, M.~Zhao, F.~Yin, S.~Zeng, and C.-L. Liu, ``A large-scale database for chemical structure recognition and preliminary evaluation,'' in \emph{Proceedings of the International Conference on Pattern Recognition}, 2022, pp. 1464--1470.

\bibitem{DBLP:conf/iclr/SaxtonGHK19}
D.~Saxton, E.~Grefenstette, F.~Hill, and P.~Kohli, ``Analysing mathematical reasoning abilities of neural models,'' in \emph{Proceedings of the International Conference on Learning Representations}.\hskip 1em plus 0.5em minus 0.4em\relax OpenReview.net, 2019.

\bibitem{DBLP:conf/eccv/ZhangJZLGQZLCQGL24}
R.~Zhang, D.~Jiang, Y.~Zhang, H.~Lin, Z.~Guo, P.~Qiu, A.~Zhou, P.~Lu, K.~Chang, Y.~Qiao, P.~Gao, and H.~Li, ``{MATHVERSE:} does your multi-modal {LLM} truly see the diagrams in visual math problems?'' in \emph{Proceedings of European Conference on Computer Vision}, ser. Lecture Notes in Computer Science, A.~Leonardis, E.~Ricci, S.~Roth, O.~Russakovsky, T.~Sattler, and G.~Varol, Eds., vol. 15066.\hskip 1em plus 0.5em minus 0.4em\relax Springer, 2024, pp. 169--186.

\bibitem{DBLP:conf/iclr/LuBX0LH0CG024}
P.~Lu, H.~Bansal, T.~Xia, J.~Liu, C.~Li, H.~Hajishirzi, H.~Cheng, K.~Chang, M.~Galley, and J.~Gao, ``Mathvista: Evaluating mathematical reasoning of foundation models in visual contexts,'' in \emph{Proceedings of the International Conference on Learning Representations}.\hskip 1em plus 0.5em minus 0.4em\relax OpenReview.net, 2024.

\bibitem{mishra2019ocr}
A.~Mishra, S.~Shekhar, A.~K. Singh, and A.~Chakraborty, ``Ocr-vqa: Visual question answering by reading text in images,'' in \emph{Proceedings of International Conference on Document Analysis and Recognition}.\hskip 1em plus 0.5em minus 0.4em\relax IEEE, 2019, pp. 947--952.

\bibitem{feng2022geometric}
H.~Feng, W.~Zhou, J.~Deng, Y.~Wang, and H.~Li, ``Geometric representation learning for document image rectification,'' in \emph{Proceedings of European Conference on Computer Vision}.\hskip 1em plus 0.5em minus 0.4em\relax Springer, 2022, pp. 475--492.

\bibitem{kafle2018dvqa}
K.~Kafle, S.~Cohen, B.~Price, and C.~Kanan, ``Dvqa: Understanding data visualizations via question answering,'' in \emph{Proceedings of the IEEE/CVF Conference on Computer Vision and Pattern Recognition}, 2018.

\bibitem{Methani_2020_WACV}
N.~Methani, P.~Ganguly, M.~M. Khapra, and P.~Kumar, ``Plotqa: Reasoning over scientific plots,'' in \emph{Proceedings of the IEEE Winter Conference on Applications of Computer Vision}, March 2020.

\bibitem{DBLP:journals/corr/abs-2104-12756}
\BIBentryALTinterwordspacing
M.~Mathew, V.~Bagal, R.~P. Tito, D.~Karatzas, E.~Valveny, and C.~V. Jawahar, ``Infographicvqa,'' \emph{CoRR}, vol. abs/2104.12756, 2021. [Online]. Available: \url{https://arxiv.org/abs/2104.12756}
\BIBentrySTDinterwordspacing

\bibitem{zhong2019image}
X.~Zhong, E.~ShafieiBavani, and A.~J. Yepes, ``Image-based table recognition: data, model, and evaluation,'' \emph{arXiv preprint arXiv:1911.10683}, 2019.

\bibitem{pasupat-liang-2015-compositional}
\BIBentryALTinterwordspacing
P.~Pasupat and P.~Liang, ``Compositional semantic parsing on semi-structured tables,'' in \emph{Proceedings of Annual Meeting of the Association for Computational Linguistics}, C.~Zong and M.~Strube, Eds.\hskip 1em plus 0.5em minus 0.4em\relax Beijing, China: Association for Computational Linguistics, Jul. 2015, pp. 1470--1480. [Online]. Available: \url{https://aclanthology.org/P15-1142}
\BIBentrySTDinterwordspacing

\bibitem{park2019cord}
S.~Park, S.~Shin, B.~Lee, J.~Lee, J.~Surh, M.~Seo, and H.~Lee, ``Cord: a consolidated receipt dataset for post-ocr parsing,'' in \emph{Advances in Neural Information Processing Systems Workshop}, 2019.

\bibitem{chen2021websrc}
X.~Chen, Z.~Zhao, L.~Chen, D.~Zhang, J.~Ji, A.~Luo, Y.~Xiong, and K.~Yu, ``Websrc: A dataset for web-based structural reading comprehension,'' \emph{arXiv preprint arXiv:2101.09465}, 2021.

\bibitem{DBLP:conf/icdar/ZhongTJ19}
X.~Zhong, J.~Tang, and A.~Jimeno{-}Yepes, ``Publaynet: Largest dataset ever for document layout analysis,'' in \emph{Proceedings of International Conference on Document Analysis and Recognition}.\hskip 1em plus 0.5em minus 0.4em\relax {IEEE}, 2019, pp. 1015--1022.

\bibitem{harley2015icdar}
A.~W. Harley, A.~Ufkes, and K.~G. Derpanis, ``Evaluation of deep convolutional nets for document image classification and retrieval,'' in \emph{Proceedings of International Conference on Document Analysis and Recognition}, 2015.

\bibitem{baechler2024screenai}
G.~Baechler, S.~Sunkara, M.~Wang, F.~Zubach, H.~Mansoor, V.~Etter, V.~Cărbune, J.~Lin, J.~Chen, and A.~Sharma, ``Screenai: A vision-language model for ui and infographics understanding,'' 2024.

\bibitem{SlideVQA2023}
R.~Tanaka, K.~Nishida, K.~Nishida, T.~Hasegawa, I.~Saito, and K.~Saito, ``Slidevqa: A dataset for document visual question answering on multiple images,'' in \emph{Proceedings of the AAAI Conference on Artificial Intelligence}, 2023.

\bibitem{kembhavi2016diagram}
A.~Kembhavi, M.~Salvato, E.~Kolve, M.~Seo, H.~Hajishirzi, and A.~Farhadi, ``A diagram is worth a dozen images,'' in \emph{Proceedings of European Conference on Computer Vision}.\hskip 1em plus 0.5em minus 0.4em\relax Springer, 2016, pp. 235--251.

\bibitem{kembhavi2017you}
A.~Kembhavi, M.~Seo, D.~Schwenk, J.~Choi, A.~Farhadi, and H.~Hajishirzi, ``Are you smarter than a sixth grader? textbook question answering for multimodal machine comprehension,'' in \emph{Proceedings of the IEEE/CVF Conference on Computer Vision and Pattern Recognition}, 2017, pp. 4999--5007.

\bibitem{lee2024modeling}
S.~Lee, B.~Lai, F.~Ryan, B.~Boote, and J.~M. Rehg, ``Modeling multimodal social interactions: New challenges and baselines with densely aligned representations,'' in \emph{Proceedings of the IEEE/CVF Conference on Computer Vision and Pattern Recognition}, 2024, pp. 14\,585--14\,595.

\bibitem{zhang2024cmmmuchinesemassivemultidiscipline}
\BIBentryALTinterwordspacing
G.~Zhang, X.~Du, B.~Chen, Y.~Liang, T.~Luo, T.~Zheng, K.~Zhu, Y.~Cheng, C.~Xu, S.~Guo, H.~Zhang, X.~Qu, J.~Wang, R.~Yuan, Y.~Li, Z.~Wang, Y.~Liu, Y.-H. Tsai, F.~Zhang, C.~Lin, W.~Huang, and J.~Fu, ``Cmmmu: A chinese massive multi-discipline multimodal understanding benchmark,'' 2024. [Online]. Available: \url{https://arxiv.org/abs/2401.11944}
\BIBentrySTDinterwordspacing

\bibitem{lu2022learn}
P.~Lu, S.~Mishra, T.~Xia, L.~Qiu, K.-W. Chang, S.-C. Zhu, O.~Tafjord, P.~Clark, and A.~Kalyan, ``Learn to explain: Multimodal reasoning via thought chains for science question answering,'' \emph{Advances in Neural Information Processing Systems}, vol.~35, pp. 2507--2521, 2022.

\bibitem{yue2024mmmu}
X.~Yue, T.~Zheng, Y.~Ni, Y.~Wang, K.~Zhang, S.~Tong, Y.~Sun, B.~Yu, G.~Zhang, H.~Sun \emph{et~al.}, ``Mmmu-pro: A more robust multi-discipline multimodal understanding benchmark,'' \emph{arXiv preprint arXiv:2409.02813}, 2024.

\bibitem{jia2024visual}
Q.~Jia, X.~Yue, S.~Huang, Z.~Qin, Y.~Liu, B.~Y. Lin, and Y.~You, ``Visual perception in text strings,'' \emph{arXiv preprint arXiv:2410.01733}, 2024.

\bibitem{yao2012detecting}
C.~Yao, X.~Bai, W.~Liu, Y.~Ma, and Z.~Tu, ``Detecting texts of arbitrary orientations in natural images,'' in \emph{Proceedings of the IEEE/CVF Conference on Computer Vision and Pattern Recognition}.\hskip 1em plus 0.5em minus 0.4em\relax IEEE, 2012, pp. 1083--1090.

\bibitem{8546143}
M.~He, Y.~Liu, Z.~Yang, S.~Zhang, C.~Luo, F.~Gao, Q.~Zheng, Y.~Wang, X.~Zhang, and L.~Jin, ``Icpr2018 contest on robust reading for multi-type web images,'' in \emph{Proceedings of the International Conference on Pattern Recognition}, 2018, pp. 7--12.

\bibitem{shi2018icdar2017competitionreadingchinese}
\BIBentryALTinterwordspacing
B.~Shi, C.~Yao, M.~Liao, M.~Yang, P.~Xu, L.~Cui, S.~Belongie, S.~Lu, and X.~Bai, ``Icdar2017 competition on reading chinese text in the wild (rctw-17),'' 2018. [Online]. Available: \url{https://arxiv.org/abs/1708.09585}
\BIBentrySTDinterwordspacing

\bibitem{TANG2019106954}
\BIBentryALTinterwordspacing
J.~Tang, Z.~Yang, Y.~Wang, Q.~Zheng, Y.~Xu, and X.~Bai, ``Seglink++: Detecting dense and arbitrary-shaped scene text by instance-aware component grouping,'' \emph{Pattern Recognition}, vol.~96, p. 106954, 2019. [Online]. Available: \url{https://www.sciencedirect.com/science/article/pii/S0031320319302511}
\BIBentrySTDinterwordspacing

\bibitem{chng2019icdar2019}
C.~K. Chng, Y.~Liu, Y.~Sun, C.~C. Ng, C.~Luo, Z.~Ni, C.~Fang, S.~Zhang, J.~Han, E.~Ding \emph{et~al.}, ``Icdar2019 robust reading challenge on arbitrary-shaped text-rrc-art,'' in \emph{Proceedings of International Conference on Document Analysis and Recognition}.\hskip 1em plus 0.5em minus 0.4em\relax IEEE, 2019, pp. 1571--1576.

\bibitem{zheng2020global}
\BIBentryALTinterwordspacing
X.~Zheng, D.~Burdick, L.~Popa, X.~Zhong, and N.~X.~R. Wang, ``Global table extractor {(GTE):} {A} framework for joint table identification and cell structure recognition using visual context,'' in \emph{Proceedings of the IEEE Winter Conference on Applications of Computer Vision}.\hskip 1em plus 0.5em minus 0.4em\relax {IEEE}, 2021, pp. 697--706. [Online]. Available: \url{https://doi.org/10.1109/WACV48630.2021.00074}
\BIBentrySTDinterwordspacing

\bibitem{dong2024internlm}
X.~Dong, P.~Zhang, Y.~Zang, Y.~Cao, B.~Wang, L.~Ouyang, S.~Zhang, H.~Duan, W.~Zhang, Y.~Li \emph{et~al.}, ``{Internlm-xcomposer2-4khd: A pioneering large vision-language model handling resolutions from 336 pixels to 4k hd},'' \emph{arXiv preprint arXiv:2404.06512}, 2024.

\bibitem{sun2024generative}
Q.~Sun, Y.~Cui, X.~Zhang, F.~Zhang, Q.~Yu, Y.~Wang, Y.~Rao, J.~Liu, T.~Huang, and X.~Wang, ``{Generative multimodal models are in-context learners},'' in \emph{Proceedings of the IEEE/CVF Conference on Computer Vision and Pattern Recognition}, 2024, pp. 14\,398--14\,409.

\bibitem{ye2024mplug}
J.~Ye, H.~Xu, H.~Liu, A.~Hu, M.~Yan, Q.~Qian, J.~Zhang, F.~Huang, and J.~Zhou, ``{mplug-owl3: Towards long image-sequence understanding in multi-modal large language models},'' \emph{arXiv preprint arXiv:2408.04840}, 2024.

\bibitem{wang2023cogvlm}
W.~Wang, Q.~Lv, W.~Yu, W.~Hong, J.~Qi, Y.~Wang, J.~Ji, Z.~Yang, L.~Zhao, X.~Song \emph{et~al.}, ``{Cogvlm: Visual expert for pretrained language models},'' \emph{arXiv preprint arXiv:2311.03079}, 2023.

\bibitem{chen2024internvl}
Z.~Chen, J.~Wu, W.~Wang, W.~Su, G.~Chen, S.~Xing, M.~Zhong, Q.~Zhang, X.~Zhu, L.~Lu \emph{et~al.}, ``{Internvl: Scaling up vision foundation models and aligning for generic visual-linguistic tasks},'' in \emph{Proceedings of the IEEE/CVF Conference on Computer Vision and Pattern Recognition}, 2024, pp. 24\,185--24\,198.

\bibitem{lu2024deepseek}
H.~Lu, W.~Liu, B.~Zhang, B.~Wang, K.~Dong, B.~Liu, J.~Sun, T.~Ren, Z.~Li, H.~Yang \emph{et~al.}, ``{Deepseek-vl: towards real-world vision-language understanding},'' \emph{arXiv preprint arXiv:2403.05525}, 2024.

\bibitem{liu2024nvila}
Z.~Liu, L.~Zhu, B.~Shi, Z.~Zhang, Y.~Lou, S.~Yang, H.~Xi, S.~Cao, Y.~Gu, D.~Li \emph{et~al.}, ``Nvila: Efficient frontier visual language models,'' \emph{arXiv preprint arXiv:2412.04468}, 2024.

\bibitem{hu2024mplug2}
A.~Hu, H.~Xu, L.~Zhang, J.~Ye, M.~Yan, J.~Zhang, Q.~Jin, F.~Huang, and J.~Zhou, ``mplug-docowl2: High-resolution compressing for ocr-free multi-page document understanding,'' \emph{arXiv preprint arXiv:2409.03420}, 2024.

\bibitem{young2024yi}
A.~Young, B.~Chen, C.~Li, C.~Huang, G.~Zhang, G.~Zhang, H.~Li, J.~Zhu, J.~Chen, J.~Chang \emph{et~al.}, ``Yi: Open foundation models by 01. ai,'' \emph{arXiv preprint arXiv:2403.04652}, 2024.

\bibitem{wu2024janus}
C.~Wu, X.~Chen, Z.~Wu, Y.~Ma, X.~Liu, Z.~Pan, W.~Liu, Z.~Xie, X.~Yu, C.~Ruan \emph{et~al.}, ``Janus: Decoupling visual encoding for unified multimodal understanding and generation,'' \emph{arXiv preprint arXiv:2410.13848}, 2024.

\bibitem{shi2024eagle}
M.~Shi, F.~Liu, S.~Wang, S.~Liao, S.~Radhakrishnan, D.-A. Huang, H.~Yin, K.~Sapra, Y.~Yacoob, H.~Shi \emph{et~al.}, ``Eagle: Exploring the design space for multimodal llms with mixture of encoders,'' \emph{arXiv preprint arXiv:2408.15998}, 2024.

\bibitem{laurenccon2024building}
H.~Lauren{\c{c}}on, A.~Marafioti, V.~Sanh, and L.~Tronchon, ``Building and better understanding vision-language models: insights and future directions,'' in \emph{Workshop on Responsibly Building the Next Generation of Multimodal Foundational Models}, 2024.

\bibitem{abouelenin2025phi}
A.~Abouelenin, A.~Ashfaq, A.~Atkinson, H.~Awadalla, N.~Bach, J.~Bao, A.~Benhaim, M.~Cai, V.~Chaudhary, C.~Chen \emph{et~al.}, ``Phi-4-mini technical report: Compact yet powerful multimodal language models via mixture-of-loras,'' \emph{arXiv preprint arXiv:2503.01743}, 2025.

\bibitem{duan2024vlmevalkit}
H.~Duan, J.~Yang, Y.~Qiao, X.~Fang, L.~Chen, Y.~Liu, X.~Dong, Y.~Zang, P.~Zhang, J.~Wang \emph{et~al.}, ``Vlmevalkit: An open-source toolkit for evaluating large multi-modality models,'' in \emph{Proceedings of the ACM International Conference on Multimedia}, 2024, pp. 11\,198--11\,201.

\bibitem{kimiteam2025kimivltechnicalreport}
\BIBentryALTinterwordspacing
K.~Team, A.~Du, B.~Yin, B.~Xing, B.~Qu, B.~Wang, C.~Chen, C.~Zhang, C.~Du, C.~Wei, C.~Wang, D.~Zhang, D.~Du, D.~Wang, E.~Yuan, E.~Lu, F.~Li, F.~Sung, G.~Wei, G.~Lai, H.~Zhu, H.~Ding, H.~Hu, H.~Yang, H.~Zhang, H.~Wu, H.~Yao, H.~Lu, H.~Wang, H.~Gao, H.~Zheng, J.~Li, J.~Su, J.~Wang, J.~Deng, J.~Qiu, J.~Xie, J.~Wang, J.~Liu, J.~Yan, K.~Ouyang, L.~Chen, L.~Sui, L.~Yu, M.~Dong, M.~Dong, N.~Xu, P.~Cheng, Q.~Gu, R.~Zhou, S.~Liu, S.~Cao, T.~Yu, T.~Song, T.~Bai, W.~Song, W.~He, W.~Huang, W.~Xu, X.~Yuan, X.~Yao, X.~Wu, X.~Zu, X.~Zhou, X.~Wang, Y.~Charles, Y.~Zhong, Y.~Li, Y.~Hu, Y.~Chen, Y.~Wang, Y.~Liu, Y.~Miao, Y.~Qin, Y.~Chen, Y.~Bao, Y.~Wang, Y.~Kang, Y.~Liu, Y.~Du, Y.~Wu, Y.~Wang, Y.~Yan, Z.~Zhou, Z.~Li, Z.~Jiang, Z.~Zhang, Z.~Yang, Z.~Huang, Z.~Huang, Z.~Zhao, and Z.~Chen, ``{Kimi-VL} technical report,'' 2025. [Online]. Available: \url{https://arxiv.org/abs/2504.07491}
\BIBentrySTDinterwordspacing

\end{thebibliography}

\newpage

\appendix
\label{appendix}

\section{Technical Appendices and Supplementary Material}

This supplementary material contains the following content:
\begin{itemize}[leftmargin=2em]
    \item \textbf{Sec.~\ref{comparison_exp}}: Comparison experiments between LMMs and some text-centric expert models.
    \item \textbf{Sec.~\ref{data_collection}}: Data collection.
    \item \textbf{Sec.~\ref{sec:task}}: Task definitions.
    \item \textbf{Sec.~\ref{more_Statistics}}: Additional statistics of \textit{OCRBench v2}.
    \item \textbf{Sec.~\ref{metric}}: Evaluation metrics.
    \item \textbf{Sec.~\ref{Experimental setting}}: Experimental setting for the evaluation process.
    \item \textbf{Sec.~\ref{Experiments compute resources}}: Compute resources for the evaluation process.
    \item \textbf{Sec.~\ref{sec:Evaluation results}}: Evaluation results for LMMs on \textit{OCRBench v2}.
    \item \textbf{Sec.~\ref{sec:Potential Factors Affecting OCR Capabilities}}: Potential factors affecting OCR capabilities
    \item \textbf{Sec.~\ref{sec:sample_for_task}}: Visualization samples for task examples.
    \item \textbf{Sec.~\ref{sample_for_limit}}: Visualization samples for failure cases.
    \item 
    \textbf{Sec.~\ref{Broader impacts}}: Discussion of broader impacts.
    \item 
    \textbf{Sec.~\ref{Limitations}}: Discussion of limitations.
\end{itemize}

\subsection{Comparison with LMMs and Text-centric Expert Models}
\label{comparison_exp}

\noindent\textbf{Comparison with text recognizers.} We compare LMMs with several representative scene text recognizers, including CRNN~\cite{shi2016end}, ABINet~\cite{fang2021read}, ASTER~\cite{shi2018aster}, MASTER~\cite{lu2021master}, and SVTR~\cite{DBLP:conf/ijcai/DuCJYZLDJ22}, on the text recognition task. The weights of these models are loaded from mmocr\footnote{\url{https://github.com/open-mmlab/mmocr}}. The results are shown in Tab.~\ref{tab:comparsion_with_recognizer}, where we selected $5$ representative LMMs, including Qwen2.5VL-7B~\cite{wang2024qwen2}, InternVL3-14B~\cite{chen2024expanding}, GPT4o~\cite{achiam2023gpt}, Gemini1.5-Pro~\cite{team2023gemini}, and Step-1V~\cite{website_step1v}. The results demonstrate that LMMs exhibit remarkable text recognition capabilities, validating our motivation to evaluate LMMs on more challenging OCR-related tasks.

\begin{table}[ht]
\centering
\small
\renewcommand{\arraystretch}{0.9}
\setlength{\tabcolsep}{3pt}
\addtolength{\tabcolsep}{2pt}
  \centering
  \begin{tabular}{@{}lc@{}}
    \toprule
    Method & Accuracy \\
    \midrule
    CRNN~\cite{shi2016end} & 38.1 \\
    ABINet~\cite{fang2021read} & 62.4 \\
    ASTER~\cite{shi2018aster} & 50.0 \\
    MASTER~\cite{lu2021master} & 54.1 \\
    SVTR~\cite{DBLP:conf/ijcai/DuCJYZLDJ22} & 57.8 \\
    \midrule
    Qwen2.5VL-7B~\cite{wang2024qwen2} & 73.0 \\
    InternVL3-14B~\cite{chen2024expanding}  & 71.1 \\
    GPT4o~\cite{achiam2023gpt} & 74.1 \\
    Gemini1.5-Pro~\cite{team2023gemini} & 64.1 \\  
    Step-1V~\cite{website_step1v} & 75.4 \\
    \bottomrule
  \end{tabular}
  \vspace{4pt}
  \caption{Comparison between LMMs and text recognizers.}
  \label{tab:comparsion_with_recognizer}
    \vspace{-10pt}
\end{table}

\noindent\textbf{Comparison with text spotters.} We also compare LMMs with ABCNet series~\cite{liu2020abcnet, liu2021abcnet} and TESTR~\cite{zhang2022text} on the text spotting task. The ABCNet series utilize the official weights\footnote{\url{https://github.com/aim-uofa/AdelaiDet}}, and TESTR is also initialized with its publicly released checkpoint\footnote{\url{https://github.com/mlpc-ucsd/TESTR}}. These models were fine-tuned with TotalText~\cite{ch2017total}. The results are shown in Tab.~\ref{tab:comparsion_with_spotter}. Although LMMs demonstrate promising capabilities in text recognition, there remains notable potential for improvement in the text spotting task. 

\begin{table}[ht]
\centering
\small
\renewcommand{\arraystretch}{0.9}
\setlength{\tabcolsep}{3pt}
\addtolength{\tabcolsep}{2pt}
  \begin{tabular}{@{}lc@{}}
    \toprule
    Method & F1 score \\
    \midrule
    ABCNet~\cite{liu2020abcnet} & 32.2 \\
    ABCNetV2~\cite{liu2021abcnet} & 44.2 \\
    TESTR~\cite{zhang2022text} & 51.8 \\
    \midrule
    Qwen2.5VL-7B~\cite{wang2024qwen2} & 1.2 \\
    InternVL3-14B~\cite{chen2024expanding} & 11.2 \\
    Gemini1.5-Pro~\cite{team2023gemini} & 13.5 \\
    GPT4o~\cite{achiam2023gpt} & 0 \\
    Step-1V~\cite{website_step1v} & 7.2 \\
    \bottomrule
  \end{tabular}  
  \vspace{4pt}
  \caption{Comparison between LMMs and text spotters.}
  \label{tab:comparsion_with_spotter}
  \vspace{-10pt}
\end{table}

\noindent\textbf{Comparison with GOT.} We notice a recent work, GOT~\cite{wei2024general}, that can parse the textual elements within images. We conduct comparison experiments between GOT and some representative LMMs, and the results are shown in Tab.~\ref{tab:comparsion_with_got}. We observe that LMMs show advantages in general text recognition, while GOT demonstrates better performance in the document parsing task.

\begin{table}[ht]
\centering
\small
\renewcommand{\arraystretch}{0.9}
\setlength{\tabcolsep}{3pt}
\addtolength{\tabcolsep}{2pt}
  \begin{tabular}{@{}lcccc@{}}
    \toprule
    Method & Rec & FG-Rec & Full-Rec & Doc-Parse \\
    \midrule
    GOT~\cite{wei2024general} & 64.1 & 52.9 & 73.3 & 53.9 \\

    \midrule
    Qwen2.5VL-7B~\cite{wang2024qwen2} & 73.0 & 36.4 & 84.2 & 39.1 \\
    InternVL3-14B~\cite{chen2024expanding} & 71.1 & 36.4 & 83.0 & 36.9 \\
    GPT4o~\cite{achiam2023gpt} & 74.1 & 13.8 & 54.1 & 35.9 \\
    Gemini1.5-Pro~\cite{team2023gemini} & 64.1 & 22.9 & 83.9 & 40.5 \\
    Step-1V~\cite{website_step1v} & 76.8 & 24.8 & 74.8 & 36.0 \\

    \bottomrule
  \end{tabular}
  \vspace{4pt}
  \caption{Comparison between LMMs and GOT~\cite{wei2024general}.}
  \label{tab:comparsion_with_got}
  \vspace{-10pt}
\end{table}


\subsection{Data Collection}
\label{data_collection}

\noindent\textbf{Text Recognition.} 
The data for text recognition task are sampled from ICDAR2013~\cite{6628859}, SVT~\cite{Shi2014EndtoendST}, IIIT5K~\cite{MishraBMVC12}, ICDAR2015~\cite{karatzas2015icdar}, SCUT-CTW1500~\cite{10.1016/j.patcog.2019.02.002}, COCO-Text~\cite{veit2016coco}, CUTE80~\cite{Risnumawan2014ARA}, TotalText, SVTP~\cite{DBLP:conf/iccv/PhanSTT13}, WordArt~\cite{10.1007/978-3-031-19815-1_18}, NonSemanticText~\cite{liu2023hidden}, IAM~\cite{marti2002iam}, ORAND-CAR-2014~\cite{6981115}, HOST~\cite{wang2021two}, and WOST~\cite{wang2021two}. Meanwhile, CAPTCHA (Completely Automated Public Turing Test to Tell Humans Apart) images are sourced from a CAPTCHA dataset\footnote{\url{https://aistudio.baidu.com/datasetdetail/159309}} and a number CAPTCHA dataset\footnote{\url{https://www.heywhale.com/mw/dataset/5e5e56b6b8dfce002d7ee42c/file}}. Additionally, dot matrix images in the text recognition task are manually collected from the web page.

\noindent\textbf{Fine-grained Text Recognition.} In the fine-grained text recognition task, images are sampled from the test sets of Fox~\cite{liu2024focus}, Totaltext, COCO-Text, CTW1500~\cite{yuliang2017detecting}, and ICDAR2015. We use the original annotations for Fox, while the other datasets are manually re-annotated.

\noindent\textbf{Full-page OCR.} The data sources for full-page OCR task include Fox, HierText~\cite{long2022towards}, CTW~\cite{1557128390388-1539606084}, RCTW-17~\cite{shi2017icdar2017}, ReCTS~\cite{liu2019icdar2019robustreading}, LSVT2019~\cite{sun2020chinesestreetviewtext}, M6Doc~\cite{cheng2023m6doclargescalemultiformatmultitype}, and CDLA\footnote{\url{https://github.com/buptlihang/CDLA}}.

\noindent\textbf{Text Grounding.} The images for the text grounding task are sampled from testset of Totaltext, COCO-Text, CTW1500, and ICDAR2015. QA pairs and bounding boxes annotations are based on their official OCR annotations.

\noindent\textbf{VQA with Position.} The images used for VQA with position task are sampled from the test sets of TextVQA~\cite{singh2019towards} and RICO~\cite{deka2017rico}, with QA pairs and bounding box annotations derived from their original datasets.

\noindent\textbf{Text Spotting.} The data sources for the text spotting task include Totaltext, COCO-Text, CTW1500, and ICDAR2015.

\noindent\textbf{Key Information Extraction.} The data sources for key information extraction task include FUNSD~\cite{jaume2019funsd}, SROIE~\cite{huang2019icdar2019}, POIE~\cite{kuang2023visual}, M6Doc, XFUND~\cite{xu-etal-2022-xfund}, ICDAR2023-SVRD~\cite{yu2023icdar2023competitionstructured}, and a private dataset of photographed receipts.

\noindent\textbf{Key Information Mapping.} The data sources for the key information mapping task include FUNSD and POIE.

\noindent\textbf{Handwritten Content Extraction.}
This task's data is our private data, which contains real exam paper data with student information removed and manually annotated QA pairs.

\noindent\textbf{Table Parsing.}
The images for table parsing task are selected from MMTab~\cite{DBLP:conf/acl/ZhengFSS0J024}, WTW~\cite{Long_2021_ICCV}, TabRecSet~\cite{yang2023large} and flush table recognition competition\footnote{\url{https://github.com/10jqka-aicubes/table-recognition}}.

\noindent\textbf{Chart Parsing.} The data sources for the chart parsing task come from OneChart~\cite{chen2024onechart} and MMC~\cite{liu2024mmc}.

\noindent\textbf{Document Parsing.} The data sources for document parsing task come from DoTA~\cite{liang2024document}, DocVQA~\cite{DBLP:conf/wacv/MathewKJ21}, M6Doc, and CDLA.

\noindent\textbf{Formula Recognition.} The data sources for the formula Recognition task includes HME100K~\cite{yuan2022syntax}, IM2LATEX-100K~\cite{DBLP:journals/corr/abs-2402-14804}, M2E~\cite{DBLP:conf/mm/YangLPJHY23}, MathWriting~\cite{DBLP:journals/corr/abs-2404-10690}, MLHME-38K\footnote{\url{https://ai.100tal.com/icdar}}, CASIA-CSDB~\cite{9956654}, and some private data.

\noindent\textbf{Math QA.} The data sources for the math QA task includes MathMatics~\cite{DBLP:conf/iclr/SaxtonGHK19}, MathVerse~\cite{DBLP:conf/eccv/ZhangJZLGQZLCQGL24}, MathVision~\cite{DBLP:journals/corr/abs-2402-14804}, and MathVista~\cite{DBLP:conf/iclr/LuBX0LH0CG024}.

\noindent\textbf{Text Counting.}
The data for the text counting task are collected from IIIT5K, SVT, ICDAR2013, HierText, and TotalText.

\noindent\textbf{Cognition VQA.} The data sources for the cognition VQA task include 
EST-VQA~\cite{wang2020general}, OCRVQA~\cite{mishra2019ocr}, ST-VQA~\cite{biten2019scene}, TEXTVQA, DIR300~\cite{feng2022geometric}, ChartQA~\cite{masry2022chartqa}, DVQA~\cite{kafle2018dvqa}, PlotQA~\cite{Methani_2020_WACV}, InfoVQA~\cite{DBLP:journals/corr/abs-2104-12756}, WTW, PubTabNet~\cite{zhong2019image}, WTQ~\cite{pasupat-liang-2015-compositional}, CORD~\cite{park2019cord}, 
LLaVAR~\cite{zhang2023llavar}, WebSRC~\cite{chen2021websrc}, DocVQA, M6Doc, XFUND, Publaynet~\cite{DBLP:conf/icdar/ZhongTJ19}, RVL-CDIP~\cite{harley2015icdar}, ScreenQA~\cite{baechler2024screenai}, SlideVQA~\cite{SlideVQA2023}, a movie poster collection dataset\footnote{\url{https://www.kaggle.com/datasets/neha1703/movie-genre-from-its-poster}}, a website screenshot collection dataset\footnote{\url{https://huggingface.co/datasets/Zexanima/website_screenshots_image_dataset/tree/main}}, and a private receipt photograph dataset.

\noindent\textbf{Diagram QA.} The data sources for the diagram QA task include AI2D~\cite{kembhavi2016diagram} and TextBookQA~\cite{kembhavi2017you}.

\noindent\textbf{Document Classification.}
The images for the document classification task are collected from RVL-CDIP.

\noindent\textbf{Reasoning VQA.} The reasoning VQA task shares some common data sources with the cognition VQA task. Additionally, portions of the reasoning VQA dataset are drawn from MMSI~\cite{lee2024modeling} and CMMMU~\cite{zhang2024cmmmuchinesemassivemultidiscipline}.

\noindent\textbf{Science QA.}
The images and annotations of the science QA task are collected from ScienceQA~\cite{lu2022learn} and MMMU-Pro~\cite{yue2024mmmu}

\noindent\textbf{APP Agent.}
The data source of the APP agent task is RICO.

\noindent\textbf{ASCII Art Classification.} The data sources for the ASCII art classification task is ASCIIEval~\cite{jia2024visual}.

\noindent\textbf{Text Translation.} The datasets collected for text translation task includes memes\footnote{\url{https://www.kaggle.com/datasets/dvishal485/meme-challenge?resource=download}}, MSRA-TD500~\cite{yao2012detecting}, MTWI2018~\cite{8546143}, M6Doc, ICDAR2023-SVRD, EST-VQA, RCTW17~\cite{shi2018icdar2017competitionreadingchinese}, DAST1500~\cite{TANG2019106954}, XFUND, ArT2019~\cite{chng2019icdar2019}, ChartQA, CDLA, ICDAR2015, SlideVQA, Fintabnet~\cite{zheng2020global}, ScienceQA, InfoVQA, COMICS-Dialogue\footnote{\url{https://huggingface.co/datasets/lmms-lab/M4-Instruct-Data}}, and ExpressExpense SRD\footnote{\url{https://expressexpense.com/blog/free-receipt-images-ocr-machine-learning-dataset/}}.

\subsection{Task Definitions}
\label{sec:task}

In this section, we introduce the definition of each task, and the visualizations for each task can be found in Sec.~\ref{sec:sample_for_task}.

\noindent\textbf{Text Recognition.} Text recognition refers to the fundamental OCR ability on text image patches, which asks LMMs to read the text content. To comprehensively evaluate LMMs' text recognition ability across diverse scenarios, our collection incorporates various text types, including regular text, irregular text, artistic text, handwriting text, digit string text, non-semantic text, occluded text, doc matrix text, and CAPTCHA text.

\noindent\textbf{Fine-grained Text Recognition.} This task requires LLMs to read and comprehend textual content within the given region. It evaluates LLMs' fine-grained perception capabilities in understanding text in natural scenes and documents.

\noindent\textbf{Full-page OCR.} Full-page OCR~\cite{liu2024focus} task requires LMMs to extract and recognize all text content from the given images. Converting text into digital format facilitates subsequent processing and analysis of text images.

\noindent\textbf{Text grounding.} In this task, users would provide a text string and require LMMs to locate its specific location, evaluating LMMs' fine-grained perception capabilities.

\noindent\textbf{VQA with Position.} For VQA with position task, LMMs need to not only respond to the question but also provide the exact position coordinates that directly correspond to the answer. We ask LMMs to output both information in JSON format for convenient evaluation, and the coordinates are required to be normalized with image sizes and scaled to the range of $[0, 1000]$.

\noindent\textbf{Text Spotting.} Text spotting task needs LMMs to output the localization and content of all appeared text simultaneously. Due to the interference of background elements and the large number of text instances, this task demands high fine-grained perception capabilities from the model. Besides, the coordinates are required to be normalized with image sizes and scaled to the range of $[0, 1000]$.

\begin{table*}
\centering
\small
\renewcommand{\arraystretch}{0.9}
\setlength{\tabcolsep}{3pt}
\addtolength{\tabcolsep}{2pt}
  \centering
  \fontsize{9}{12}\selectfont
  \begin{tabular}{@{}c c | c c | c c@{}}
    \toprule
    \textbf{Scene} & \textbf{Number} & \textbf{Scene} & \textbf{Number} & \textbf{Scene} & \textbf{Number} \\
    \midrule
    Schematic diagram & 1238 & Scientific paper & 799 & Word & 728 \\
    Table(filled) & 705 & Chart & 620 & Receipts & 609 \\
    Questions & 581 & Mathematical formula & 475 & Product labels & 434 \\
    Phone screenshot & 431 & Indoor scenes & 395 & Industry research reports & 343 \\
    Poster & 264 & Street scene & 224 & ASCII Art & 199 \\
    Shop sign & 189 & Financial reports & 153 & Chemical formula & 149 \\
    Textbook & 148 & Magazine & 146 & Email & 111 \\
    Web screenshot & 99 & Details page & 95 & Verification code & 87 \\
    Resumes & 67 & Illustration & 61 & Newspaper & 52 \\
    Road signs & 43 & Menus & 31 & Notify & 30 \\
    Questionnaire & 29 \\
    \bottomrule
  \end{tabular}
  \vspace{4pt}
  \caption{The number of images included in each scene category in public data.}
  \label{tab:scene_count}
\end{table*}


\begin{figure}[t!]
  \centering   \includegraphics[width=0.95\linewidth]{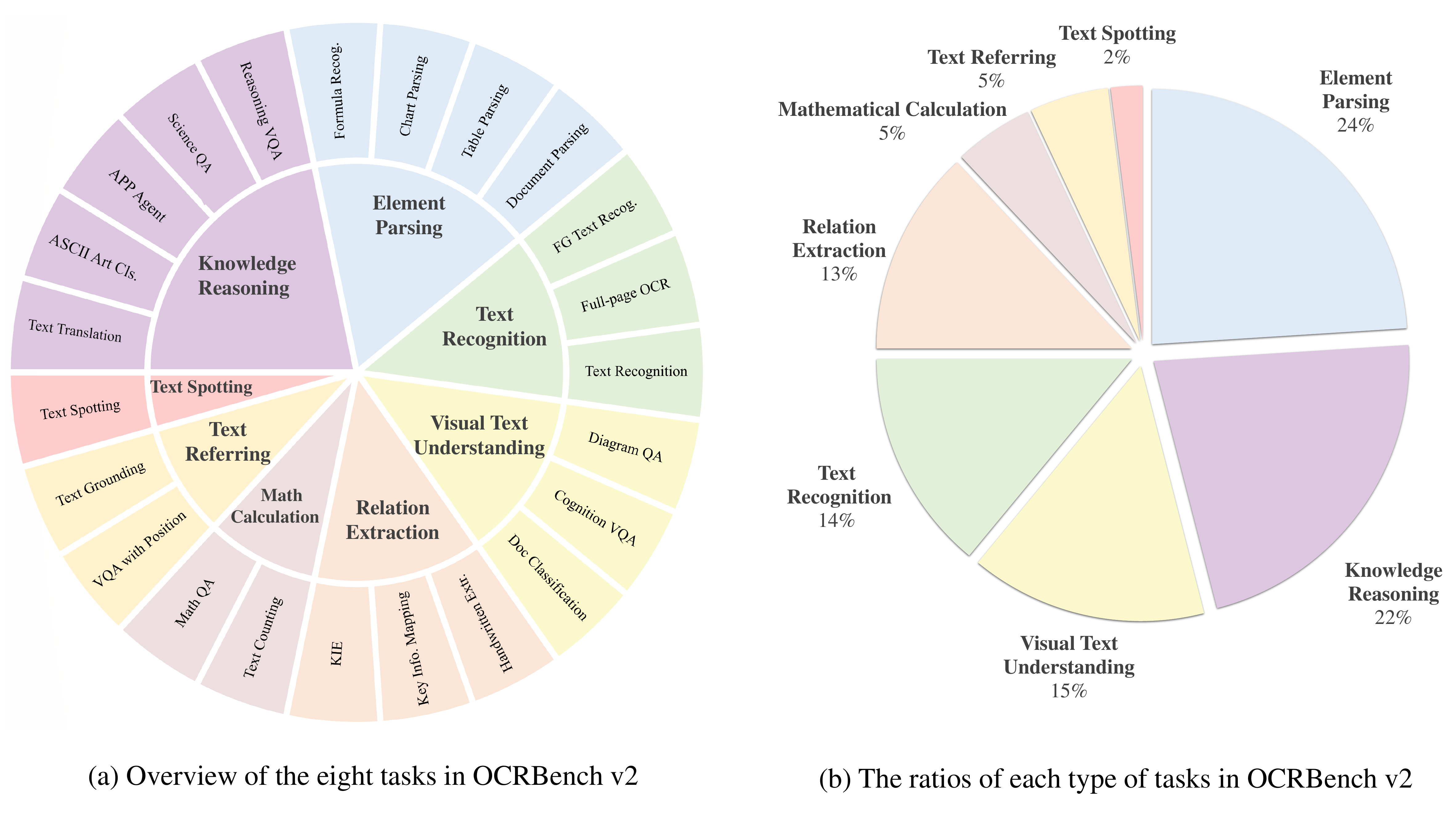}
   \captionsetup{width=0.95\linewidth}
   \vspace{-5pt}
   \caption{\textbf{Overview of the eight testable text-reading capabilities and associated tasks in \datasetname}. Each color represents a distinct capability type.}
   \label{fig:ocrbench-tasks}
   \vspace{-10pt}
\end{figure}

\begin{figure*}[t]
  \centering   
  \includegraphics[width=\linewidth]{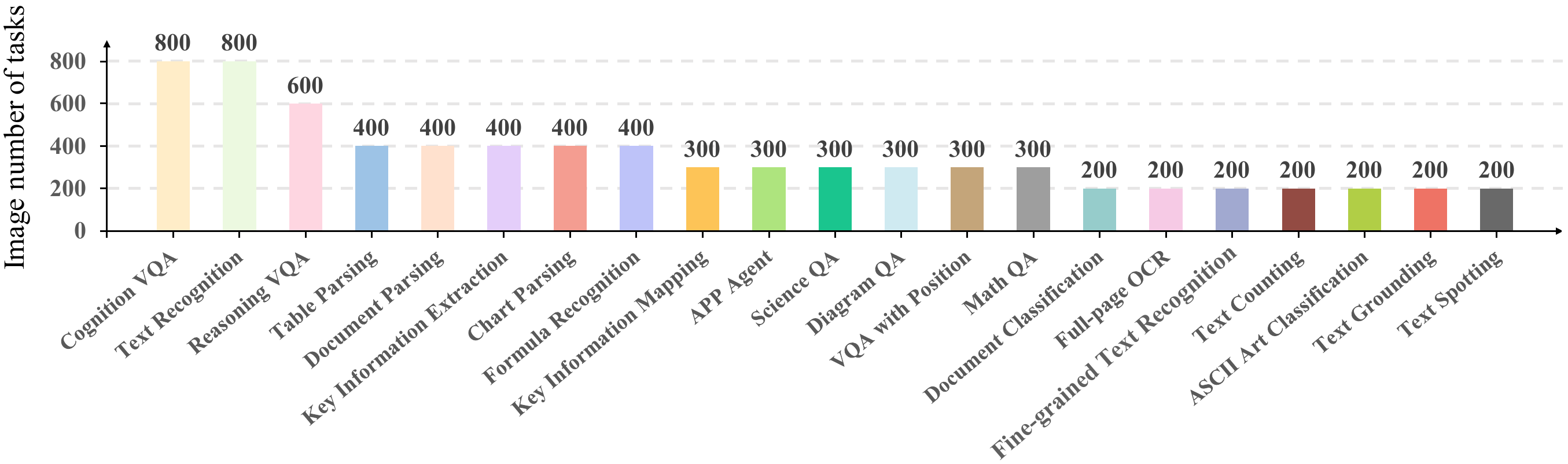}
\vspace{-5pt}
  \caption{The quantity distribution of English tasks of public data.}
   \label{fig:Eng_sub_tasks_num}
   \vspace{-10pt}
\end{figure*}

\begin{figure}[ht]
  \centering   
  \includegraphics[width=0.95\linewidth]{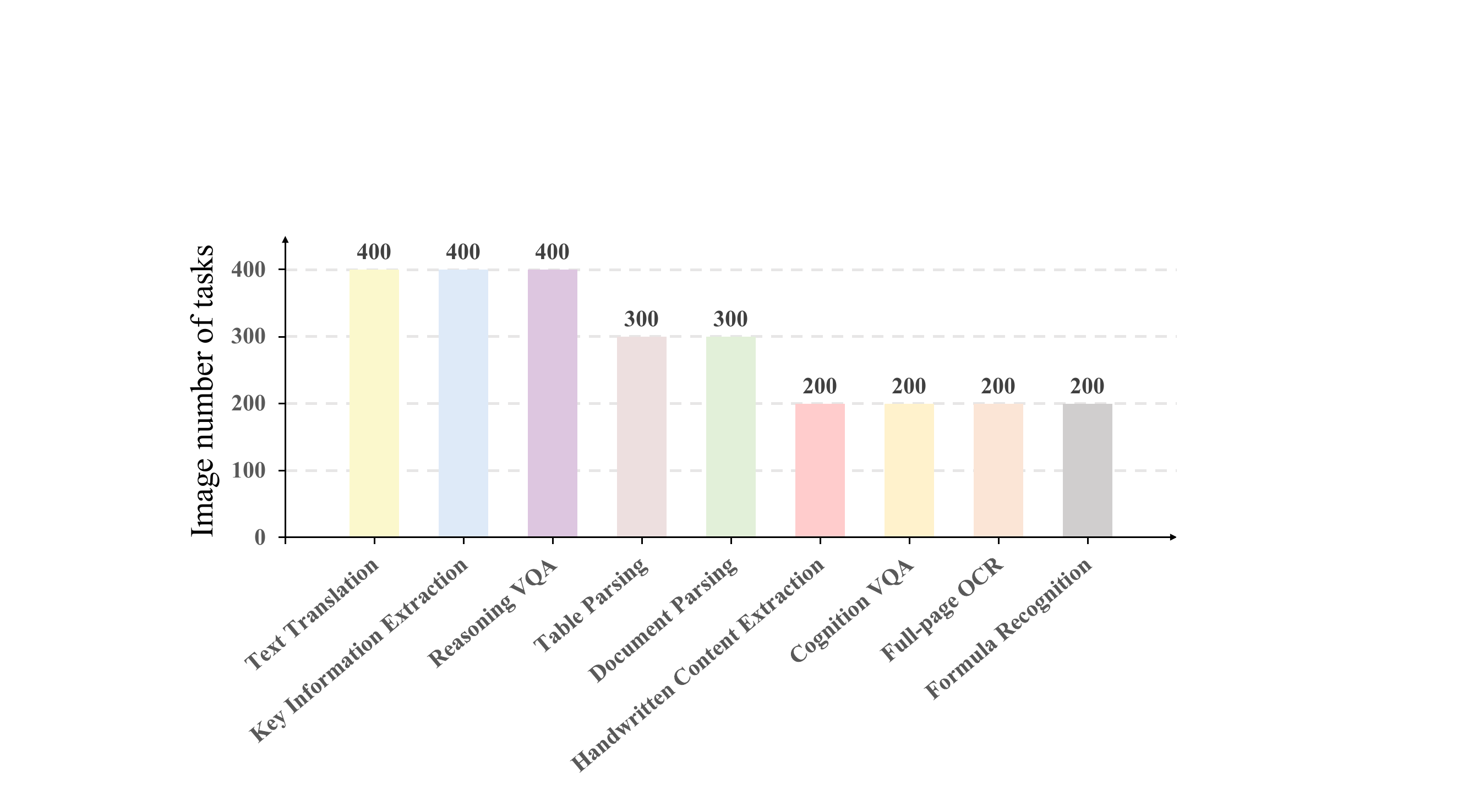}
  \vspace{-5pt}
  \caption{The quantity distribution of Chinese tasks of public data.}
   \label{fig:Chi_sub_tasks_num}
   \vspace{-10pt}
\end{figure}

\begin{figure*}[t]
  \centering   
  \includegraphics[width=\linewidth]{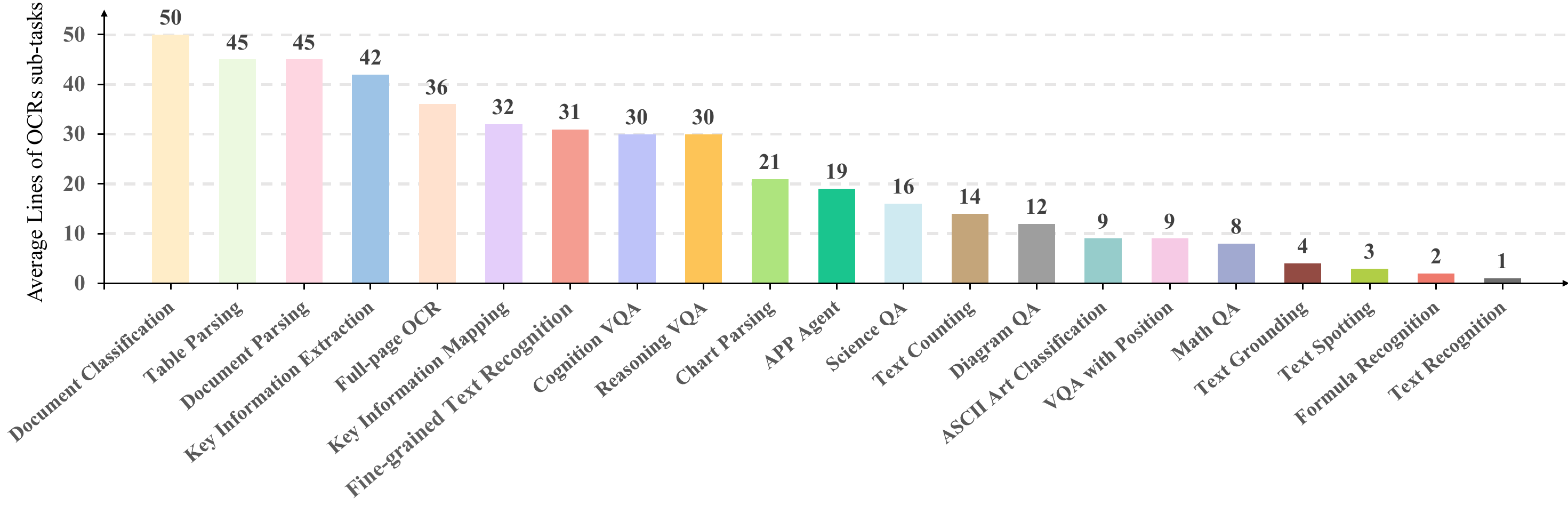}
  \vspace{-5pt}
  \caption{The OCR lines distribution of English tasks of public data.}
   \label{fig:Eng_sub_tasks_ocr}
   \vspace{-10pt}
\end{figure*}

\begin{figure}[t]
  \centering   
  \includegraphics[width=0.95\linewidth]{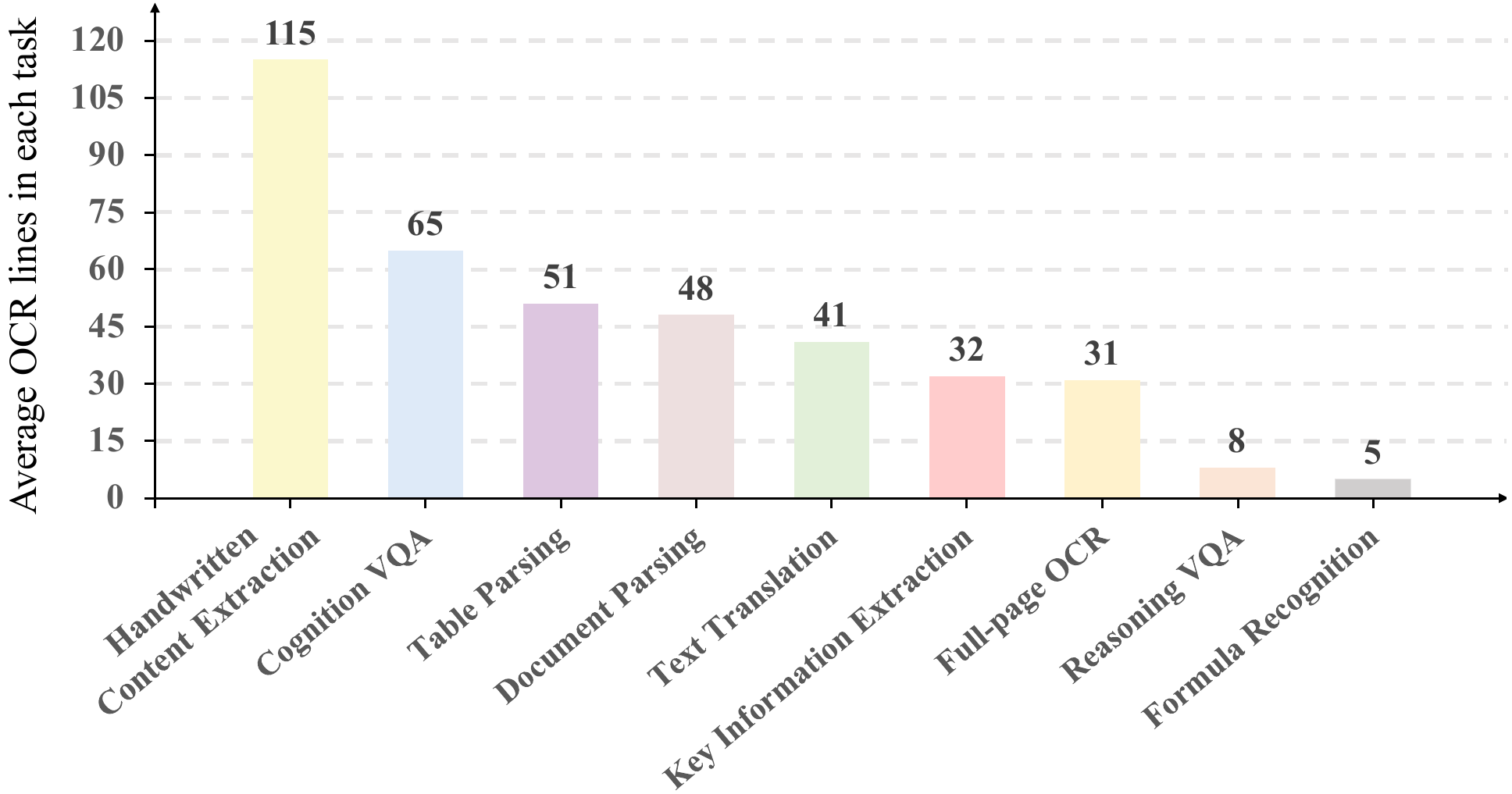}
   \vspace{-5pt}
  \caption{The OCR lines distribution of Chinese tasks of public data.}
   \label{fig:Chi_sub_tasks_ocr}
   \vspace{-10pt}
\end{figure}

\noindent\textbf{Key Information Extraction.} The key information extraction task is to extract the necessary information from densely arranged text. In this task, we provide some desired entities as keys and demand LMMs to output the corresponding values to form the output JSON string.  

\noindent\textbf{Key Information Mapping.} In this task, we provide a set of entity keys and their corresponding values in the prompt. The LMMs are then asked to match and pair these keys with their respective values into groups.

\noindent\textbf{Handwritten Content Extraction.} To investigate the information extraction capabilities of LMMs in educational scenarios, we collect some Chinese examination papers, containing both printed question text and handwritten student responses. There are four types of questions in these examination papers, including single-choice, multiple-choice, true or false, and brief response questions. The prompts require LMMs to extract the handwritten content for specific questions.

\noindent\textbf{Table Parsing.} Table parsing task requires LMMs to parse the given table into structured text, including Markdown and HTML format.

\noindent\textbf{Chart Parsing.} Apart from tables, charts can also be converted to structured information. In this task, LLMs are required to transform visual charts into JSON format.

\noindent\textbf{Document Parsing.} In the document parsing task, both text and the complex elements, including charts, tables, and formulas, are required to be parsed.

\noindent\textbf{Formula Recognition.} This task asks LMMs to recognize the given formula in the LaTeX format. The collection includes mathematical and chemical formulas. 

\noindent\textbf{Math QA.} Math QA task evaluates the LMMs' mathematical calculation ability. In particular, we render the mathematical problem description and related figures into images and ask LMMs to answer the questions within the images.

\noindent\textbf{Text Counting.} Text counting task is built to evaluate the quantity property perceiving ability of LMMs, including the character frequency in words and the word counting in the given image.

\noindent\textbf{Cognition VQA.} In \textit{OCRBench v2}, we split text-centric VQA instructions into cognition VQA and Reasoning VQA based on whether the answers can be directly found in the images. Cognition VQA task refers to the instructions where answers are explicitly present in the given image. This task evaluates the fundamental text-centric question-answering ability based on visual content.

\noindent\textbf{Diagram QA.} In the diagram QA task, LMMs need to respond to the question about the given diagrams, reflecting LMMs' ability to understand the relationship between the visual elements.

\noindent\textbf{Document Classification.} Document classification task asks LMMs to classify the category of the given document image. The included categories are letters, forms, emails, handwritten documents, advertisements, scientific reports, scientific publications, specifications, file folders, news articles, budgets, invoices, presentations, questionnaires, resumes, and memos.

\noindent\textbf{Reasoning VQA.} In reasoning VQA tasks, the answers often do not directly appear in the image. This forces LMMs to perform logical reasoning to respond to questions based on visual information.

\noindent\textbf{Science QA.} In the Science QA task, LMMs are required to respond to the scientific problem. We use PaddleOCR\footnote{\url{https://github.com/PaddlePaddle/PaddleOCR}} to extract text from the collected images and filter out those with fewer than four OCR results. Additionally, when extra subject-related knowledge is provided by the source, we incorporate it by rendering it into the images.

\noindent\textbf{APP Agent.} For the APP agent task, LMMs need to understand the relationship between textual content, icons, and world knowledge to respond to the question from the user, simulating the real-world application scene.

\noindent\textbf{ASCII Art Classification.} We incorporate a recent image classification task that uses images composed purely of ASCII characters~\cite{jia2024visual}. This task is included in \textit{OCRBench v2} to evaluate LMMs' ability to assess LMMs' pattern recognition and visual abstraction abilities.

\noindent\textbf{Text Translation.} In the text translation task, LMMs need to execute translation between Chinese and English texts, evaluating LMMs' semantic understanding abilities.

\subsection{Additional Statistics of OCRBench v2}
\label{more_Statistics}

\noindent\textbf{Scene Coverage.} Our dataset can be divided into $31$ classic scenes according to the scene of the image. The specific scenes and the corresponding number of pictures are shown in Tab.~\ref{tab:scene_count}.

\noindent\textbf{Statistics of each task.} Fig.~\ref{fig:ocrbench-tasks} shows an overview of each task in \textit{OCRBench v2}.The distribution of $23$ tasks in \textit{OCRBench v2} is displayed in Fig.~\ref{fig:Eng_sub_tasks_num} and Fig.~\ref{fig:Chi_sub_tasks_num}. Additionally, we calculate and present the average number of OCR text lines per task in Fig.~\ref{fig:Eng_sub_tasks_ocr} and Fig.~\ref{fig:Chi_sub_tasks_ocr}. As illustrated in these figures, the task distribution is well-balanced, with each task containing adequate textual information for analysis.

\subsection{Evaluation Metrics}
\label{metric}

\noindent\textbf{Parsing Type.} 
We use Tree-Edit-Distance-based Similarity (TEDS)~\cite{zhong2020image} to evaluate parsing tasks, which require LMMs to transform the images to structured formats. Tree Edit Distance (TED) refers to the minimum number of edits to transform one tree into another. TEDS is based on TED to calculate the similarity of two trees. Assuming $T_1$ and $T_2$ are two different trees, $TED(T_1, T_2)$ refers to their TED, and the TEDS is defined as:
\begin{equation}
TEDS(T_1, T_2) = 1 - \frac{TED(T_1, T_2)}{\max(|T_1|, |T_2|)},
\end{equation}

\noindent where $|T_1|$ and $|T_2|$ is the number of nodes of trees, $TED(T_1, T_2)$ can be calculated by dynamic programming algorithm. If $T_1$ and $T_2$ are identical, then their TEDS equals $1$. As the structural difference between two trees increases, their TED value becomes larger, resulting in the TEDS approaching $0$.

\noindent\textbf{Localization Type.} 
In the text referring and spotting tasks, LMMs are required to provide regression bounding boxes of target objects. IoU score is adopted to measure the distance between the predicted regions and the ground truth.
\begin{equation}
IoU(B_1, B_2)=\frac{Intersect(B_1,B_2)}{Union(B_1, B_2)},
\end{equation}

\noindent where $Intersect(B_1,B_2)$ refers to the overlap area of bounding box $B_1$ and $B_2$, while $Union(B_1, B_2)$ refers to their union area.

\noindent\textbf{Extraction Type.} 
The F1 score is used to evaluate LMMs' relation extraction capability. Given the predicted and ground truth Key-Value pairs, the F1 score is formulated as follows:

\begin{align}
&Precision = \frac{N_3}{N_2},  \label{eq:precision}\\
&Recall = \frac{N_3}{N_1}, \label{eq:recall}\\
&Fmean = \frac{2*Precision*Recall}{Precision + Recall}, \label{eq:fmean}
\end{align}

\noindent where $N_1$, $N_2$, and $N_3$ denote the number of ground-truth Key-Value pairs, predicted Key-Value pairs, and correctly matched Key-Value pairs, respectively.

\noindent\textbf{Long Reading Type.} 
To evaluate LMMs' ability to recognize text across entire paragraphs or pages, BLEU~\cite{papineni2002bleu}, METEOR~\cite{banerjee2005meteor}, F1 score, and normalized edit distance are employed. And the final score is the average value of these metrics.

BLEU evaluates prediction quality by comparing n-gram match rates between the prediction and ground truth sequences. For each n-gram type, precision is calculated as the ratio of matching n-grams to total predicted n-grams. The final BLEU score is the geometric mean of these precision values multiplied by a penalty $BP$, which is defined as:

\begin{align}
&BLEU = BP * exp(\sum^N_{n=1} {w_n \log{p_n}}), \\
&BP = \begin{cases} 
      1 & L_p \geq L_g \\
      e^{(1 - \frac{L_p}{L_g})} & L_p < L_g
   \end{cases},
\end{align}

\noindent where $p_n$ represents the precision of n-grams, $L_p$ represents the length of prediction sequence, $L_g$ represents the length of ground truth sequence, $w_n$ is weight factor, usually evenly distributed ($w_n = \frac{1}{N}$). Typically, $N$ is set to 4.

METEOR employs a semantic-aware matching strategy with four levels. 
1) Exact Match: words in the prediction that are identical to the ground truth. 
2) Stem match: matching words that have the same word stem. 
3) Synonym Match: matching words based on synonymous relationships. 
4) Paraphrase Match: Matching similar phrases at the phrase level. 
These matches are combined to calculate precision and recall, from which a weighted harmonic mean F1 score is derived as:

\begin{align}
&P_{meteor} = \frac{N_{match}}{N_{pred}}, \\
&R_{meteor}=\frac{N_{match}}{N_{gt}}, \\
&F_{meteor}=\frac{10 * P_{meteor} * R_{meteor}}{P_{meteor} + 9*R_{meteor}},
\end{align}

\noindent where $N_{match}$, $N_{pred}$, and $N_{gt}$ represent the number of matched items, words in prediction, and words in ground truth, respectively. The final METEOR score is obtained by multiplying the $F_{meteor}$ by the penalty adjustment factor. The calculation is formulated as follows:
\begin{align}
&METEOR = F_{meteor} * (1-BP_{meteor}), \\
&BP_{meteor} = 0.5* \frac{N_{chunk}}{N_{match}},
\end{align}

\noindent where $N_{chunk}$ refers to the number of contiguous matching phrases. More chunks indicate greater word order differences, resulting in a heavier penalty.

The calculation method of the F1 score in long reading metrics follows the same approach as discussed in extraction metrics, as shown in Equations~\ref{eq:precision}, \ref{eq:recall}, \ref{eq:fmean}.

Normalized Edit Distance (NED) measures string similarity by computing the minimum number of operations needed to transform one string into another. And then NED is normalized by the length of the longer string. The calculation is formulated as follows:

\begin{align}
NED(S_1, S_2)=\frac{ED(S_1, S_2)}{\text{max}(len(S_1), len(S_2))}
\end{align}

\noindent where $ED(S_1, S_2)$ represents the edit distance between the prediction string $S_1$ and the ground truth $S_2$. The $NED$ value of 0 indicates identical strings, while 1 indicates completely different strings.

\noindent\textbf{Counting Type.}
In \textit{OCRBench v2}, character frequency counting and word counting tasks are included. For character frequency, we use exact match evaluation since the answers are typically single-digit integers. For word counting, we evaluate using the L1 distance between predicted and ground truth counts, normalized to $[0,1]$ based on the ground truth. This can be formulated as follows:

\begin{align}
score = \begin{cases} 
      0 & C_{pred} \leq 0 \\
      1 - \frac{|C_{pred}-C_{gt}|}{C_{gt}}) & 0 < C_{pred} < 2 * C_{gt} \\
      0 & C_{pred} \geq 2 * C_{gt} \\
   \end{cases},
\end{align}

\noindent where $C_{pred}$ and $C_{gt}$ denote the predicted count and ground truth count, respectively.

\noindent\textbf{Basic VQA Type.}
The remaining tasks in \textit{OCRBench v2} are basic VQA types, and we employ different evaluation metrics based on question types. For multiple-choice questions, we use exact matching between predictions and answer options. In other cases, we check whether the ground truth is contained in the prediction for answers shorter than $5$ words, and use ANLS for longer answers.

\subsection{Experimental setting}
\label{Experimental setting}
The detailed public data construction are shown in Sec.~\ref{data_collection} and Sec.~\ref{metric}. Private data consists of unlabeled images collected manually from websites and real life. At the same time, we annotated and checked the private test set to ensure the quality. The environment configuration of each open-source model experiment strictly complies with the official version and uses the official pre-trained model and inference code. The model parameters of the open-source model and the API parameters of the closed-source model use the official default parameters for fair. Specifically, we use the official API versions: GPT-4o (gpt-4o-2024-08-06), GPT-4o-mini (gpt-4o-mini-2024-07-18), and Gemini 1.5 Pro (gemini1.5-pro-002).

\subsection{Compute resources}
\label{Experiments compute resources}
Evaluations of open-source models were conducted on 8×NVIDIA GeForce RTX 4090 (24GB) and a NVIDIA H800 Tensor Core GPU (80GB). The closed-source experiments obtained the results by calling the official API.

\subsection{Results and Discussions}
\label{sec:Evaluation results}


Tab.~\ref{tab:all_English_subsets}, Tab.~\ref{tab:all_Chinese_subsets}, Tab.~\ref{tab:all_private_English_subsets}, and Tab.~\ref{tab:all_private_Chinese_subsets} exhibit the results of $39$ open-source models and $5$ closed-source models on the public and private test sets of \textit{OCRBench v2}

\noindent\textbf{Evaluation results on public data} are shown in Tab.~\ref{tab:all_English_subsets} and Tab.~\ref{tab:all_Chinese_subsets}. Most LMMs performed well in tasks such as Understanding, Recognition, Extraction, which shows that current models have basic OCR capabilities. However, they performed poorly in tasks such as Referring, Spotting, Parsing, and Calculation. The scores of all models are basically below 50 points, which shows that the models still lack the ability in text localization, logical reasoning, and understanding complex elements.

\noindent\textbf{Evaluation results on private data} are shown in Tab.~\ref{tab:all_private_English_subsets} and Tab.~\ref{tab:all_private_Chinese_subsets}. The performance trends of the models on private and public datasets are consistent. In addition, most models perform worse on private datasets than on public datasets, which shows that private data may be more challenging for LMMs due to the lack of training, and also reflects the importance of private data construction.

\begin{table*}
  \centering
  \renewcommand{\arraystretch}{0.9} 
  \addtolength{\tabcolsep}{-2pt} 
  \fontsize{9}{11}\selectfont
  \resizebox{\textwidth}{!}{
  \begin{tabular}{ 
                  @{}
                  l
                  >{\centering\arraybackslash}p{1.25cm}
                  >{\centering\arraybackslash}p{1.25cm}
                  >{\centering\arraybackslash}p{1.25cm}
                  >{\centering\arraybackslash}p{1.25cm}
                  >{\centering\arraybackslash}p{1.25cm}
                  >{\centering\arraybackslash}p{1.25cm}
                  >{\centering\arraybackslash}p{1.25cm}
                  >{\centering\arraybackslash}p{1.25cm}
                  >{\centering\arraybackslash}p{1.25cm}
                  @{}
                  }
    \toprule
    {\fontsize{7.5}{9}\selectfont Method} & 
    \multicolumn{1}{>{\centering\arraybackslash}p{1.25cm}}{\fontsize{7.5}{9}\selectfont Recognition} & 
    \multicolumn{1}{>{\centering\arraybackslash}p{1.25cm}}{\fontsize{7.5}{9}\selectfont Referring} & 
    \multicolumn{1}{>{\centering\arraybackslash}p{1.25cm}}{\fontsize{7.5}{9}\selectfont Spotting} &{\fontsize{8}{9}\selectfont Extraction} & 
    \multicolumn{1}{>{\centering\arraybackslash}p{1.25cm}}{\fontsize{7.5}{9}\selectfont Parsing} & 
    \multicolumn{1}{>{\centering\arraybackslash}p{1.25cm}}{\fontsize{7.5}{9}\selectfont Calculation} & 
    \multicolumn{1}{>{\centering\arraybackslash}p{1.25cm}}{\fontsize{7.5}{9}\selectfont Understanding} & 
    \multicolumn{1}{>{\centering\arraybackslash}p{1.25cm}}{\fontsize{7.5}{9}\selectfont Reasoning} & 
    \multicolumn{1}{>{\centering\arraybackslash}p{1.25cm}}{\fontsize{7.5}{9}\selectfont Average} \\
    \midrule
    \rowcolor{gray!20}
    \multicolumn{10}{c}{Open-source LMMs} \\
    LLaVA-Next-8B~\cite{liu2024llava} & 41.3 & 18.8 & 0 & 49.5 & 21.2 & 17.3 & 55.2 & 48.9 & 31.5 \\
    LLaVA-OV-7B~\cite{li2024llava} & 46.0 & 20.8 & 0.1 & 58.3 & 25.3 & 23.3 & 64.4 & 53.0 & 36.4 \\
    Monkey~\cite{li2024monkey} & 35.2 & 0 & 0 & 16.6 & 16.3 & 14.4 & 59.8 & 42.3 & 23.1 \\
    TextMonkey~\cite{liu2024textmonkey} & 39.1 & 0.7 & 0 & 19.0 & 12.2 & 19.0 & 61.1 & 40.2 & 23.9 \\
    XComposer2-4KHD~\cite{dong2024internlm} & 45.1 & 21.8 & 0.1 & 15.9 & 11.7 & 15.7 & 66.8 & 45.9 & 27.9 \\
    Molmo-7B~\cite{deitke2024molmo} & 52.4 & 21.3 & 0.1 & 45.5 & 7.6 & 28.5 & 65.3 & 55.0 & 34.5 \\
    Cambrian-1-8B~\cite{tongcambrian} & 45.3 & 21.5 & 0 & 53.6 & 19.2 & 19.5 & 63.5 & 55.5 & 34.7 \\
    Pixtral-12B~\cite{agrawal2024pixtral} & 48.9 & 21.6 & 0 & 66.3 & 35.5 & 29.8 & 66.9 & 53.7 & 40.3 \\
    EMU2-chat~\cite{sun2024generative} & 42.1 & 0.2 & 0 & 12.5 & 8.1 & 11.2 & 42.7 & 33.4 & 18.8 \\
    mPLUG-Owl3~\cite{ye2024mplug} & 41.6 & 14.0 & 0.6 & 24.4 & 10.9 & 11.1 & 52.2 & 46.0 & 25.1 \\
    CogVLM-chat~\cite{wang2023cogvlm} & 50.9 & 0 & 0 & 0.2 & 8.4 & 15.0 & 58.1 & 41.7 & 21.8 \\
    Qwen-VL~\cite{bai2023qwen} & 34.6 & 7.5 & 0 & 18.2 & 20.0 & 8.1 & 57.2 & 41.1 & 23.3 \\
    Qwen-VL-chat~\cite{bai2023qwen} & 34.5 & 4.1 & 0 & 25.9 & 14.0 & 13.8 & 55.7 & 39.5 & 23.4 \\
    Qwen2-Vl-7B~\cite{wang2024qwen2} & \underline{72.1} & \textbf{47.9} & \textbf{17.5} & 82.5 & 25.5 & 25.4 & 78.4 & 61.5 & 51.4 \\
    Qwen2.5-VL-7B~\cite{qwen2.5-VL} & 68.8 & 25.7 & 1.2 & 80.2 & 30.4 & 38.2 & 73.2 & 56.2 & 46.7 \\
    InternVL2-8B~\cite{chen2024internvl} & 49.9 & 23.1 & 0.5 & 65.2 & 24.8 & 26.7 & 73.5 & 52.9 & 39.6 \\
    InternVL2-26B~\cite{chen2024internvl} & 63.4 & 26.1 & 0 & 76.8 & 37.8 & 32.3 & \textbf{79.4} & 58.9 & 46.8 \\
    InternVL2.5-8B~\cite{chen2024expanding} & 59.0 & 25.0 & 1.4 & 77.5 & 35.1 & 29.4 & 75.3 & 57.2 & 45.0 \\
    InternVL2.5-26B~\cite{chen2024expanding} & 65.6 & 26.1 & 1.6 & \underline{86.9} & 36.2 & 37.4 & 78.3 & \textbf{62.9} & 49.4 \\
    InternVL3-8B~\cite{chen2024expanding} & 68.6 & 30.4 & 8.8 & 85.3 & 34.0 & 27.1 & 77.5 & 60.3 & 49.0 \\
    InternVL3-14B~\cite{chen2024expanding} & 67.3 & 36.9 & 11.2 & \textbf{89.0} & 38.4 & 38.4 & \underline{79.2} & 60.5 & \textbf{52.6} \\
    Deepseek-VL-7B~\cite{lu2024deepseek} & 37.1 & 15.4 & 0 & 23.5 & 14.6 & 20.8 & 53.3 & 52.9 & 27.2 \\
    Deepseek-VL2-Small~\cite{wu2024deepseek} & 62.7 & 28.0 & 0.1 & 77.5 & 32.7 & 14.3 & 77.1 & 53.9 & 43.3 \\
    MiniCPM-V-2.6~\cite{yao2024minicpm} & 66.8 & 6.0 & 0.8 & 62.0 & 28.8 & 32.4 & 73.7 & 52.1 & 40.3 \\
    MiniCPM-o-2.6~\cite{yao2024minicpm} & 66.9 & 29.5 & 0.5 & 70.8 & 33.4 & 31.9 & 69.9 & 57.9 & 45.1 \\
    GLM-4V-9B~\cite{glm2024chatglm} & 61.8 & 22.6 & 0 & 71.7 & 31.6 & 22.6 & 72.1 & 58.4 & 42.6 \\
    VILA1.5-8B~\cite{liu2024nvila} & 35.3 & 15.5 & 0 & 21.1 & 12.7 & 17.3 & 46.3 & 40.3 & 23.6 \\
    LLaVAR~\cite{zhang2023llavar} & 37.3 & 0 & 0 & 1.0 & 9.9 & 12.3 & 34.6 & 27.0 & 15.3 \\
    UReader~\cite{ye2023ureader} & 22.4 & 0.1 & 0 & 0 & 9.2 & 7.9 & 41.0 & 29.1 & 13.7 \\
    DocOwl2~\cite{hu2024mplug2} & 24.0 & 9.7 & 0 & 13.4 & 13.5 & 8.8 & 53.7 & 32.0 & 19.4 \\
    Yi-VL-6B~\cite{young2024yi} & 28.9 & 2.9 & 0 & 9.7 & 12.9 & 15.8 & 36.1 & 32.0 & 17.3 \\
    Janus-1.3B~\cite{wu2024janus} & 46.1 & 0 & 0 & 0.2 & 14.5 & 13.5 & 36.0 & 39.1 & 18.7 \\
    Eagle-X5-7B~\cite{shi2024eagle} & 34.7 & 17.8 & 0 & 21.7 & 20.6 & 21.5 & 61.0 & 42.6 & 27.5 \\
    Idefics3-8B~\cite{laurenccon2024building}  & 23.8 & 13.2 & 0 & 63.2 & 23.8 & 23.0 & 65.8 & 44.9 & 32.2 \\
    Phi-4-MultiModal~\cite{abouelenin2025phi} & 63.7 & 16.4 & 0 & 40.4 & 19.1 & 18.3 & 69.8 & 53.9 & 35.2 \\
    SAIL-VL-1.6-8B~\cite{duan2024vlmevalkit} & 67.7 & 28.6 & 2.8 & 70.5 & 25.9 & 29.5 & 73.9 & 59.7 & 44.8 \\
    Kimi-VL-A3B-16B~\cite{kimiteam2025kimivltechnicalreport} & 56.5 & 13.8 & 0 & 59.2 & 33.8 & 32.9 & 75.5 & 56.7 & 41.1 \\
    Ovis1.6-3B~\cite{lu2024ovis}  & 59.2 & 14.3 & 0 & 65.0 & 32.1 & 29.0 & 69.8 & 56.8 & 40.8 \\
    Ovis2-8B~\cite{lu2024ovis} & \textbf{73.2} & 24.6 & 0.7 & 62.4 & \textbf{44.8} & 40.6 & 72.7 & \underline{62.6} & 47.7 \\
    \rowcolor{gray!20}
    \multicolumn{10}{c}{Closed-source LMMs} \\
    GPT-4o~\cite{achiam2023gpt} & 61.2 & 26.7 & 0 & 77.5 & 36.3 & \underline{43.4} & 71.1 & 55.5 & 46.5 \\
    GPT-4o-mini~\cite{gpt4omini} & 57.9 & 23.3 & 0.6 & 70.8 & 31.5 & 38.8 & 65.9 & 55.1 & 43.0 \\
    Gemini-Pro~\cite{team2023gemini} & 61.2 & \underline{39.5} & \underline{13.5} & 79.3 & \underline{39.2} & \textbf{47.7} & 75.5 & 59.3 & \underline{51.9} \\
    Claude3.5-sonnet~\cite{Anthropic2024Claude3.5Sonnet} & 62.2 & 28.4 & 1.3 & 56.6 & 37.8 & 40.8 & 73.5 &  60.9 & 45.2 \\
    Step-1V~\cite{website_step1v} & 67.8 & 31.3 & 7.2 & 73.6 & 37.2 & 27.8 & 69.8 & 58.6 & 46.7 \\
    \bottomrule
  \end{tabular}
  }
  \caption{\textbf{Evaluation of existing LMMs on English tasks of OCRBench v2's public data}. ``Recognition'', ``Referring'', ``Spotting'', ``Extraction'', ``Parsing'', ``Calculation'', ``Understanding'', and ``Reasoning'' refer to text recognition, text referring, text spotting, relation extraction, element parsing, mathematical calculation, visual text understanding, and knowledge reasoning, respectively. Higher values indicate better performance. Best performance is in boldface, and the second best is underlined. The notations apply to all subsequent figures.}
  \label{tab:all_English_subsets}
  \vspace{-10pt}
\end{table*}

\begin{table*}
  \centering
  \renewcommand{\arraystretch}{0.85} 
  \addtolength{\tabcolsep}{-0.5pt} 
  \fontsize{9}{11}\selectfont
  \resizebox{\textwidth}{!}{
  \begin{tabular}{
                  @{}
                  l
                  >{\centering\arraybackslash}p{1.5cm}
                  >{\centering\arraybackslash}p{1.5cm}
                  >{\centering\arraybackslash}p{1.25cm}
                  >{\centering\arraybackslash}p{1.25cm}
                  >{\centering\arraybackslash}p{1.25cm}
                  >{\centering\arraybackslash}p{1.25cm}
                  >{\centering\arraybackslash}p{1.25cm}
                  >{\centering\arraybackslash}p{1.25cm}
                  @{}
                  }
    \toprule
    Method & 
    \multicolumn{1}{>{\centering\arraybackslash}p{1.5cm}}{\fontsize{8}{9}\selectfont LLM Size} & 
    \multicolumn{1}{>{\centering\arraybackslash}p{1.25cm}}{\fontsize{8}{9}\selectfont Recognition} & 
    \multicolumn{1}{>{\centering\arraybackslash}p{1.25cm}}{\fontsize{8}{9}\selectfont Extraction} & 
    \multicolumn{1}{>{\centering\arraybackslash}p{1.25cm}}{\fontsize{8}{9}\selectfont Parsing} & 
    \multicolumn{1}{>{\centering\arraybackslash}p{1.25cm}}{\fontsize{8}{9}\selectfont Understanding} & 
    \multicolumn{1}{>{\centering\arraybackslash}p{1.25cm}}{\fontsize{8}{9}\selectfont Reasoning} & 
    \multicolumn{1}{>{\centering\arraybackslash}p{1.25cm}}{\fontsize{8}{9}\selectfont Average} \\
    \midrule
    \rowcolor{gray!20}
    \multicolumn{8}{c}{Open-source LMMs} \\
    LLaVA-Next-8B~\cite{liu2024llava} & 8B & 5.7 & 2.9 & 12.2 & 7.5 & 17.2 & 9.1 \\
    LLaVA-OV-7B~\cite{li2024llava} & 8B & 14.8 & 15.7 & 13.7 & 16.0 & 28.7 & 17.8 \\
    Monkey~\cite{li2024monkey} & 8B & 4.6 & 11.2 & 8.4 & 21.5 & 20.0 & 13.1 \\
    TextMonkey~\cite{liu2024textmonkey} & 8B &  23.5 & 14.8 & 8.4 & 19.9 & 12.2 & 15.8 \\
    XComposer2-4KHD~\cite{dong2024internlm} & 7B & 16.7 & 18.8 & 12.1 & 27.5 & 2.3 & 15.5 \\
    Molmo-7B~\cite{deitke2024molmo} & 8B & 7.1 & 15.0 & 9.2 & 9.0 & 23.7 & 12.8 \\
    Cambrian-1-8B~\cite{tongcambrian} & 8B & 5.3 & 14.9 & 12.6 & 8.5 & 8.1 & 9.9 \\
    Pixtral-12B~\cite{agrawal2024pixtral} & 12B & 13.4 & 10.9 & 21.0 & 7.0 & 20.7 & 14.6 \\
    EMU2-chat~\cite{sun2024generative} & 37B & 2.3 & 0.5 & 8.5 & 1.0 & 7.3 & 3.9 \\
    mPLUG-Owl3~\cite{ye2024mplug} & 8B & 6.6 & 17.9 & 9.7 & 6.0 & 26.1 & 13.3 \\
    CogVLM-chat~\cite{wang2023cogvlm} & 7B & 5.5 & 10.0 & 9.8 & 1.5 & 2.5 & 5.9 \\
    Qwen-VL~\cite{bai2023qwen} & 8B & 7.2 & 5.3 & 10.7 & 11.5 & 11.2 & 9.2 \\
    Qwen-VL-chat~\cite{bai2023qwen} & 8B & 9.5 & 8.2 & 9.3 & 11.0 & 21.1 & 11.8 \\
    Qwen2-Vl-7B~\cite{wang2024qwen2} & 7B & 51.3 & 51.4 & 21.6 & 52.5 & 37.5 & 42.9 \\
    Qwen2.5-VL-7B~\cite{qwen2.5-VL} & 7B & \textbf{75.3} & 61.4 & \textbf{41.8} & \underline{59.3} & 40.4 & \underline{55.6} \\
    InternVL2-8B~\cite{chen2024internvl} & 8B & 20.6 & 45.2 & 23.2 & 54.4 & 38.1 & 36.3 \\
    InternVL2-26B~\cite{chen2024internvl} & 26B & 21.9 & 46.0 & 34.8 & 50.9 & 34.8 & 37.7 \\
    InternVL2.5-8B~\cite{chen2024expanding} & 8B & 52.8 & 52.8 & 28.6 & 56.4 & 40.5 & 46.2 \\
    InternVL2.5-26B~\cite{chen2024expanding} & 26B & 32.4 & 56.1 & 32.6 & 56.3 & 43.6 & 44.2 \\
    InternVL3-8B~\cite{chen2024expanding} & 8B & 68.9 & \underline{62.0} & 31.6 & 57.9 & \underline{47.3} & 53.5 \\ 
    InternVL3-14B~\cite{chen2024expanding} & 14B & 66.2 & \textbf{64.8} & 33.5 & \textbf{63.4} & \textbf{50.6} & \textbf{55.7} \\
    Deepseek-VL-7B~\cite{lu2024deepseek} & 7B & 8.0 & 13.3 & 15.7 & 5.5 & 18.5 & 12.2 \\
    Deepseek-VL2-Small~\cite{wu2024deepseek} & 16B & 60.9 & 50.6 & 28.3 & 53.0 & 20.5 & 42.7 \\
    MiniCPM-V-2.6~\cite{yao2024minicpm} & 8B & 51.0 & 29.9 & 21.2 & 34.0 & 33.6 & 33.9 \\
    MiniCPM-o-2.6~\cite{yao2024minicpm} & 7B & 53.0 & 49.4 & 27.1 & 43.5 & 32.7 & 41.1 \\
    GLM-4V-9B~\cite{glm2024chatglm} & 9B & 24.4 & 60.6 & 20.4 & 52.8 & 25.2 & 36.6 \\
    VILA1.5-8B~\cite{liu2024nvila} & 8B & 5.4 & 8.8 & 8.5 & 3.0 & 15.5 & 8.2 \\
    LLaVAR~\cite{zhang2023llavar} & 13B & 2.3 & 1.7 & 8.9 & 0 & 2.5 & 3.1 \\
    UReader~\cite{ye2023ureader} & 7B & 6.8 & 2.7 & 8.4 & 2.5 & 7.2 & 5.5 \\
    DocOwl2~\cite{hu2024mplug2} & 7B & 4.2 & 10.3 & 8.6 & 4.0 & 9.6 & 7.3 \\
    Yi-VL-6B~\cite{young2024yi} & 6B & 4.8 & 4.4 & 8.5 & 4.0 & 25.0 & 9.4 \\
    Janus-1.3B~\cite{wu2024janus} & 1.3B & 7.6 & 8.7 & 11.4 &  4.5 & 10.7 & 8.6 \\
    Eagle-X5-7B~\cite{shi2024eagle} & 8B & 7.5 & 12.0 & 11.6 & 5.0 & 19.2 & 11.1 \\
    Idefics3-8B~\cite{laurenccon2024building} & 8B & 7.0 & 15.5 & 15.9 & 9.0 & 18.1 & 13.1 \\
    Phi-4-MultiModal~\cite{abouelenin2025phi}  & 5.6B & 51.5 & 32.3 & 12.1 & 34.4 & 23.0 & 30.7 \\
    SAIL-VL-1.6-8B~\cite{duan2024vlmevalkit} & 8B & 31.2 & 40.0 & 23.9 & 42.3 & 35.0 & 34.5 \\
    Kimi-VL-A3B-16B~\cite{kimiteam2025kimivltechnicalreport}& 16B & 57.2 & 54.7 & 31.5 & 52.5 & 31.4 & 45.5 \\
    Ovis1.6-3B~\cite{lu2024ovis} & 3B & 11.5 & 23.7 & 22.8 & 28.8 & 18.9 & 21.1 \\
    Ovis2-8B~\cite{lu2024ovis} & 7B & \underline{72.2} & 50.8 & \underline{37.7} & 47.9 & 37.4 & 49.2 \\
    \rowcolor{gray!20}
    \multicolumn{8}{c}{Closed-source LMMs} \\
    GPT-4o~\cite{achiam2023gpt} & - & 21.6 & 53.0 & 29.8 & 38.5 & 18.2 & 32.2 \\
    GPT-4o-mini~\cite{gpt4omini} & - & 13.1 & 38.9 & 27.2 & 28.8 & 16.9 & 25.0 \\
    Gemini-Pro~\cite{team2023gemini} & - & 52.5 & 47.3 & 30.9 & 51.5 & 33.4 & 43.1 \\
    Claude3.5-sonnet~\cite{Anthropic2024Claude3.5Sonnet} & - & 21.0 & 56.2 & 35.2 & 55.0 & 30.5 & 39.6 \\
    Step-1V~\cite{website_step1v} & - & 56.7 & 41.1 & 37.6 & 38.3 & 39.2 & 42.6 \\
    \bottomrule
  \end{tabular}
  }
  \caption{\textbf{Evaluation of existing LMMs on Chinese tasks of OCRBench v2' public data}. ``LLM Size'' indicates the number of parameters of the language model employed in each method.}
  \label{tab:all_Chinese_subsets}
  \vspace{-5pt}
\end{table*}

\begin{table*}
  \centering
  \renewcommand{\arraystretch}{0.9} 
  \addtolength{\tabcolsep}{-2pt} 
  \fontsize{9}{11}\selectfont
  \resizebox{\textwidth}{!}{
  \begin{tabular}{ 
                  @{}
                  l
                  >{\centering\arraybackslash}p{1.25cm}
                  >{\centering\arraybackslash}p{1.25cm}
                  >{\centering\arraybackslash}p{1.25cm}
                  >{\centering\arraybackslash}p{1.25cm}
                  >{\centering\arraybackslash}p{1.25cm}
                  >{\centering\arraybackslash}p{1.25cm}
                  >{\centering\arraybackslash}p{1.25cm}
                  >{\centering\arraybackslash}p{1.25cm}
                  >{\centering\arraybackslash}p{1.25cm}
                  @{}
                  }
    \toprule
    {\fontsize{7.5}{9}\selectfont Method} & 
    \multicolumn{1}{>{\centering\arraybackslash}p{1.25cm}}{\fontsize{7.5}{9}\selectfont Recognition} & 
    \multicolumn{1}{>{\centering\arraybackslash}p{1.25cm}}{\fontsize{7.5}{9}\selectfont Referring} & 
    \multicolumn{1}{>{\centering\arraybackslash}p{1.25cm}}{\fontsize{7.5}{9}\selectfont Spotting} &{\fontsize{8}{9}\selectfont Extraction} & 
    \multicolumn{1}{>{\centering\arraybackslash}p{1.25cm}}{\fontsize{7.5}{9}\selectfont Parsing} & 
    \multicolumn{1}{>{\centering\arraybackslash}p{1.25cm}}{\fontsize{7.5}{9}\selectfont Calculation} & 
    \multicolumn{1}{>{\centering\arraybackslash}p{1.25cm}}{\fontsize{7.5}{9}\selectfont Understanding} & 
    \multicolumn{1}{>{\centering\arraybackslash}p{1.25cm}}{\fontsize{7.5}{9}\selectfont Reasoning} & 
    \multicolumn{1}{>{\centering\arraybackslash}p{1.25cm}}{\fontsize{7.5}{9}\selectfont Average} \\
    \midrule
    \rowcolor{gray!20}
    \multicolumn{10}{c}{Open-source LMMs} \\
    LLaVA-Next-8B~\cite{liu2024llava} & 41.4 & 17.0 & 0 & 49.0 & 12.9 & 16.1 & 60.9 & 30.5 & 28.5 \\
    LLaVA-OV-7B~\cite{li2024llava} & 45.4 & 18.5 & 0 & 60.0 & 15.5 & 32.0 & 59.0 & 39.3 & 33.7 \\
    Monkey~\cite{li2024monkey} & 31.5 & 0.1 & 0 & 34.4 & 26.3 & 17.7 & 61.4 & 22.4 & 24.2 \\
    TextMonkey~\cite{liu2024textmonkey} & 39.8 & 1.6 & 0 & 27.6 & 24.8 & 10.2 & 62.3 & 21.2 & 23.4 \\
    XComposer2-4KHD~\cite{dong2024internlm} & 39.5 & 12.0 & 0 & 69.7 & 26.0 & 20.2 & 68.2 & 35.8 & 33.9 \\
    Molmo-7B~\cite{deitke2024molmo} & 40.8 & 19.5 & 0 & 51.7 & 10.0 & 33.9 & 67.0 & 48.0 & 33.9 \\
    Cambrian-1-8B~\cite{tongcambrian} & 44.0 & 19.0 & 0 & 52.3 & 19.0 & 20.7 & 64.0 & 39.3 & 32.3 \\
    Pixtral-12B~\cite{agrawal2024pixtral} & 45.1 & 21.8 & 0 & 71.6 & 21.7 & 30.4 & 77.3 & 39.5 & 38.4 \\
    EMU2-chat~\cite{sun2024generative} & 34.3 & 0 & 0 & 20.4 & 21.3 & 20.3 & 47.1 & 18.3 & 20.2 \\
    mPLUG-Owl3~\cite{ye2024mplug} & 34.9 & 17.0 & 0 & 12.0 & 14.9 & 24.1 & 50.7 & 25.5 & 22.4 \\
    CogVLM-chat~\cite{wang2023cogvlm} & 40.8 & 0 & 0 & 1.6 & 18.6 & 10.9 & 60.2 & 26.8 & 19.9 \\
    Qwen-VL~\cite{bai2023qwen} & 35.9 & 4.2 & 0 & 38.7 & 28.5 & 13.8 & 60.1 & 16.9 & 24.8 \\
    Qwen-VL-chat~\cite{bai2023qwen} & 34.1 & 12.6 & 0.1 & 42.6 & 19.5 & 18.4 & 58.3 & 20.3 & 25.7 \\
    Qwen2-Vl-7B~\cite{wang2024qwen2} & 47.0 & \textbf{42.0} & 1.5 &  \textbf{90.2} & 13.7 & 36.4 & 71.1 & 36.6 & 42.3 \\
    Qwen2.5-VL-7B~\cite{wang2024qwen2} & 51.5 & 24.5 & \underline{3.1} & 64.8 & 13.1 & 53.3 & \underline{78.6} & 45.5 & 41.8 \\
    InternVL2-8B~\cite{chen2024internvl} & 43.0 & 21.6 & 0 & 70.2 & 19.2 & 35.6 & 65.9 & 33.6 & 36.1 \\
    InternVL2-26B~\cite{chen2024internvl} & 56.0 & 21.2 & 0 & 80.5 & 23.9 & 40.3 & 72.1 & 40.7 & 41.8 \\
    InternVL2.5-8B~\cite{chen2024expanding} & 48.9 & 21.2 & 0 & 82.1 & 20.3 & 41.2 & 67.8 & 42.3 & 40.5 \\
    InternVL2.5-26B~\cite{chen2024expanding} & 53.5 & 21.4 & 0 & 84.0 & 21.4 & 51.5 & 67.5 & 41.5 & 42.6 \\
    InternVL3-8B~\cite{chen2024expanding} & 49.7 & 22.3 & 0.2 & 86.8 & 22.4 & 57.0 & 70.7 & 53.0 & 45.3 \\
    InternVL3-14B~\cite{chen2024expanding} & 55.8 & 24.5 & 2.1 & 89.3 & 21.0 & \underline{59.5} & 72.0 & 50.0 & 46.8  \\
    Deepseek-VL-7B~\cite{lu2024deepseek} & 33.5 & 13.7 & 0 & 19.1 & 11.7 & 24.8 & 60.5 & 32.5 & 24.5 \\
    Deepseek-VL2-Small~\cite{wu2024deepseek} & 56.6 & 23.7 & 0 & 86.4 & 18.9 & 30.6 & 72.2 & 39.5 & 41.0 \\
    MiniCPM-V-2.6~\cite{yao2024minicpm} & 52.2 & 18.6 & 0.3 & 45.8 & 19.6 & 20.9 & 68.9 & 37.3 & 33.0 \\
    MiniCPM-o-2.6~\cite{yao2024minicpm} & 54.1 & 24.7 & 0.3 & 74.4 & 17.6 & 39.2 & 75.7 & 47.0 & 41.6 \\
    GLM-4v-9B~\cite{glm2024chatglm} & 52.7 & 20.6 & 0 & 79.4 & 15.9 & 21.5 & 74.7 & 32.0 & 37.1 \\
    VILA1.5-8B~\cite{liu2024nvila} & 36.0 & 14.5 & 0 & 26.0 & 17.4 & 20.3 & 44.7 & 27.0 & 23.2 \\
    LLaVAR~\cite{zhang2023llavar} & 13.8 & 0 & 0 & 8.3 & 15.2 & 4.4 & 42.4 & 15.0 & 12.4 \\
    UReader~\cite{ye2023ureader} & 20.9 & 0 & 0 & 0 & 20.7 & 11.3 & 39.0 & 20.8 & 14.1 \\
    DocOwl2~\cite{hu2024mplug2} & 25.4 & 7.5 & 0 & 47.1 & 26.2 & 8.3 & 52.8 & 19.5 & 23.4 \\
    Yi-VL-6B~\cite{young2024yi} & 31.1 & 4.0 & 0 & 23.4 & 22.5 & 18.1 & 43.0 & 15.5 & 19.7 \\
    Janus-1.3B~\cite{wu2024janus} & 32.6 & 0 & 0 & 0.3 & 13.0 & 18.4 & 32.1 & 17.9 & 14.3 \\
    Eagle-X5-7B~\cite{shi2024eagle} & 34.6 & 18.5 & 0 & 9.7 & 18.5 & 24.0 & 63.1 & 37.0 & 25.7 \\
    Idefics3-8B~\cite{laurenccon2024building} & 37.4 & 13.0 & 0 & 28.9 & 19.4 & 21.1 & 65.4 & 21.8 & 26.0 \\
    Phi-4-MultiModal~\cite{abouelenin2025phi} & 58.4 & 19.0 & 0 & 53.5 &  \textbf{38.7} & 28.7 & 66.8 & 39.8 & 38.1 \\
    SAIL-VL-1.6-8B~\cite{duan2024vlmevalkit} & 56.7 & 24.1 & 2.2 & 79.3 & 22.8 & 45.4 & 69.2 & 45.3 & 43.1 \\
    Kimi-VL-A3B-16B~\cite{kimiteam2025kimivltechnicalreport} & 49.1 & 13.5 & 0 & 28.8 & 21.9 & 37.6 & 69.4 & 36.2 & 32.1 \\
    Ovis1.6-3B~\cite{lu2024ovis} & 48.5 & 19.5 & 0 & 69.2 & 20.7 & 22.1 & 74.6 & 49.5 & 38.0 \\
    Ovis2-8B~\cite{lu2024ovis} & 54.2 & 20.9 & 0 & 83.6 & 24.2 & 54.7 & 74.1 & 57.3 & 46.1 \\
    \rowcolor{gray!20}
    \multicolumn{10}{c}{Closed-source LMMs} \\
    GPT-4o~\cite{achiam2023gpt} & \underline{58.6} & 23.4 & 0 & 87.4 & 23.1 & 51.6 & 74.4 & \textbf{62.3} & \underline{47.6} \\
    GPT-4o-mini~\cite{gpt4omini} & 55.3 & 21.8 & 0 & 85.4 & 20.6 & 45.2 & 75.5 & 49.0 & 44.1 \\
    Gemini1.5-Pro~\cite{team2023gemini} & \textbf{59.1} & \underline{41.2} & \textbf{6.6} & \underline{89.5} & 22.4 & 54.7 & \textbf{78.8} & \underline{60.3} & \textbf{51.6} \\
    Claude3.5-sonnet~\cite{Anthropic2024Claude3.5Sonnet} & 52.9 & 24.9 & 2.5 & 86.9 & 23.8 & \textbf{61.4} & 74.4 & 53.0 & 47.5 \\
    Step-1V~\cite{website_step1v} & 56.7 & 27.4 & 2.6 & 86.3 & \underline{33.3} & 42.6 & 76.6 & 48.7 & 46.8 \\
    \bottomrule
  \end{tabular}
  }
  \caption{\textbf{Evaluation of existing LMMs on English tasks of OCRBench v2's private data}.}
  \label{tab:all_private_English_subsets}
  \vspace{-5pt}
\end{table*}

\begin{table*}
  \centering
  \renewcommand{\arraystretch}{0.85} 
  \addtolength{\tabcolsep}{-0.5pt} 
  \fontsize{9}{11}\selectfont
  \resizebox{\textwidth}{!}{
  \begin{tabular}{
                  @{}
                  l
                  >{\centering\arraybackslash}p{1.5cm}
                  >{\centering\arraybackslash}p{1.5cm}
                  >{\centering\arraybackslash}p{1.25cm}
                  >{\centering\arraybackslash}p{1.25cm}
                  >{\centering\arraybackslash}p{1.25cm}
                  >{\centering\arraybackslash}p{1.25cm}
                  >{\centering\arraybackslash}p{1.25cm}
                  >{\centering\arraybackslash}p{1.25cm}
                  @{}
                  }
    \toprule
    Method & 
    \multicolumn{1}{>{\centering\arraybackslash}p{1.5cm}}{\fontsize{8}{9}\selectfont LLM Size} & 
    \multicolumn{1}{>{\centering\arraybackslash}p{1.25cm}}{\fontsize{8}{9}\selectfont Recognition} & 
    \multicolumn{1}{>{\centering\arraybackslash}p{1.25cm}}{\fontsize{8}{9}\selectfont Extraction} & 
    \multicolumn{1}{>{\centering\arraybackslash}p{1.25cm}}{\fontsize{8}{9}\selectfont Parsing} & 
    \multicolumn{1}{>{\centering\arraybackslash}p{1.25cm}}{\fontsize{8}{9}\selectfont Understanding} & 
    \multicolumn{1}{>{\centering\arraybackslash}p{1.25cm}}{\fontsize{8}{9}\selectfont Reasoning} & 
    \multicolumn{1}{>{\centering\arraybackslash}p{1.25cm}}{\fontsize{8}{9}\selectfont Average} \\
    \midrule
    \rowcolor{gray!20}
    \multicolumn{8}{c}{Open-source LMMs} \\
    LLaVA-Next-8B~\cite{liu2024llava} & 8B & 2.8 & 0.9 & 14.9 & 20.0 & 7.4 & 9.2 \\
    LLaVA-OV-7B~\cite{li2024llava} & 8B & 5.4 & 13.6 & 20.3 & 34.0 & 13.6 & 17.4 \\
    Monkey~\cite{li2024monkey} & 8B & 1.5 & 28.4 & 29.1 & 40.0 & 8.3 & 21.5 \\
    TextMonkey~\cite{liu2024textmonkey} & 8B & 10.5 & 15.2 & 30.2 & 44.0 & 7.6 & 21.5 \\
    XComposer2-4KHD~\cite{dong2024internlm}  & 7B & 12.9 & 38.6 & \underline{37.5} & 60.0 & 13.1 & 32.4 \\
    Molmo-7B~\cite{deitke2024molmo} & 8B & 3.4 & 29.8 & 6.6 & 24.0 & 11.1 & 15.0 \\
    Cambrian-1-8B~\cite{tongcambrian} & 8B & 2.4 & 19.8 & 26.7 & 36.0 & 7.6 & 18.5 \\
    Pixtral-12B~\cite{agrawal2024pixtral} & 12B & 6.2 & 22.3 & 11.4 & 26.0 & 14.0 & 16.0 \\
    EMU2-chat~\cite{sun2024generative} & 37B & 1.2 & 3.0 & 29.3 & 4.0 & 3.6 & 8.2 \\
    mPLUG-Owl3~\cite{ye2024mplug} & 8B & 1.6 & 27.4 & 27.3 & 16.0 & 10.0 & 16.5 \\
    CogVLM-chat~\cite{wang2023cogvlm} & 7B & 2.4 & 16.2 & 22.5 & 20.0 & 3.1 & 12.8 \\
    Qwen-VL~\cite{bai2023qwen} & 8B & 4.3 & 0 & 30.6 & 38.0 & 5.1 & 15.6 \\
    Qwen-VL-chat~\cite{bai2023qwen} & 8B & 9.1 & 3.6 & 18.9 & 44.0 & 7.1 & 16.5 \\
    Qwen2-Vl-7B~\cite{wang2024qwen2} & 7B & 23.7 & 63.5 & 27.9 & 80.0 & 28.5 & 44.7 \\
    Qwen2.5-VL-7B~\cite{wang2024qwen2} & 8B & 24.4 & \textbf{78.9} & 33.1 & \underline{82.0} & 29.0 & 49.5 \\
    InternVL2-8B~\cite{chen2024internvl} & 8B & 35.2 & 42.8 & 26.1 & 78.0 & 24.4 & 41.3 \\
    InternVL2-26B~\cite{chen2024internvl} & 26B & 20.4 & 50.7 & 29.0 & 76.0 & 14.5 & 38.1 \\
    InternVL2.5-8B~\cite{chen2024expanding} & 8B & 42.8 & 47.9 & 27.3 & 80.0 & 23.5 & 44.3 \\
    InternVL2.5-26B~\cite{chen2024expanding} & 26B & 40.2 & 42.7 & 25.6 & 74.0 & 27.0 & 41.9 \\
    InternVL3-8B~\cite{chen2024expanding} & 8B & 57.7 & 55.8 & 29.9 & 72.0 & 29.4 & 49.0 \\
    InternVL3-14B~\cite{chen2024expanding} & 14B & 62.1 & 59.5 & 33.2 & 80.0 & 29.2 & 52.8 \\
    Deepseek-VL-7B~\cite{lu2024deepseek} & 7B & 3.2 & 14.7 & 10.7 & 30.0 & 9.8 & 13.7 \\
    DeepSeek-VL2-Small~\cite{wu2024deepseek} & 16B & 51.6 & 56.3 & 27.8 & 79.6 & 25.3 & 48.1 \\
    MiniCPM-V-2.6~\cite{yao2024minicpm} & 8B & 53.1 & 53.2 & 32.8 & 76.0 & 23.4 & 47.7  \\
    MiniCPM-o-2.6~\cite{yao2024minicpm} & 7B & 54.0 & 62.4 & 24.1 & 68.0 & 29.8 & 47.7 \\
    GLM-4v-9B~\cite{glm2024chatglm} & 9B & 60.6 & 65.2 & 32.4 & \underline{82.0} & 18.2 & 51.7 \\
    VILA1.5-8B~\cite{liu2024nvila} & 8B & 1.4 & 9.1 & 22.2 & 16.0 & 6.4 & 11.0 \\
    LLaVAR~\cite{zhang2023llavar} & 13B & 2.2 & 2.0 & 27.1 & 10.0 & 1.9 & 8.6 \\
    UReader~\cite{ye2023ureader} & 7B & 0.3 & 2.0 & 28.1 & 12.0 & 2.4 & 9.0 \\
    DocOwl2~\cite{hu2024mplug2} & 7B & 1.0 & 17.8 & 29.4 & 20.0 & 3.9 & 14.4 \\
    Yi-VL-6B~\cite{young2024yi} & 6B & 1.6 & 6.4 & 28.8 & 10.0 & 5.3 & 10.4 \\
    Janus-1.3B~\cite{wu2024janus} & 1.3B & 4.1 & 2.2 & 10.4 & 14.0 & 6.7 & 7.5 \\
    Eagle-X5-7B~\cite{shi2024eagle} & 8B & 1.9 & 16.1 & 13.6 & 22.0 & 8.1 & 12.3 \\
    Idefics3-8B~\cite{laurenccon2024building} & 8B & 2.9 & 29.0 & 12.3 & 26.0 & 7.9 & 15.6 \\
    Phi-4-MultiModal~\cite{abouelenin2025phi} & 5.6B & 30.5 & 40.5 & 42.7 & 56.0 & 16.9 & 37.3 \\
    SAIL-VL-1.6-8B~\cite{duan2024vlmevalkit} & 8B & 35.8 & 41.5 & 35.7 & 76.0 & 23.9 & 42.6 \\
    Kimi-VL-A3B-16B~\cite{kimiteam2025kimivltechnicalreport}& 16B & 54.0 & \underline{71.1} & 32.5 & \textbf{84.0} & 28.7 & 54.1 \\
    Ovis1.6-3B~\cite{lu2024ovis} & 3B & 22.5 & 33.3 & 31.5 & 54.0 & 17.0 & 31.7 \\
    Ovis2-8B~\cite{lu2024ovis} & 7B & 61.0 & 67.7 & \textbf{43.6} & \underline{82.0} & 25.6 & \textbf{56.0} \\
    \rowcolor{gray!20}
    \multicolumn{8}{c}{Closed-source LMMs} \\
    GPT-4o~\cite{achiam2023gpt} & - & 41.7 & 52.1 & 29.0 & 76.0 & 29.4 & 45.7 \\
    GPT-4o-mini~\cite{gpt4omini} & - & 20.0 & 53.6 & 27.9 & 66.0 & 19.6 & 37.4 \\
    Gemini1.5-Pro~\cite{team2023gemini} & - & \textbf{71.4} & 63.8 & 30.5 & \underline{82.0} & \underline{29.9} & \underline{55.5} \\
    Claude3.5-sonnet~\cite{Anthropic2024Claude3.5Sonnet} & - & 34.2 & 62.5 & 35.2 & 78.0 & \textbf{32.2} & 48.4 \\
    Step-1V~\cite{website_step1v} & - & \underline{65.2} & 64.9 & 33.1 & 78.0 & 25.5 & 53.4 \\
    \bottomrule
  \end{tabular}
  }
  \caption{\textbf{Evaluation of existing LMMs on Chinese tasks of OCRBench v2's private data}.}
  \label{tab:all_private_Chinese_subsets}
  \vspace{-15pt}
\end{table*}

\subsection{Potential Factors Affecting OCR Capabilities}
\label{sec:Potential Factors Affecting OCR Capabilities}

\noindent\textbf{High-Res Visual Encoders.} Since text often appears small in images, the resolution setting of the visual encoder could be a key factor affecting the text perception ability~\cite{li2024monkey}. Here we change the input resolution of the LMMs and observe the performance changes. In particular, InternVL2-8B is chosen, and the resolution setting includes $448$, $896$, and dynamic. Tab.~\ref{tab:resolution} lists the results. 
Indeed, when the input resolution increases from $448$ to $896$, the performance increases by $4.1\%$.

\noindent\textbf{Pre-provided OCR Information.}
To study the impact of OCR information, we use PaddleOCR\footnote{\url{https://github.com/PaddlePaddle/PaddleOCR}} to pre-extract OCR results 
and incorporate them with prompts. 
Tab.~\ref{tab:ocr_information} shows the results. 
We observe that adding OCR information does not help much. 
This suggests that \textit{OCRBench v2} evaluates LMMs capabilities across multiple dimensions, rather than solely focusing on text recognition abilities.    

\noindent\textbf{
Connection Between OCR and LLMs.}
We further explore a direct pipeline by first extracting OCR information and then by feeding it directly into Qwen2.5. Unlike 
LMMs, this pipeline separates OCR and language modeling into distinct stages. The results shown in Tab.~\ref{tab:ocr_information} suggest that Qwen2-VL-7B outperforms Qwen2.5 with OCR information, demonstrating LMMs' remarkable ability to incorporate both textual and visual features efficiently. 

\begin{table*}
\centering
\small
  \renewcommand{\arraystretch}{0.85} 
  \addtolength{\tabcolsep}{-0.5pt} 
  \fontsize{9}{11}\selectfont
  \begin{adjustbox}{max width=\textwidth}
  \begin{tabular}{ 
                  @{}
                  l
                  >{\centering\arraybackslash}p{1.02cm}
                  >{\centering\arraybackslash}p{1.02cm}
                  >{\centering\arraybackslash}p{1.02cm}
                  >{\centering\arraybackslash}p{1.02cm}
                  >{\centering\arraybackslash}p{1.02cm}
                  >{\centering\arraybackslash}p{1.02cm}
                  >{\centering\arraybackslash}p{1.02cm}
                  >{\centering\arraybackslash}p{1.02cm}
                  >{\centering\arraybackslash}p{1.02cm}
                  >{\centering\arraybackslash}p{1.02cm}
                  @{}}
    \toprule
    Method & 
    \multicolumn{1}{>
    {\centering\arraybackslash}p{1cm}}{\fontsize{7}{9}\selectfont Resolition} & 
    \multicolumn{1}{>
    {\centering\arraybackslash}p{1cm}}{\fontsize{7}{9}\selectfont Recognition} & 
    \multicolumn{1}{>{\centering\arraybackslash}p{1cm}}{\fontsize{7}{9}\selectfont Referring} & 
    \multicolumn{1}{>{\centering\arraybackslash}p{1cm}}{\fontsize{7}{9}\selectfont Spotting} & 
    \multicolumn{1}{>{\centering\arraybackslash}p{1cm}}{\fontsize{7}{9}\selectfont Extraction} & 
    \multicolumn{1}{>{\centering\arraybackslash}p{1cm}}{\fontsize{7}{9}\selectfont Parsing} & 
    \multicolumn{1}{>{\centering\arraybackslash}p{1cm}}{\fontsize{7}{9}\selectfont Calculation} & 
    \multicolumn{1}{>{\centering\arraybackslash}p{1cm}}{\fontsize{6.8}{9}\selectfont Understanding} & 
    \multicolumn{1}{>{\centering\arraybackslash}p{1cm}}{\fontsize{6.8}{9}\selectfont Reasoning} & 
    \multicolumn{1}{>{\centering\arraybackslash}p{1cm}}{\fontsize{7}{9}\selectfont Average} \\
    \midrule
    \multirow{3}{*}{InternVL2-8B~\cite{chen2024internvl}} & 448 & 47.3 & 19.1 & \underline{0.1} & 52.8 & \textbf{27.3} & 25.4 & 61.1 & 49.1 & 35.3 \\
     & 896 & \underline{48.7} & \underline{23.0} & \textbf{0.5} & \textbf{66.2} & \underline{26.2} & \underline{25.9} & \underline{73.2} & \underline{51.9} & \underline{39.4} \\
     & dynamic & \textbf{49.9} & \textbf{23.1} & \textbf{0.5} & \underline{65.2} & 24.8 & \textbf{26.7} & \textbf{73.5} & \textbf{52.9} & \textbf{39.6} \\
    \bottomrule
  \end{tabular}
  \end{adjustbox}
  \vspace{3pt}
  \caption{Evaluation of InternVL2-8B with different resolution settings on the English tasks of OCRBench v2's public data.}
  \label{tab:resolution}
\end{table*}

\begin{table*}
  \centering
  \renewcommand{\arraystretch}{0.85} 
  \addtolength{\tabcolsep}{-0.5pt} 
  \fontsize{9}{11}\selectfont
  \begin{adjustbox}{max width=\textwidth}
  \begin{tabular}{ 
                  @{}
                  l
                  >{\centering\arraybackslash}p{1.2cm}
                  >{\centering\arraybackslash}p{1.2cm}
                  >{\centering\arraybackslash}p{1.2cm}
                  >{\centering\arraybackslash}p{1.2cm}
                  >{\centering\arraybackslash}p{1.2cm}
                  >{\centering\arraybackslash}p{1.2cm}
                  >{\centering\arraybackslash}p{1.2cm}
                  >{\centering\arraybackslash}p{1.2cm}
                  >{\centering\arraybackslash}p{1.2cm}
                  @{}}
    \toprule
    Method & 
    \multicolumn{1}{>
    {\centering\arraybackslash}p{1.2cm}}{\fontsize{7.5}{9}\selectfont Recognition} & 
    \multicolumn{1}{>{\centering\arraybackslash}p{1.2cm}}{\fontsize{7.5}{9}\selectfont Referring} & 
    \multicolumn{1}{>{\centering\arraybackslash}p{1.2cm}}{\fontsize{7.5}{9}\selectfont Spotting} & 
    \multicolumn{1}{>{\centering\arraybackslash}p{1.2cm}}{\fontsize{7.5}{9}\selectfont Extraction} & 
    \multicolumn{1}{>{\centering\arraybackslash}p{1.2cm}}{\fontsize{7.5}{9}\selectfont Parsing} & 
    \multicolumn{1}{>{\centering\arraybackslash}p{1.2cm}}{\fontsize{7.5}{9}\selectfont Calculation} & 
    \multicolumn{1}{>{\centering\arraybackslash}p{1.25cm}}{\fontsize{7.5}{9}\selectfont Understanding} & 
    \multicolumn{1}{>{\centering\arraybackslash}p{1.25cm}}{\fontsize{7.5}{9}\selectfont Reasoning} & 
    \multicolumn{1}{>{\centering\arraybackslash}p{1.25cm}}{\fontsize{7.5}{9}\selectfont Average} \\
    \midrule
    Qwen2-VL-7B~\cite{wang2024qwen2} & \textbf{72.1} & \underline{47.9} & \underline{17.5} & \textbf{82.5} & \underline{25.5} & 25.4 & \textbf{78.4} & \textbf{61.5} & \underline{51.4} \\
    Qwen2-VL-7B+OCR & \underline{69.8} & \textbf{50.4} & \textbf{20.1} & \underline{79.1} & \textbf{29.4} & \underline{28.0} & \underline{77.7} & \underline{60.0} & \textbf{51.8} \\
    Qwen2.5-8B+OCR & 28.6 & 13.8 & 0 & 45.9 & 24.2 & \textbf{31.3} & 61.1 & 40.5 & 30.7 \\
    \bottomrule
  \end{tabular}
    \end{adjustbox}
  \vspace{3pt}
  \caption{Evaluation of Qwen2-VL-7B and Qwen2.5-7B with pre-provided OCR information on English tasks of OCRBench v2's public data.}
  \label{tab:ocr_information}
  \vspace{-15pt}
\end{table*}

\subsection{Samples for Each Task}
\label{sec:sample_for_task}

\begin{figure*}[t]
  \centering   
  \includegraphics[width=0.98\linewidth]{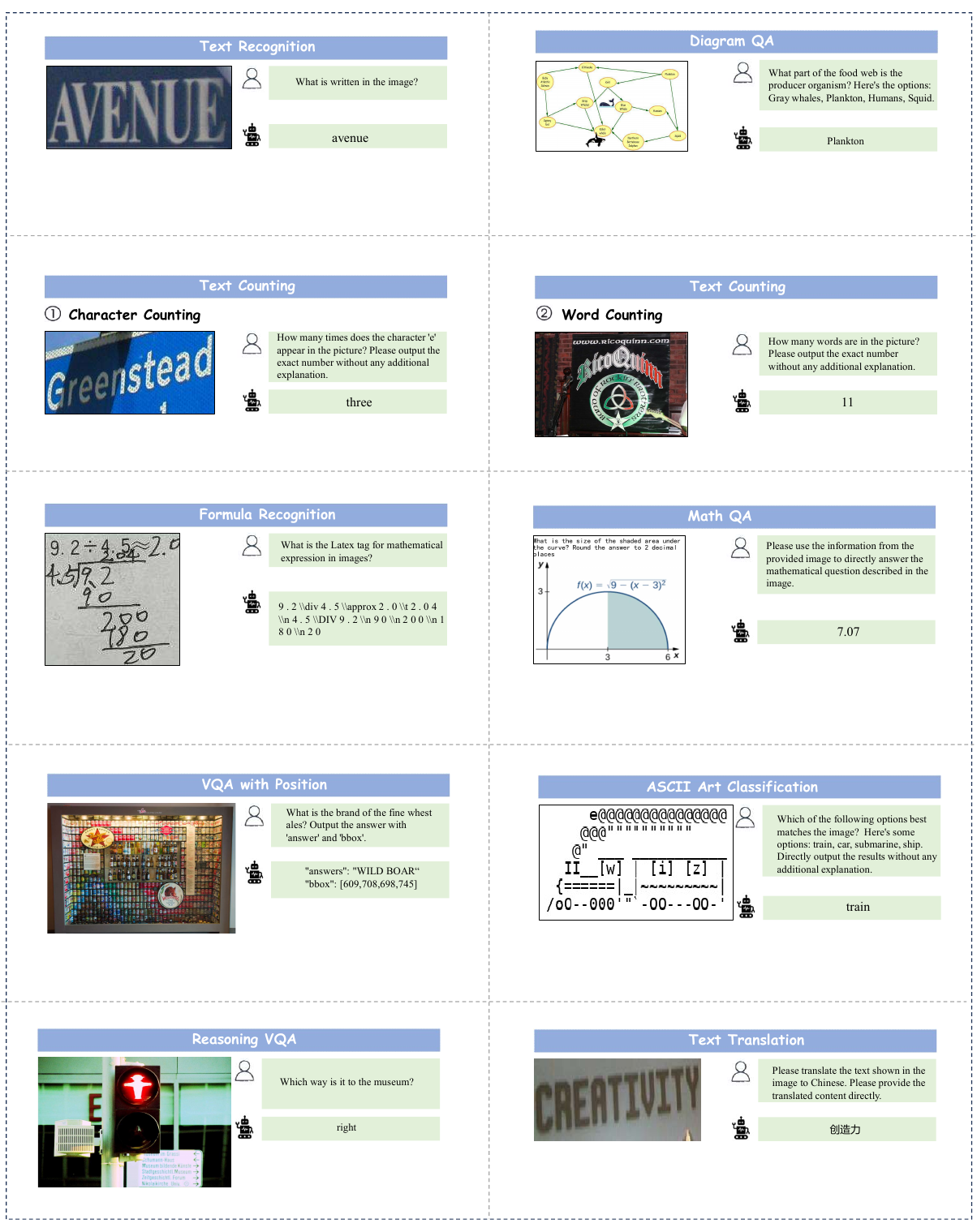}
  \caption{Samples for each task.}
   \label{fig:sample_tasks1}
\end{figure*}

\begin{figure*}[t]
  \centering   
  \includegraphics[width=0.98\linewidth]{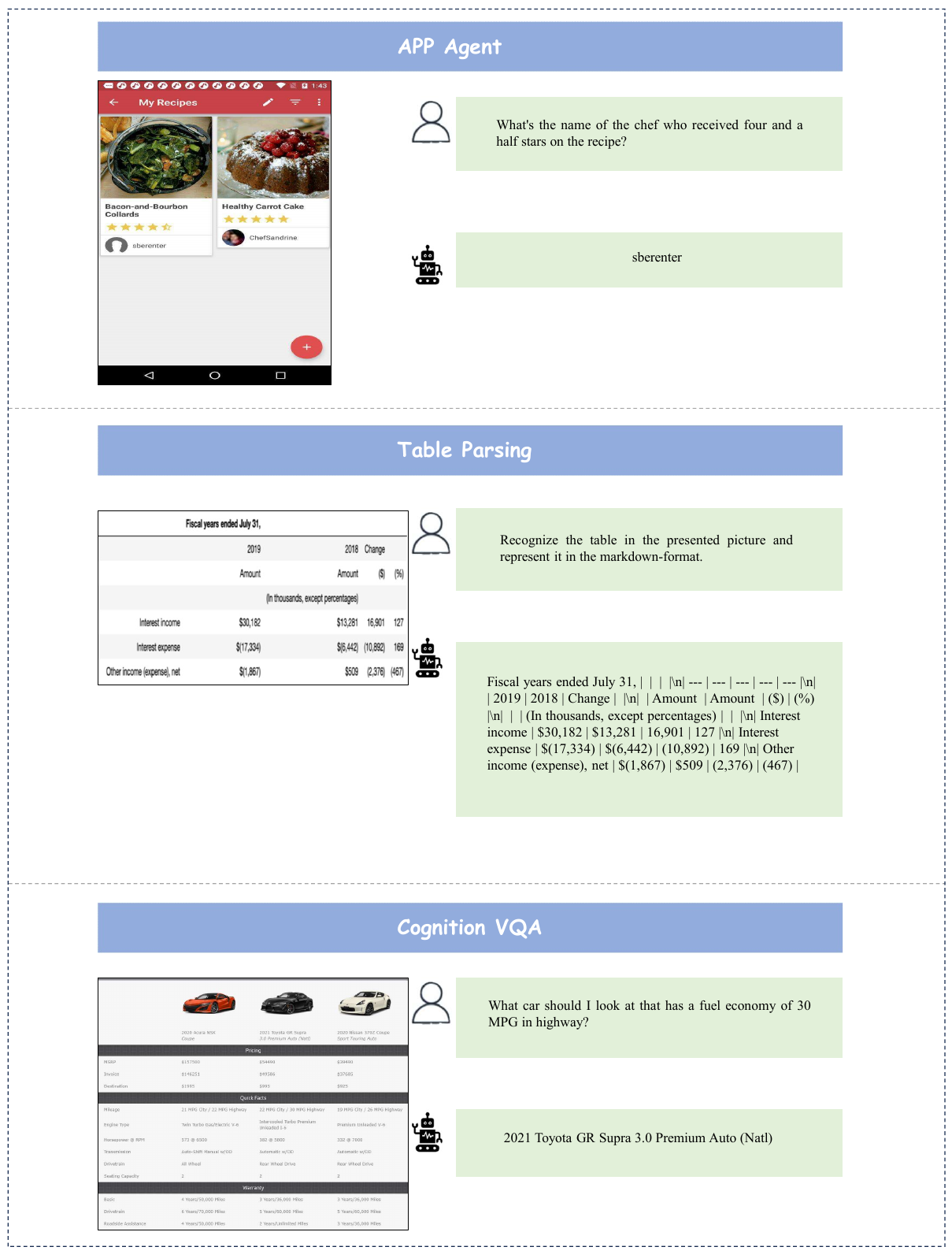}
  \caption{Samples for each task.}
   \label{fig:sample_tasks2}
\end{figure*}

\begin{figure*}[t]
  \centering   
  \includegraphics[width=0.98\linewidth]{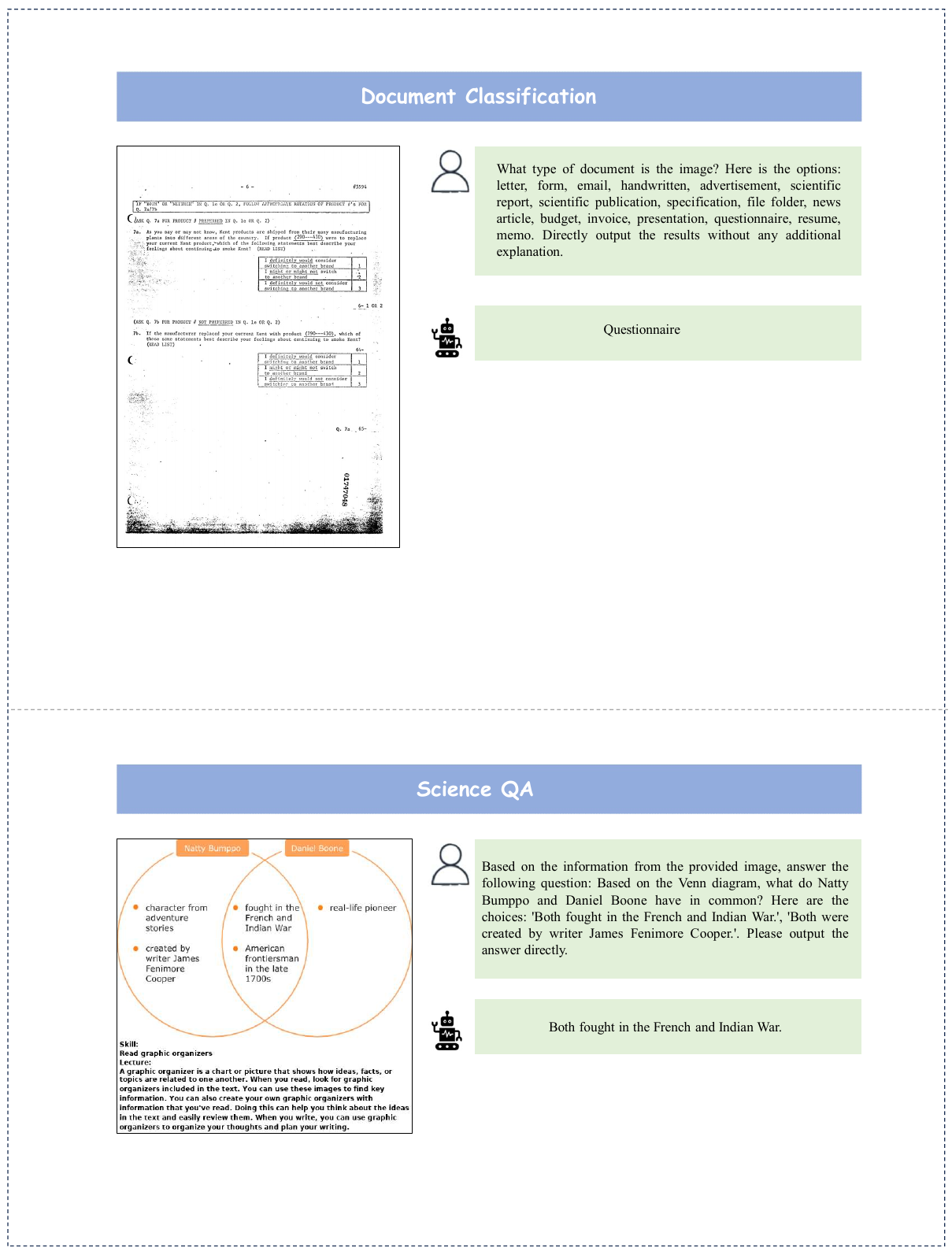}
  \caption{Samples for each task.}
   \label{fig:sample_tasks3}
\end{figure*}

\begin{figure*}[t]
  \centering   
  \includegraphics[width=0.98\linewidth]{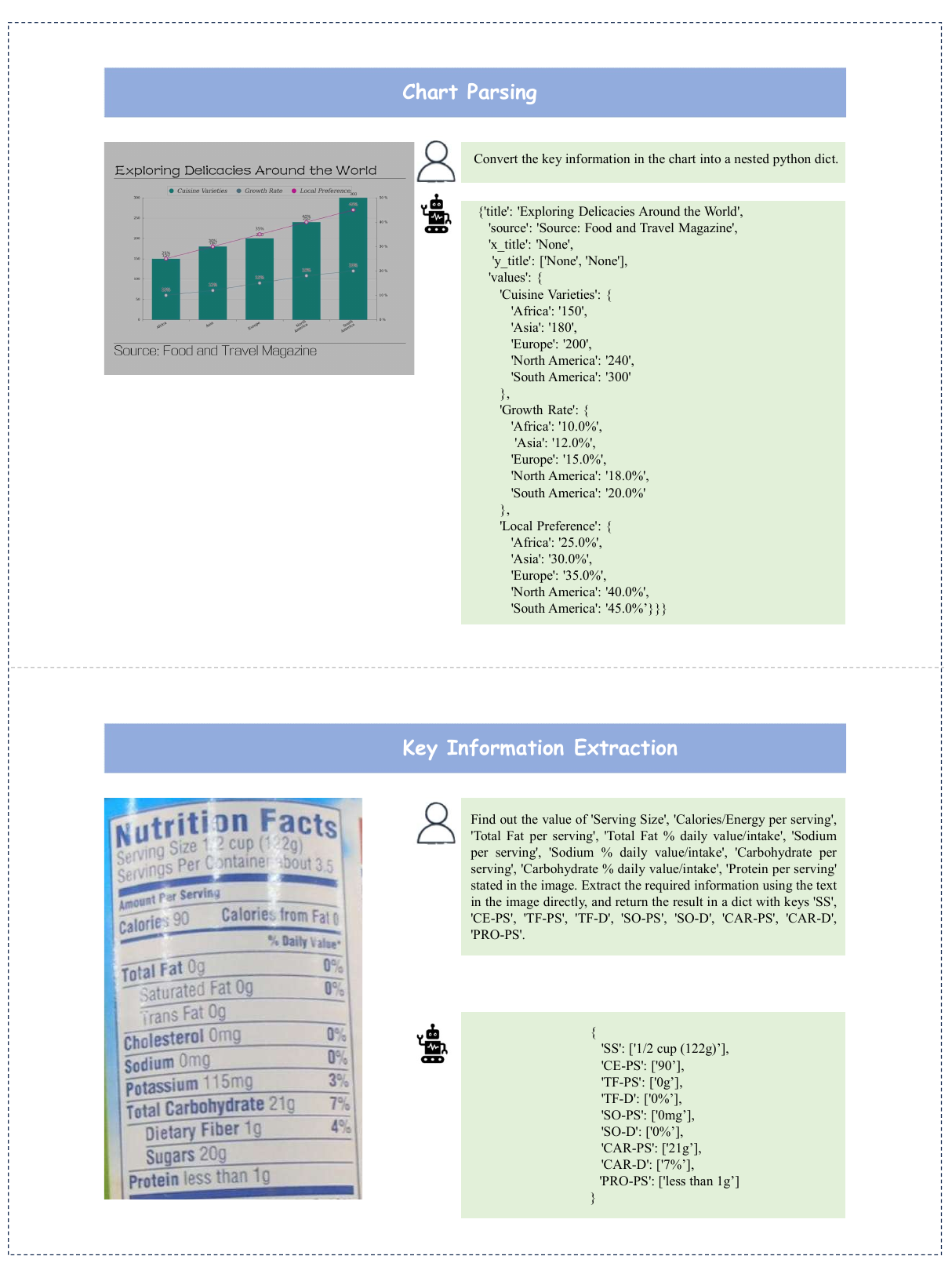}
  \caption{Samples for each task.}
   \label{fig:sample_tasks4}
\end{figure*}

\begin{figure*}[t]
  \centering   
  \includegraphics[width=0.98\linewidth]{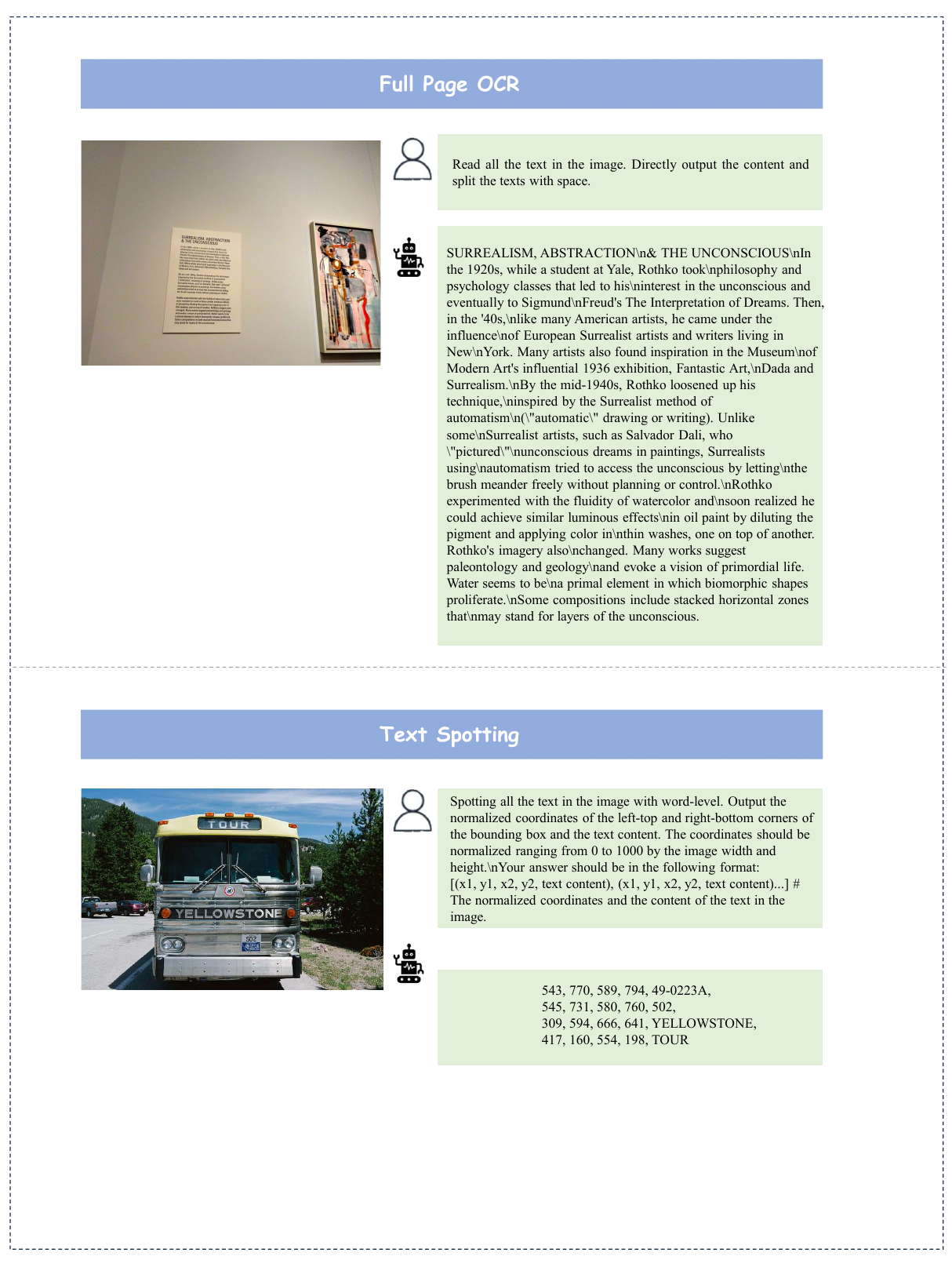}
  \caption{Samples for each task.}
   \label{fig:sample_tasks5}
\end{figure*}

\begin{figure*}[t]
  \centering   
  \includegraphics[width=0.98\linewidth]{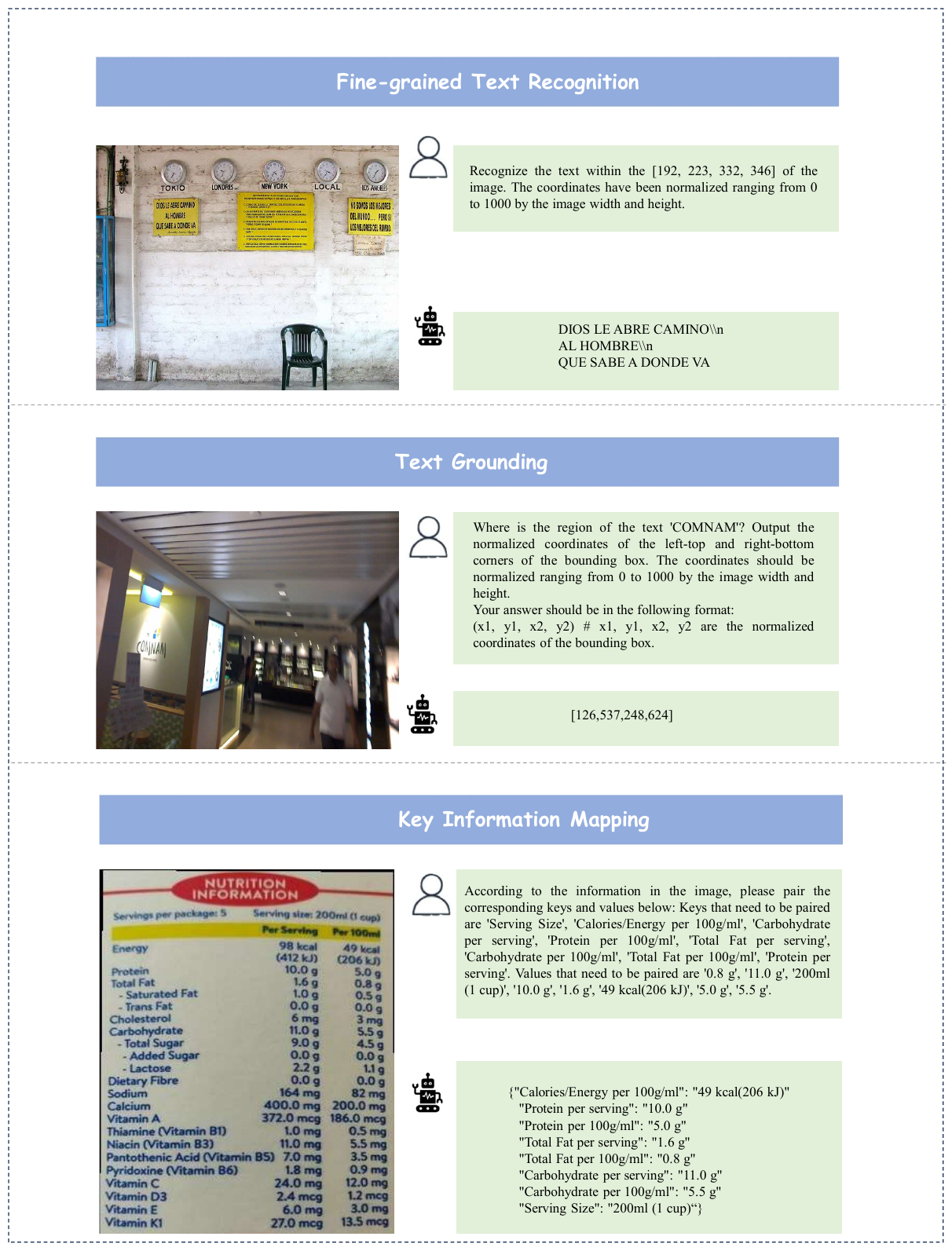}
  \caption{Samples for each task.}
   \label{fig:sample_tasks6}
\end{figure*}

\begin{figure*}[t]
  \centering   
  \includegraphics[width=0.98\linewidth]{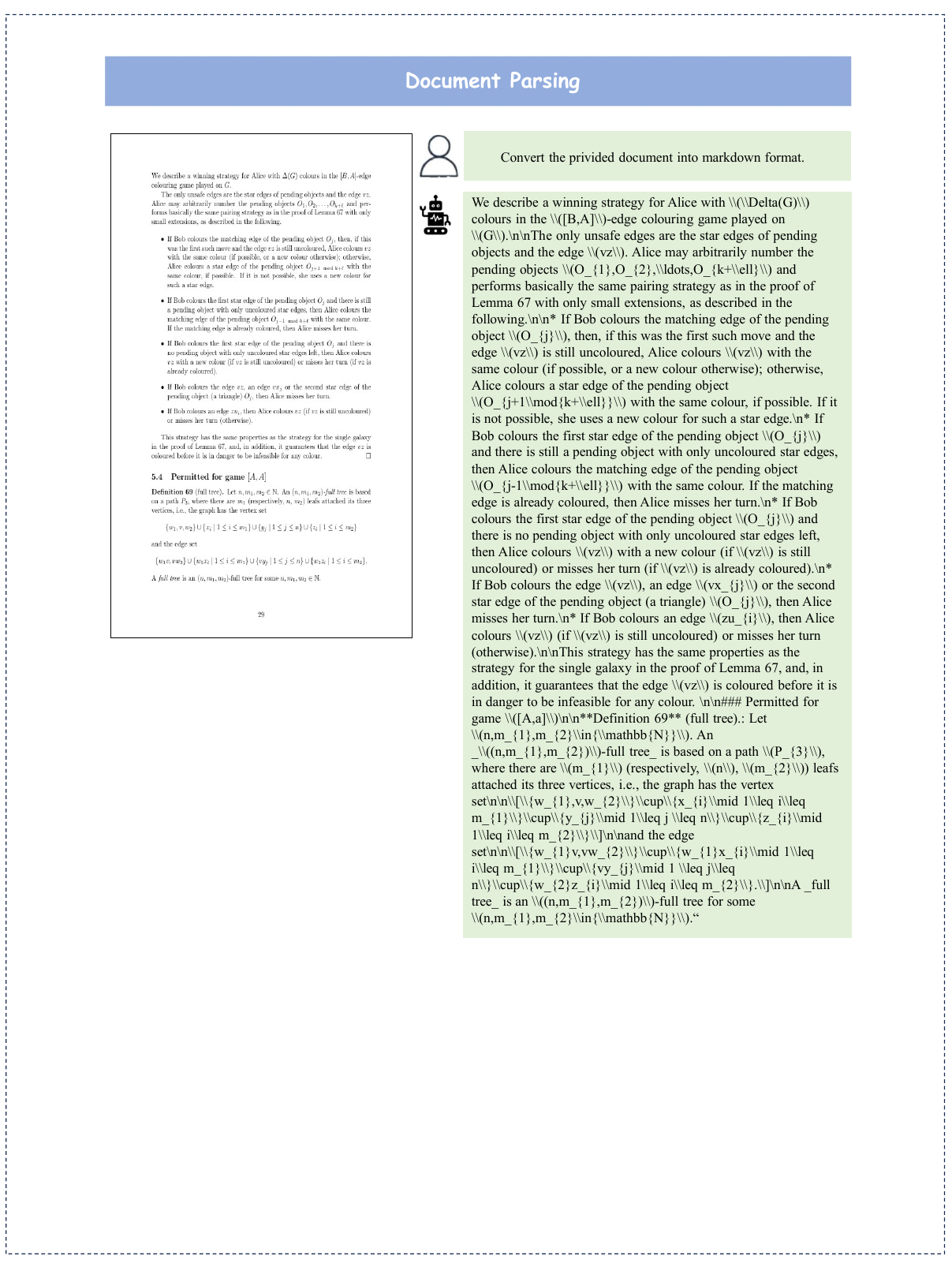}
  \caption{Samples for each task.}
   \label{fig:sample_tasks7}
\end{figure*}

\begin{figure*}[t]
  \centering   
  \includegraphics[width=\linewidth]{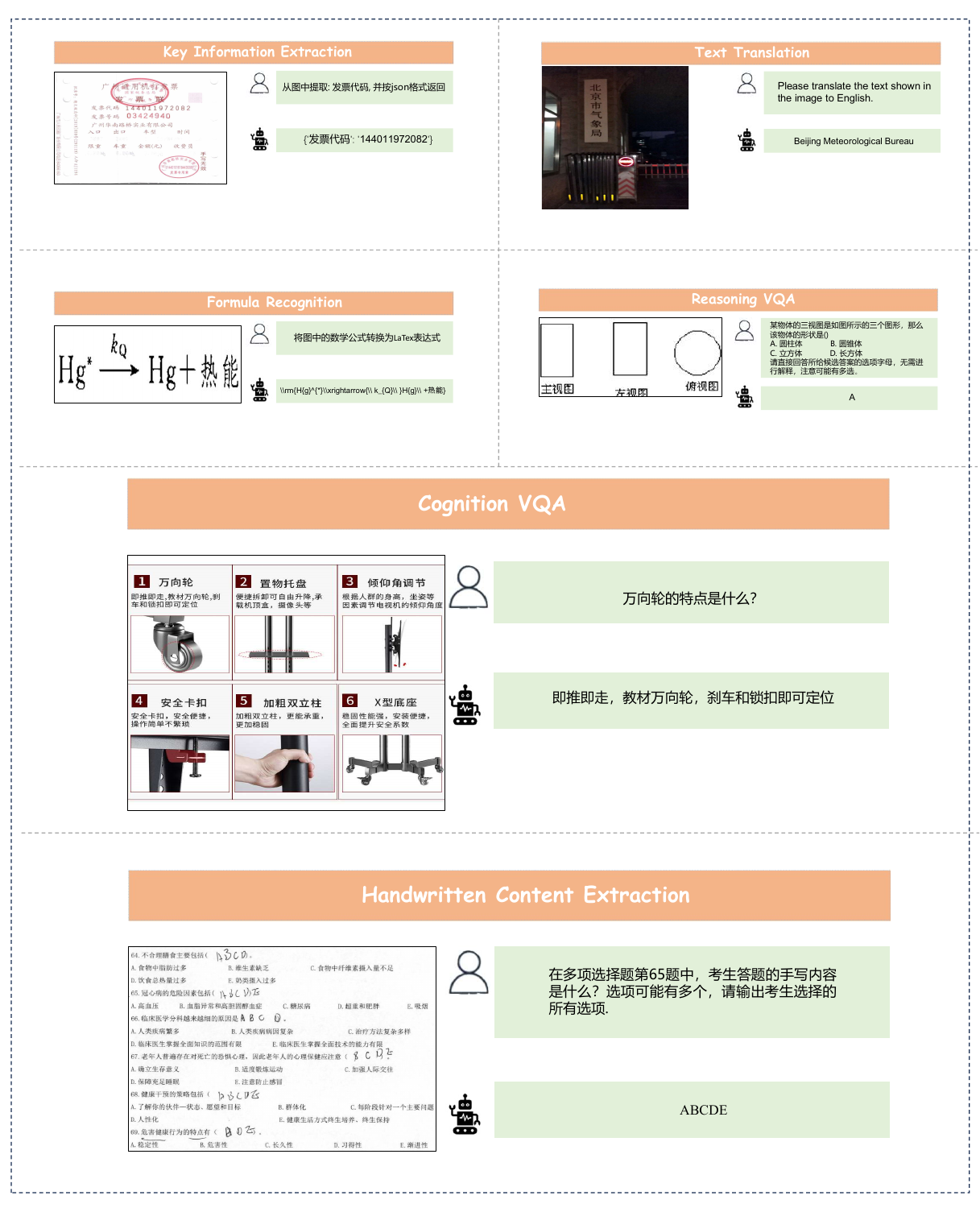}
  \caption{Samples for each task.}
   \label{fig:sample_tasks8}
\end{figure*}

\begin{figure*}[t]
  \centering   
  \includegraphics[width=0.98\linewidth]{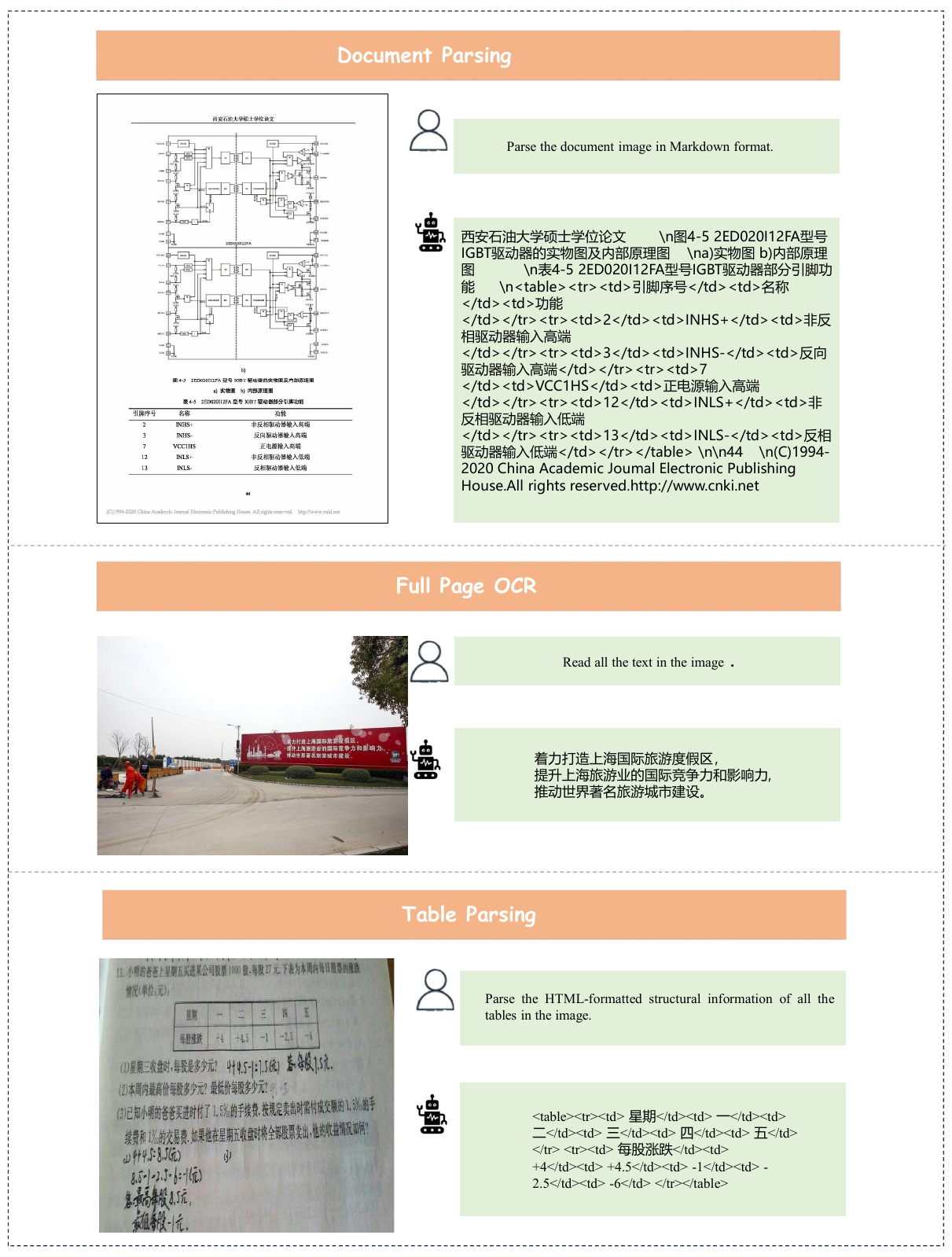}
  \caption{Samples for each task.}
   \label{fig:sample_tasks9}
\end{figure*}

As show in Fig.~\ref{fig:sample_tasks1} to Fig.~\ref{fig:sample_tasks9} , there are $23$ OCR tasks included in \textit{OCRBench v2}. Among them, Fig.~\ref{fig:sample_tasks1} to Fig.~\ref{fig:sample_tasks7} present examples of English tasks, including text recognition, diagram QA, text counting, formula recognition, math QA, VQA with position, ASCII art classification, reasoning VQA, text translation, APP agent, table parsing, cognition VQA, document classification, science QA, chart parsing, key information extraction, full-page OCR, text spotting, fine-grained text recognition, text grounding, key information mapping, and document parsing. These figures show corresponding images and QA pairs for each of the $23$ tasks. Fig.~\ref{fig:sample_tasks8} to Fig.~\ref{fig:sample_tasks9} provide examples of Chinese tasks, including key information extraction, text translation, formula recognition, reasoning VQA, cognition VQA, handwritten content extraction, document parsing, full-page OCR, and table parsing, along with their associated images and QA pairs.

\subsection{Samples for LMMs' Limitations}
\label{sample_for_limit}
\begin{figure*}[t]
  \centering   
  \includegraphics[width=\linewidth]{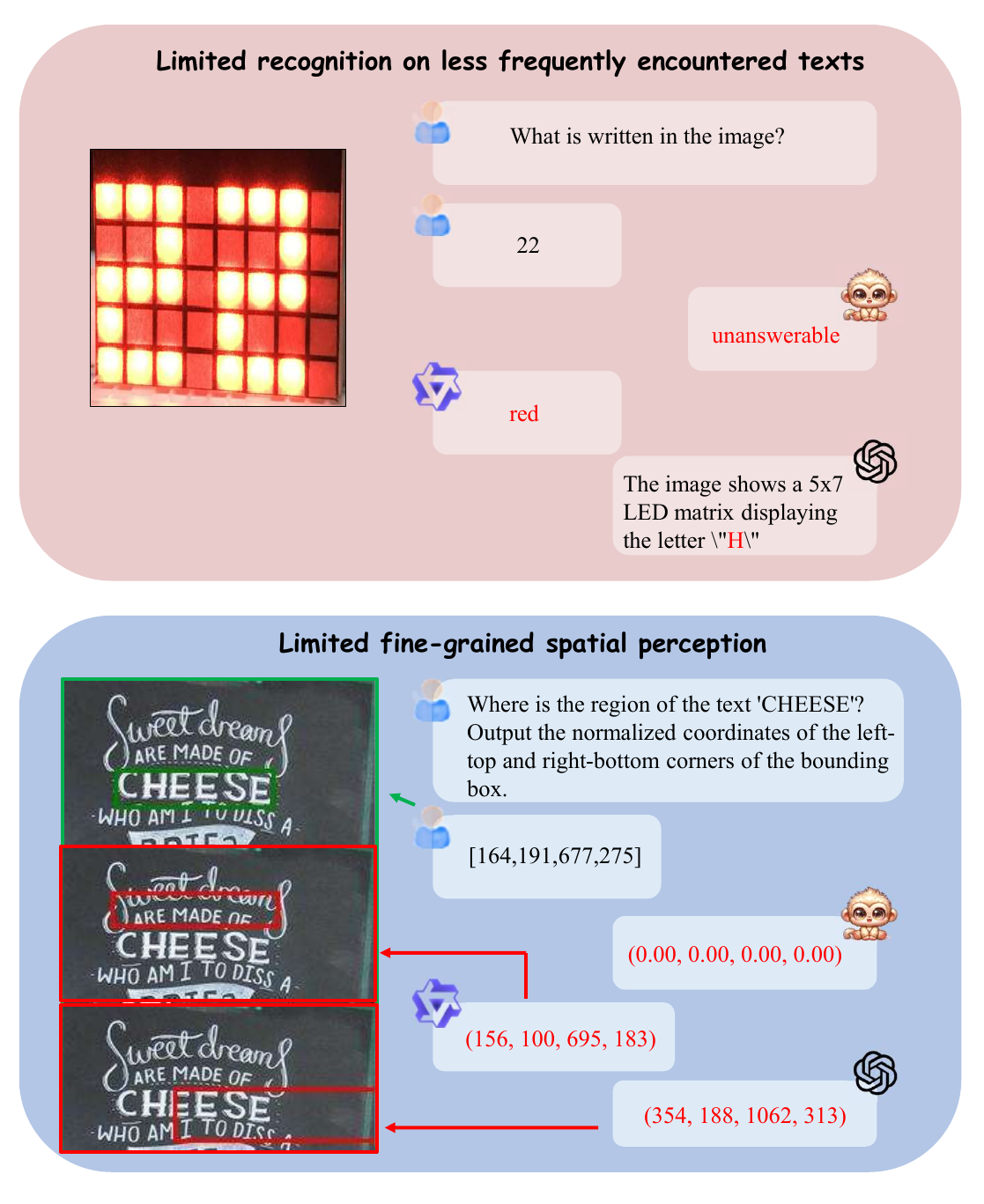}
  \caption{\textbf{Samples for LMM'S Limitations}. The portion of LLM's response marked in red is incorrect content, and the content in the red dashed box is missing information.}
   \label{fig:sample_limit1}
\end{figure*}

\begin{figure*}[t]
  \centering   
  \includegraphics[width=0.98\linewidth]{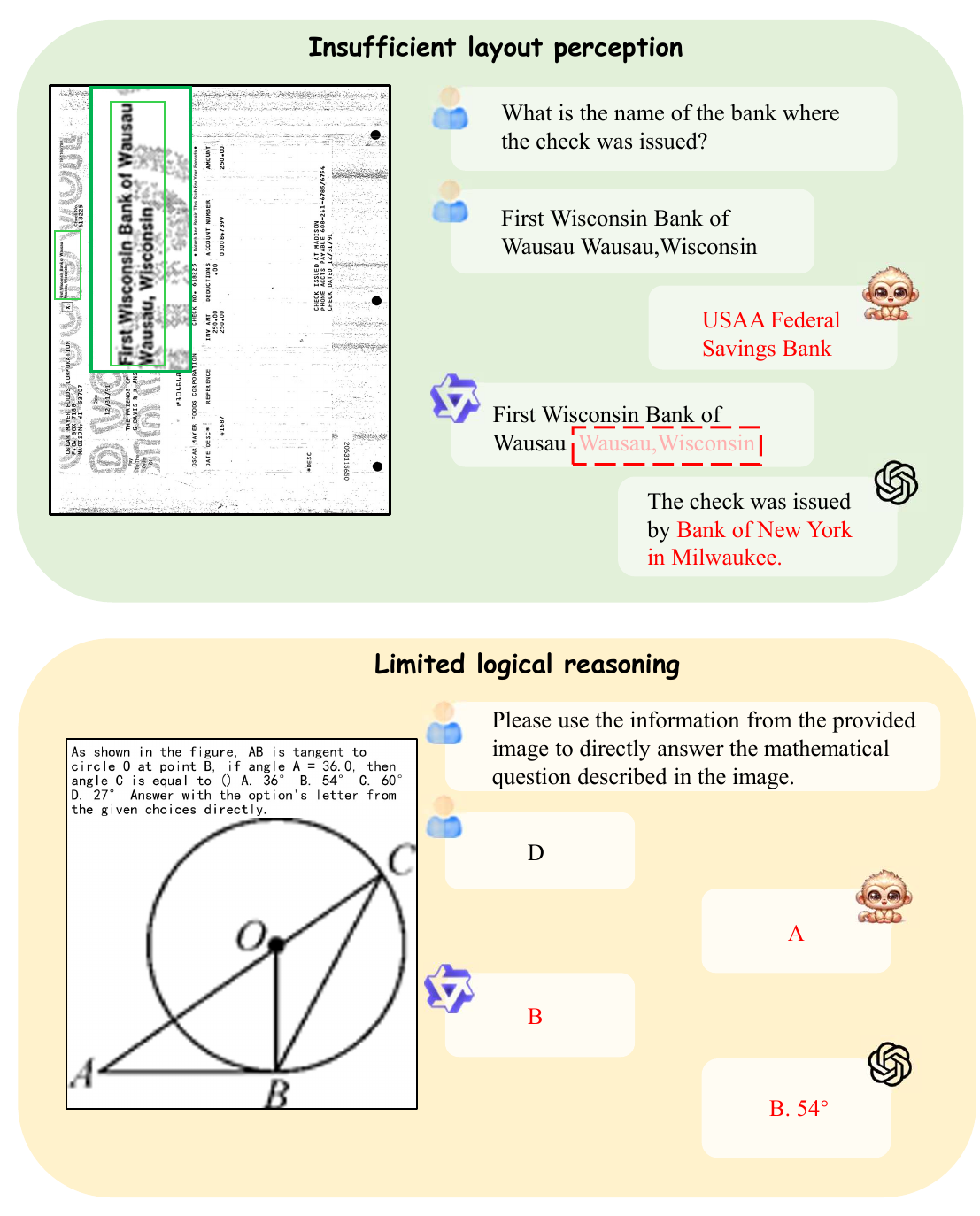}
  \caption{\textbf{Samples for LMM'S Limitations}. The portion of LLM's response marked in red is incorrect content, and the content in the red dashed box is missing information.}
   \label{fig:sample_limit2}
\end{figure*}

\begin{figure*}[t]
  \centering   
  \includegraphics[width=0.95\linewidth]{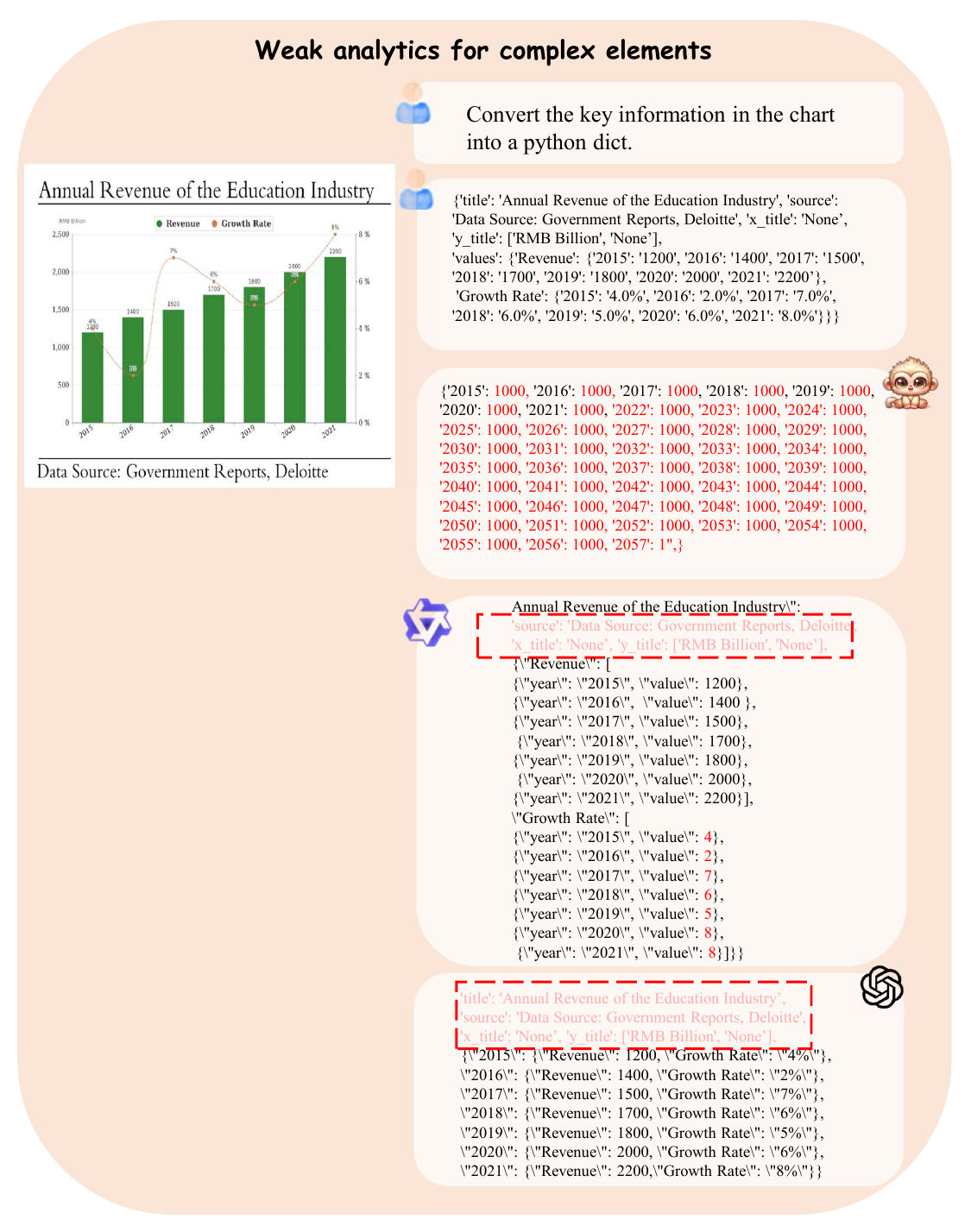}
  \caption{\textbf{Samples for LMM'S Limitations}. The portion of LLM's response marked in red is incorrect content, and the content in the red dashed box is missing information.}
   \label{fig:sample_limit3}
\end{figure*}

Fig.~\ref{fig:sample_limit1} to Fig.~\ref{fig:sample_limit3} provide examples corresponding to the findings discussed in Sec. 5.3 of the main text, which show error results of GPT-4o~\cite{achiam2023gpt}, Monkey~\cite{li2024monkey}, and Qwen2VL-8B on various tasks in \textit{OCRBench v2}. These examples highlight the current limitations of LLMs on OCR tasks. For instance, LLMs exhibit poor recognition of less frequently encountered texts, struggle to accurately locate text in tasks involving text and coordinates, and demonstrate insufficient perception of text in complex layouts such as rotated texts. Additionally, their logical reasoning abilities are limited when addressing mathematical problems, and their analysis of complex elements in charts remains weak. These are the capabilities of LLMs in OCR tasks that require further improvement.

\subsection{Broader Impacts}
\label{Broader impacts}
Our benchmark aims to enhance the evaluation of LMMs in text-oriented visual comprehension tasks. By establishing comprehensive benchmarks that reveal deficiencies in models' OCR capabilities, we provide insights for improving model performance. This advancement will elevate processing efficiency across scenarios such as document automation, assisted reading tools, and complex layout analysis, thereby benefiting applications in domains like healthcare and education. However, enhanced OCR functionality also introduces risks of misuse, including unauthorized extraction of sensitive information from images, surveillance-related applications, or generation of forged documents. To mitigate these risks, we restrict the use of this benchmark solely to research purposes and urge the community to prioritize privacy and fairness considerations in future model development.

\subsection{Limitations}
\label{Limitations}
One challenge we encountered is that LMMs sometimes produce responses that deviate from the given instructions, making it difficult to extract the desired answers. In future work, we plan to develop a more objective assessment framework to address this issue.

Another limitation arises when evaluating commercial LMMs, as some models occasionally refuse to answer certain questions due to safety filters or unclear content policies. This can lead to incomplete or biased performance assessments compared to open-source models that do not exhibit such behavior.

\end{document}